\documentclass[acmsmall]{acmart}
\usepackage{xcolor}
\usepackage[utf8]{inputenc}
\usepackage{enumitem}
\usepackage{booktabs,array}
\usepackage{tabularx}
\usepackage{graphicx}
\usepackage{amsthm}
\usepackage{amsmath}
\usepackage{makecell} % added by junjie
\usepackage{hyperref}  
\usepackage{multirow}
\usepackage{subcaption}
\usepackage{natbib}
\usepackage{wrapfig}

\newcommand{\suhang}[1]{\textcolor{blue}{SW: #1}}
\theoremstyle{definition}
\newtheorem{definition}{Definition}[section]

\title{A Comprehensive Survey on Trustworthy Graph Neural Networks: Privacy,  Robustness, Fairness, and Explainability}

\author{Enyan Dai}
\affiliation{%
 \institution{The Pennsylvania State University}
 \country{USA}}
 \email{emd5759@psu.edu}

\author{Tianxiang Zhao}
\affiliation{%
 \institution{The Pennsylvania State University}
 \country{USA}}
 \email{tkz5084@psu.edu}
 
\author{Huaisheng Zhu}
\affiliation{%
 \institution{The Pennsylvania State University}
 \country{USA}}
 \email{hvz5312@psu.edu}
 
\author{Junjie Xu}
\affiliation{%
 \institution{The Pennsylvania State University}
 \country{USA}}
 \email{jmx5097@psu.edu}
\author{Zhimeng Guo}
\affiliation{%
 \institution{The Pennsylvania State University}
 \country{USA}}
 \email{zhimeng@psu.edu}

\author{Hui Liu}
\affiliation{%
 \institution{Michigan State University}
 \country{USA}}
 \email{liuhui7@msu.edu}

\author{Jiliang Tang}
\affiliation{%
 \institution{Michigan State University}
 \country{USA}}
 \email{tangjili@msu.edu}
 
\author{Suhang Wang}
\affiliation{%
 \institution{The Pennsylvania State University}
 \country{USA}}
 \email{szw494@psu.edu}

\begin{document}
\begin{abstract}
Graph Neural Networks (GNNs) have made rapid developments in the recent years. Due to their great ability in modeling graph-structured data, GNNs are vastly used in various applications, including high-stakes scenarios such as financial analysis, traffic predictions, and drug discovery.  Despite their great potential in benefiting  humans in the real world, recent study shows that GNNs can leak private information, are vulnerable to adversarial attacks, can inherit and magnify societal bias from training data and lack interpretability, which have risk of causing unintentional harm to the users and society. For example, existing works demonstrate that attackers can fool the GNNs to give the outcome they desire with unnoticeable perturbation on training graph. GNNs trained on social networks may embed the discrimination in their decision process, strengthening the undesirable societal bias. Consequently, trustworthy GNNs in various aspects are emerging to prevent the harm from GNN models and increase the users' trust in GNNs. In this paper, we give a comprehensive survey of GNNs in the computational aspects of privacy, robustness, fairness, and explainability. For each aspect, we give the taxonomy of the related methods and formulate the general frameworks for the multiple categories of trustworthy GNNs. We also discuss the future research directions of each aspect and connections between these aspects to help achieve trustworthiness.
\end{abstract}


\maketitle

% \section{Outlines for Trustworthy GNNs}
\section{Introduction}
Graph-structured data such as bioinformatics network~\cite{kawahara2017brainnetcnn}, trading network~\cite{wang2021review}, and social network~\cite{hamilton2017inductive} are pervasive in the real-world. Inspired by the great success of deep learning on independent and identically distributed (i.i.d) data such as images, Graph Neural Networks (GNNs)~\cite{kipf2016semi,chen2018fastgcn,xiao2021learning,zhao2022exploring} are investigated to generalize deep neural networks to model graph-structured data. GNNs have shown great performance for various applications across various domains including finance~\cite{lv2019auto,harl2020explainable}, healthcare~\cite{li2021braingnn} and social analysis~\cite{fan2019graph,tan2019deep}. The success of GNNs relies on the message-passing
mechanism, where node representations are updated by aggregating the information from neighbors. With this mechanism, node representations can capture node features, information of neighbors and local graph structure, which facilitate various graph mining tasks, such as node classification~\cite{kipf2016semi}, link prediction~\cite{chen2018link} and graph classification~\cite{jin2019power}.

Despite their achievements in modeling graphs, the concerns in the trustworthiness of GNNs are rising. 
Firstly, GNN models are vulnerable to the attacks that steal the private data information or affect the behaviors of the model. For example, hackers can utilize the embeddings of nodes to infer their attribute information and friendship information in social network~\cite{he2021stealing,zhang2021inference}. They also can easily fool the GNNs to give target prediction to a node by injecting malicious nodes to the network~\cite{sun2019node}. 
Secondly, GNN models themselves have problems in fairness and interpretability. More specifically, GNN models can magnify the bias in the training data, resulting discrimination towards the people with certain genders, skin colors, and other protected sensitive attributes~\cite{dai2021say,buyl2020debayes}. 
Finally, due to the high nonlinearity of the model, predictions from the GNNs are difficult to understand.  The lacking of interpretability also make the GNNs untrustworthy, which largely limit the applications of GNNs. Those weaknesses significantly hinder the adoption of GNNs in real-world applications, especially those high-stake scenarios such as finance and healthcare. Therefore, how to build trustworthy GNN models has become a focal topic. 

\begin{figure}
    \centering
    \includegraphics[width=0.85\linewidth]{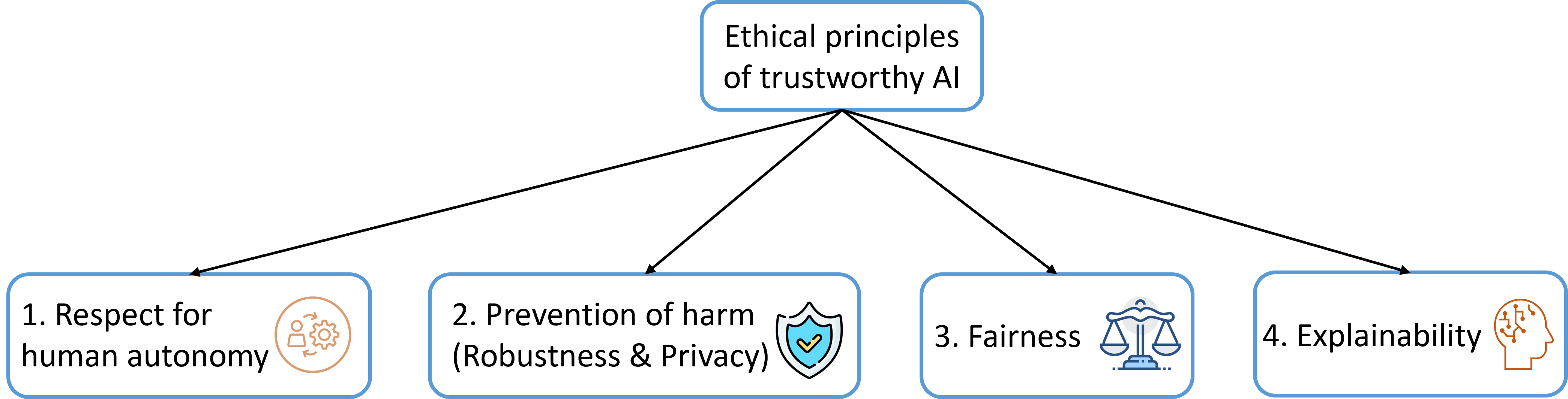}
    \vskip -1em
    \caption{The ethical principles of trustworthy AI.}
    \vskip -1em
    \label{fig:guideline}
\end{figure}

Recently, a guideline of trustworthy AI system have been proposed by the European Union~\cite{smuha2019ethics}. As shown in Figure~\ref{fig:guideline}, the guideline indicates that trustworthy AI should obey the following four ethical principles: \textit{Respect for human autonomy}, \textit{Prevention of harm}, \textit{Fairness}, and \textit{Explainability}. The principle of respect for human autonomy requires AI systems to follow human-centric design principles and leave meaningful opportunity for human choice. This generally fails in the domain of human-computer interaction. Therefore, we do not focus on this direction of trustworthy GNNs in this survey. According to the principle of prevention of harm, AI systems should be technically robust and be ensured not open to malicious use, which corresponds to the robustness and privacy aspects of our survey. The principle of fairness requires that AI systems should ensure the individuals and groups free from unfair bias, discrimination and stigmatisation. As for the explainability, it requires the decision process of AI to be transparent and explainable. It is worth to mention that the four aspects are not isolated with each other. For instance, the attacker may poison the training data to degrade the fairness of the model~\cite{solans2020poisoning,mehrabi2020exacerbating} or mislead the GNN explainer model~\cite{fan2021jointly}. And the explanations from explainable GNN methods can also be helpful for other aspects. Specifically, based on the explanations from the explainable GNN model, the human can debug the model to avoid the adversarial attacks. In addition, with the analysis of the explanations, we can evaluate whether the deployed model is giving biased predictions. Therefore, it is important to explore the connections of these aspects to finally achieve trustworthy GNNs that simultaneously address the concerns in robustness, privacy, fairness, and explainability. In this survey, we also have some discussions about the interactions of the trustworthiness aspects in the future directions.

Due to the demand for trustworthy GNNs,  a large number of literature in different aspects of trustworthy GNNs are emerging in recent years. For example, robust GNN against the perturbations from attackers have been developed~\cite{wu2019adversarial,jin2020graph,dai2022towards}. To prevent the private information, Privacy-preserving GNN models~\cite{liao2021information,wu2021fedgnn} are also proposed  in various real-world applications such as financial analysis. Fair GNNs~\cite{dai2021say} and explainable GNNs~\cite{ying2019gnnexplainer,dai2021towards} also become hot topics to address the concerns in trustworthiness. There are several surveys of GNNs in robustness~\cite{sun2018adversarial,wu2022recent,jin2020adversarial,zhu2021deep}, explainability~\cite{yuan2021explainability}, and fairness~\cite{dong2023fairness}. However, none of them thoroughly discuss about the trustworthiness of GNNs, which should also cover the dimensions of privacy and fairness. For the aspects of robustness and explainability, they also do not include the emerging directions and techniques such as scalable attacks, backdoor attacks, and self-explainable GNNs, which are discussed in this survey. A recent survey~\cite{liu2021trustworthy} gives a review about the trustworthy AI systems. But it mainly focus on the techniques of trustworthy AI systems on i.i.d data. Considering the complexity of the graph topology and the deployment of message-passing mechanism in GNNs, the trustworthy AI designed for i.i.d data generally can not be adopted to process graph-structured data. 
There is a concurrent survey in trustworthy in graph neural networks~\cite{zhang2022trustworthy}. Compared with~\cite{zhang2022trustworthy}, we cover more recent advanced topics of trustworthiness such as machine unlearning, model ownership verification, scalable adversarial attacks, fair contrastive learning, explanation-enhanced fairness, and self-explainable GNNs. To summarize, our major contributions are:
\begin{itemize} [leftmargin=*]
    \item In Section~\ref{sec:privacy}, we give a comprehensive survey of the existing works in privacy attacks and defense on GNNs followed by the future directions. The graph datasets in privacy domain are also listed. 
    \item Various categories of adversarial attack and defense methods on GNN models are discussed in Section~\ref{sec:robust}. Some recent advances about the robustness of GNN such as scalable attacks, graph backdoor attacks, and self-supervised learning defense methods are further introduced.
    \item The fairness of trustworthy GNNs is thoroughly discussed in the Section~\ref{sec:fairness} of this survey, which includes the the biases and fairness definitions on graph-structured data, vairous fair GNN models and the datasets they applied. 
    \item A comprehensive survey of GNN explainability is presented in Section~\ref{sec:explain}, in which we go through the motivations, challenges, and experiment settings adopted by existing works. A taxonomic summary of methodologies is also introduced.
\end{itemize}

% \suhang{avoid using users as (i) user can be a user in social network or the end-user of GNN models; (ii) we are talking about general case, the graph can be from various domains. use node or other words. Let's have a section here to introduce the necessary background on graph neural networks, e.g., graph and GNN}

\section{Preliminaries of Graph Neural Networks}
To facilitate the discussion of trustworthy GNNs, we firstly introduce notations, the basic design of GNNs, and graph analysis tasks in this section.
\subsection{Notations} \label{sec:notations}
We use $\mathcal{G}=(\mathcal{V},\mathcal{E})$ to denote a graph, where $\mathcal{V}=\{v_1,...,v_N\}$ is the set of $N$ nodes, $\mathcal{E} \subseteq \mathcal{V} \times \mathcal{V}$ is the set of edges. 
The graph can be either attributed or plain graph. For attributed graph, node attributes $\mathbf{X}=\{\mathbf{x}_1,...,\mathbf{x}_N\}$ are provided, where $\mathbf{x}_i \in \mathbb{R}^d$ corresponds to the $d$-dimensional attributes of node $v_i$. $\mathbf{A} \in \mathbb{R}^{N \times N}$ is the adjacency matrix of the graph $\mathcal{G}$, where $\mathbf{A}_{ij}=1$ if nodes ${v}_i$ and ${v}_j$ are connected; otherwise, $\mathbf{A}_{ij}=0$. 
% In the semi-supervised node classification setting, part of nodes $v \in \mathcal{V}_L$ are provided with labels $y_v \in \mathcal{Y}$, where $\mathcal{V}_L \subseteq \mathcal{V}$ denotes nodes with labels, and $\mathcal{Y}$ is the set of labels. As for the 

\subsection{Inner Working of Graph Neural Networks}
\label{sec:design_GNN}
Apart from the node features, the graph topology also offers crucial important information for representation learning. Generally, GNNs adopt the message-passing mechanism to learn node representations that capture both node features and graph topology information. Specifically, in 
each layer, GNNs will update the representations of a node by aggregating the information from their neighborhood nodes. As a result, a $k$-layers GNN model would capture the information of the local graph containing $k$-hop neighbors of the central nodes. The general form of updating representations in the $k$-th layer of GNNs can be formulated as:
\begin{equation}
\begin{aligned}
    \mathbf{h}^{(k)}_{v} & =\text{COMBINE}^{(k)}(\mathbf{h}^{(k-1)}_v, \text{AGGREGATE}^{(k-1)}(\{\mathbf{h}^{(k-1)}_u: u \in \mathcal{N}(v)\})),
    \label{eq:GNN}
    % \label{eq:GNN_h}
\end{aligned}
\end{equation}
where $\mathbf{h}^{(k)}_v$ stands for the representation of node $v \in \mathcal{V}$ after the $k$-th GNN layer and $\mathcal{N}(v)$ denotes  the set of neighborhoods of node $v$. For node classification, a linear classifier can be applied to the representation $\mathbf{h}_v$ to predict the label of node $v$. For graph classification, a READOUT function will summarize the node embeddings to a graph embedding $\mathbf{h}^G$ for future predictions:
\begin{equation}
    \mathbf{h}^G = \text{READOUT}(\{\mathbf{h}_v^{K}|v \in \mathcal{G}\}),
    \label{eq:readout}
\end{equation}
where READOUT can be various graph pooling functions such as max pooling and average pooling. Similar to node classification, graph classification can be conducted by applying a linear classifier on the graph embedding $\mathbf{h}^G$. 

Extensive graph neural networks that follow the Eq.(\ref{eq:GNN}) have been proposed. Here, we only introduce the design of GCN~\cite{kipf2016semi}, which is one the most popular GNN architectures. For the design of other GNNs, please refer to the survey of GNN models~\cite{zhou2020graph}.
More specifically, each layer of GCN can be written as:
\begin{equation}
    \mathbf{H}^{(k)} = \sigma(\tilde{\mathbf{A}}\mathbf{H}^{(k-1)}\mathbf{W}^{(k)}),
\end{equation}
where $\mathbf{H}^{(k)}$ denote the representations of all the nodes after the $k$-th layer; $\mathbf{W}^{(k)}$ stands for the  parameters of the $k$-th layer. $\tilde{\mathbf{A}}$ is the normalized adjacency matrix. Generally, the symmetric normalized form is used, which can be written as $\tilde{\mathbf{A}}=\mathbf{D}^{-\frac{1}{2}}(\mathbf{A}+\mathbf{I})\mathbf{D}^{-\frac{1}{2}}$, and $\mathbf{D}$ is a diagonal matrix with $D_{ii}=\sum_{i}A_{ij}$. $\mathbf{I}$ is the identity matrix. $\sigma$ is the activation function such as ReLU.

\subsection{Graph Analysis Tasks}
The learned node representations by GNNs can facilitate various tasks, such as node classification, link prediction, community detection and graph classification. Next, we briefly introduce them. 
% \suhang{add community detection}  \suhang{add citations of the applications in the paragraphs below}

\vspace{0.2em} \noindent \textbf{Node Classification.} Many real-world problems can be treated as the node classification problem such as  such as  user attribute prediction in social media~\cite{dai2021say,zhao2021graphsmote}, fraud detection in transaction networks~\cite{wang2021review,xu2021towards}, and protein function prediction on protein-protein interaction networks~\cite{yang2020graph}. 
In node-level classification, the GNN model aims to infer the labels of the test nodes $\mathcal{V}_T$  given the graph $\mathcal{G}=(\mathcal{V}, \mathcal{E})$.
Generally, the node classification task is semi-supervised where only partial nodes $\mathcal{V}_L \in \mathcal{V}$ of the are provided with labels $\mathcal{Y}$. 
Based on whether the test samples $\mathcal{V}_T$ are seen during the training phase, node classification can be split into \textit{transductive setting} and \textit{inductive setting}. In transductive setting, the test nodes $\mathcal{V}_T$ are available during the training phase. The node features of the test nodes can be utilized for better prediction performance. In contrast, in inductive setting the test nodes are totally new for the trained GNN model.

\vspace{0.2em} \noindent \textbf{Community Detection.} Communities are subgraphs of a network, which are more densely connected to each other than the rest nodes in the network. Formally, the set of communities in the network can be represented by $\{\mathcal{C}_1,\dots \mathcal{C}_K\}$, where $\mathcal{C}_i \subset \mathcal{G}$ is a partition of the whole graph $\mathcal{G}$. These communities could be either disjointed or overlapping. The goal of the community detection is to identity which communities each node $v \in \mathcal{G}$ belongs to. Community detection is often unsupervised~\cite{zhang2019attributed,wang2021unsupervised,shchur2019overlapping}. Recently, supervised community detection based on GNNs are also investigated~\cite{chen2017supervised}. Community detection can be useful for various domains such as social network analysis~\cite{girvan2002community} and functional region identification in brain~\cite{garcia2018applications}.

\vspace{0.2em} \noindent \textbf{Link Prediction.} Link Given a graph $\mathcal{G}=(\mathcal{V}, \mathcal{E})$, the link prediction model will predict the existence of link between nodes $v_i \in \mathcal{V}$ and $v_j \in \mathcal{V}$, where $(v_i, v_j) \notin \mathcal{E}$. A very common form of link prediction is to give prediction based on the representations of two nodes from a GNN model. Let $\mathbf{h}_i$ and $\mathbf{h}_j$ denote the representations of node $v_i$ and $v_j$, it can be formulated as $g(v_i,v_j) = MLP(\mathbf{h}_i, \mathbf{h}_j)$.
Link prediction have various applications such as friend recommendation on social media~\cite{sankar2021graph} and knowledge graph completion~\cite{arora2020survey}.

\vspace{0.2em} \noindent \textbf{Graph Classification.} For graph classification, each graph instance belongs to a certain class. The training set of the graph classification can be denoted as $\mathcal{D}_T = \{(\mathcal{G}_i, y_i)\}_{i=1}^{|\mathcal{D}_T|}$, where $y_i$ denotes the label of graph $\mathcal{G}_i$ and $|\mathcal{D}_T|$ represents the number of graphs in the training set. The goal of the graph classification is to learn a function $f_\theta: \mathcal{G} \rightarrow y$ to classify the unlabeled test graphs $\mathcal{D}_U$. As it is mentioned in Section~\ref{sec:design_GNN}, a READOUT function is added to the GNN model to obtain graph embedding for classification. Similarly, there are many applications of graph classification such as property prediction of drugs~\cite{jiang2021could} where each drug is represented as a graph.

\section{Privacy of Graph Neural Networks} \label{sec:privacy}
Similar to deep learning algorithms on images and texts, the remarkable achievements of GNNs also rely on the big data.  Extensive sensitive data are collected from users to obtain powerful GNN models for various services in critical domains such as healthcare~\cite{li2020graph}, banking systems~\cite{wang2021review}, and bioinformatics~\cite{li2021braingnn}. For example, GNNs have been applied on brain networks for FMRI analysis~\cite{li2021braingnn}.  In addition, the GNN model owner may provide the query API service to share the knowledge learned by GNNs. It is also very common that the pretrained GNN models are released to third parties for knowledge distillation or various downstream tasks~\cite{long2020pre}. 
However, the collection and utilization of private data for GNN model training, the API service and model release are threatening the safety of the private and sensitive information.  \textit{First}, GNNs are generally trained in a centralized way, where the users' data and models are stored in the centralized server. In case of an untrustworthy centralized server, the collected sensitive attributes might be leaked by unauthorized usage or data breach. 
For instance, the personal data of more than half a billion Facebook users was leaked online for free in a hacker forum in 2021\footnote{https://www.bbc.com/news/technology-56745734}.
\textit{Second}, the private information of users can also be leaked from model release or the provided services due to privacy attacks. Taking the online service for brain disease classification as example,  membership inference attack can figure out the patients that covered in the training dataset, which severely threaten the privacy of patients. Moreover, various types of privacy attacks~\cite{duddu2020quantifying,zhang2021inference} such as link inference and attribute inference have been proved effective to steal users' information from the pretrained model.
Therefore, it is crucial to develop privacy-preserving GNNs to achieve trustworthiness. 

% From a legal perspective, laws from the state level to the global level have begun to provide mandatory
% regulations for data privacy.
% Therefore, 
There are several surveys on the privacy aspect of machine learning model. In~\cite{rigaki2020survey}, it comprehensively reviews the current privacy attacks in i.i.d data. Privacy-preserving methods are reviewed in~\cite{al2019privacy,kairouz2019advances,ji2014differential,yang2020local} as well. However, they are overwhelmingly dedicated to the privacy issue on models for i.i.d data such as images and text, and rarely discuss the privacy attacks and defense methods on graphs; while these methods are challenged by the topology information in graphs and the message-passing mechanism of GNN models. Therefore, in this section, we give the overview of the privacy attacks on GNNs and privacy-preserving GNNs to defend against privacy attacks. We also include related datasets and the applications of privacy-preserving GNNs followed by future research directions on privacy-preserving GNNs. 

\subsection{Taxonomy of Privacy Attacks}
In this subsection, we will introduce the categorization of privacy attacks on GNNs according to the target private information. We will also briefly explain two settings of the attacker's accessible knowledge for conducting privacy attacks. Finally, we will present more details of existing privacy attack methods on GNNs. 

\begin{table}[t]
    % \small
    \scriptsize	
    \centering
    \caption{Different types of privacy attack methods on GNNs.}
    \vskip -1.5em
    \begin{tabularx}{0.65\linewidth}{XX}
    \toprule
    Privacy attack types &  References \\
    \midrule
    Membership inference &
    \cite{duddu2020quantifying},~\cite{olatunji2021membership},~\cite{he2021node},~\cite{wu2021adapting} \\
    \midrule
    Property Inference & \cite{zhang2021inference} \\
    \midrule 
    Reconstruction attack & \cite{he2021stealing},~\cite{zhang2021inference},~\cite{duddu2020quantifying},~\cite{zhang2021graphmi}\\
    \midrule
    Model extraction & \cite{wu2020model},~\cite{shen2021model}
    \\
    \bottomrule
    \end{tabularx}
    \label{tab:privacy_attack}
    \vskip -1em
\end{table}

\subsubsection{Types of Privacy Attacks on GNNs}
\label{sec:privacy_attack_types}
The goal of privacy attacks on GNNs is to extract information that is not intended to be shared. The target information can be about the training graph such as the membership, sensitive attributes of nodes, and connections of nodes. In addition, some attackers aim to extract the model parameters of GNNs. Based on the target knowledge, the privacy attacks can generally be split into four categories:
\begin{itemize}[leftmargin=*]
    \item \textbf{Membership Inference Attack}: In membership inference attack, the attackers try to determine whether a target sample is part of the training set. For example, suppose researchers train a GNN model on social network of COVID-19 patients to analyze the propagation of virus. The membership inference attack can identify if a target subject is in training patient network, resulting in information leakage of the subject. 
    Different from i.i.d data, the format of the target samples can be nodes or graphs. For instance, for node classification task, the target samples can be subgraph of the target node's local graph~\cite{olatunji2021membership} or only contain the node attributes~\cite{he2021node}. For graph classification task, the target sample is a graph to be classified~\cite{wu2021adapting}. 
    \item \textbf{Reconstruction Attack}: Reconstruction attack, also known as model inversion attack, aims to infer the private information of the input graph. Since the graph-structure data is composed of graph topology and node attributes, the reconstruction attack on GNNs can be split into \textit{structure reconstruction}, i.e., infer the structures of target samples, and \textit{attribute reconstruction} (also known as attribute inference attack), i.e., infer the attributes of target samples. Generally, the embeddings of the target samples are required to conduct the reconstruction attack. 
    \item \textbf{Property Inference Attack}: Different from attribute reconstruction attack, property inference attack aims to infer dataset properties that are not encoded as features. For instance, one may want to infer the ratio of women and men in a social network, where this information is not contained in node attributes. The attacker may also be interested in structure-related properties such as degrees of a node, which is the number of friends of the target user in a social network~\cite{zhang2021inference}.
    \item \textbf{Model Extraction Attack}: This attack aims to extract the target model information by learning a model that behaves similarly to the target model. It may focus on different aspects of the model information, which results in two goals in model extraction: (i) The attacker aims to obtain a model that matches the accuracy of the target model; (ii) The attacker tries to replicate the decision boundary of the target model.  Model Extraction Attack can threaten the security of model for API service~\cite{niu2020dual} and can be a stepping stone for various privacy attacks and adversarial attacks. 
\end{itemize}
Table~\ref{tab:privacy_attack} categorizes existing methods on privacy attacks on GNNs based on the attack types. We will introduce the details of these methods in Section~\ref{sec:atk_methods}.
\begin{table}[t]
    % \small
    \scriptsize	
    \centering
    \caption{Categorization of attacker's knowledge.}
    \vskip -1.5em
    \begin{tabular}{ll}
    \toprule
    Knowledge & References  \\
    \midrule
    White-box & ~\cite{duddu2020quantifying},~\cite{zhang2021graphmi} \\
    \midrule
    Black-box &~\cite{duddu2020quantifying},~\cite{olatunji2021membership},~\cite{he2021node},~\cite{wu2021adapting},~\cite{wu2020model},~\cite{shen2021model},~\cite{zhang2021inference},~\cite{he2021stealing}\\
    \midrule
    \end{tabular}
    \label{tab:privacy_know}
    \vskip -1em
\end{table}
\subsubsection{Threat Models of Privacy Attacks}
% \enyan{consider to explain other knowledge categories}

To conduct privacy attacks, auxiliary knowledge about the target GNN and/or dataset is usually possessed by attackers. In this subsection, we introduce the categorization of threat models of privacy attacks in the aspect of attackers' knowledge. 
Generally, based on whether the model parameters of the target GNN are available,  attacker's knowledge about the threat model can be split into two settings, i.e., white-box attack and black-box attack:
\begin{itemize}[leftmargin=*]
    \item \textbf{White-Box Attack}: In white-box attack, the model parameters or the gradients during training is accessible for the attackers. %This can be a practical adversary assumption in federated learning~\cite{duddu2020quantifying} and pre-trained GNNs releasing~\cite{rong2020self}, where the intermediate computations or the model parameters are shared. 
    Apart from the knowledge about the trained GNNs, the attacks may require some other knowledge such as the nodes/graphs to be attacked in inference attacks and a shadow dataset, i.e., dataset that follows the same distribution as the training dataset of the target GNN. The white-box attack can be used to attack pretrained GNNs whose model is publicly releasedg~\cite{rong2020self}. It is also practical during the training process of federated learning~\cite{duddu2020quantifying} where the intermediate computations.
    \item \textbf{Black-Box Attack}: In contrast to white-box attack, the parameters of the target GNN are unknown in black-box attack; while the architecture of the target GNN and hyperparameters during training may be known. In this setting, attackers are generally allowed to query the target GNN model to get the prediction vectors or embeddings of the queried samples. Similar to the white-box attacks, shadow datasets and the target nodes/graphs are also required to conduct black-box attacks. A practical example of the black-box privacy attack is to attack the API service that sends the output of the GNN models when receiving queries from the users.
\end{itemize}
The categorization of existing privacy attack methods according to the assumption on attacker's knowledge is shown in Table~\ref{tab:privacy_know}.

\subsection{Methods of Privacy Attack on GNNs} \label{sec:atk_methods}

\begin{figure}[t]
\centering
\begin{subfigure}{0.49\linewidth}
    \centering
    \includegraphics[width=0.9\linewidth]{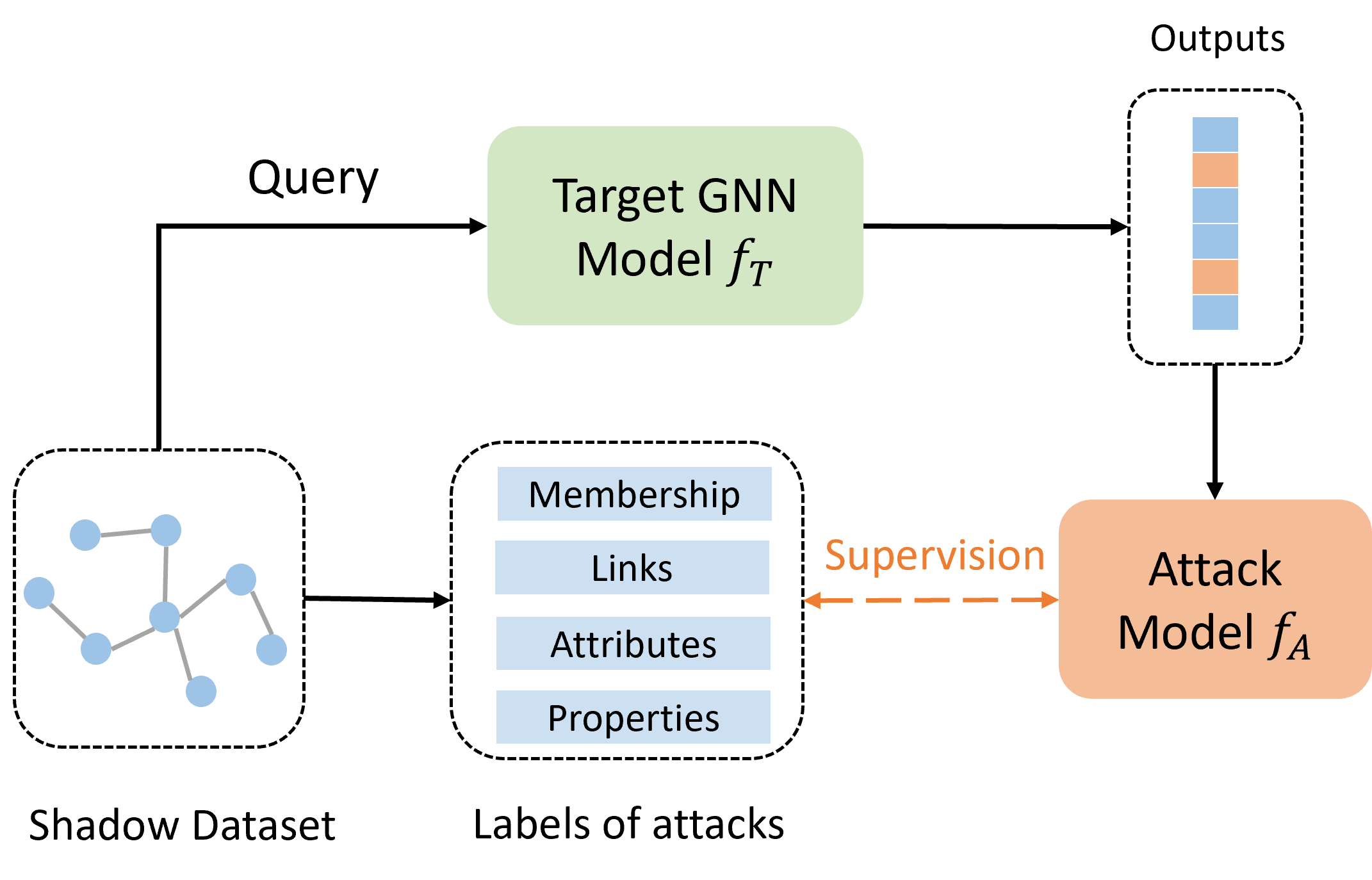}
    \vskip -1em
    \caption{Unified framework of privacy attacks.}
    \label{fig:privacy_unify}
\end{subfigure}
\begin{subfigure}{0.49\linewidth}
    \centering
    \includegraphics[width=0.92\linewidth]{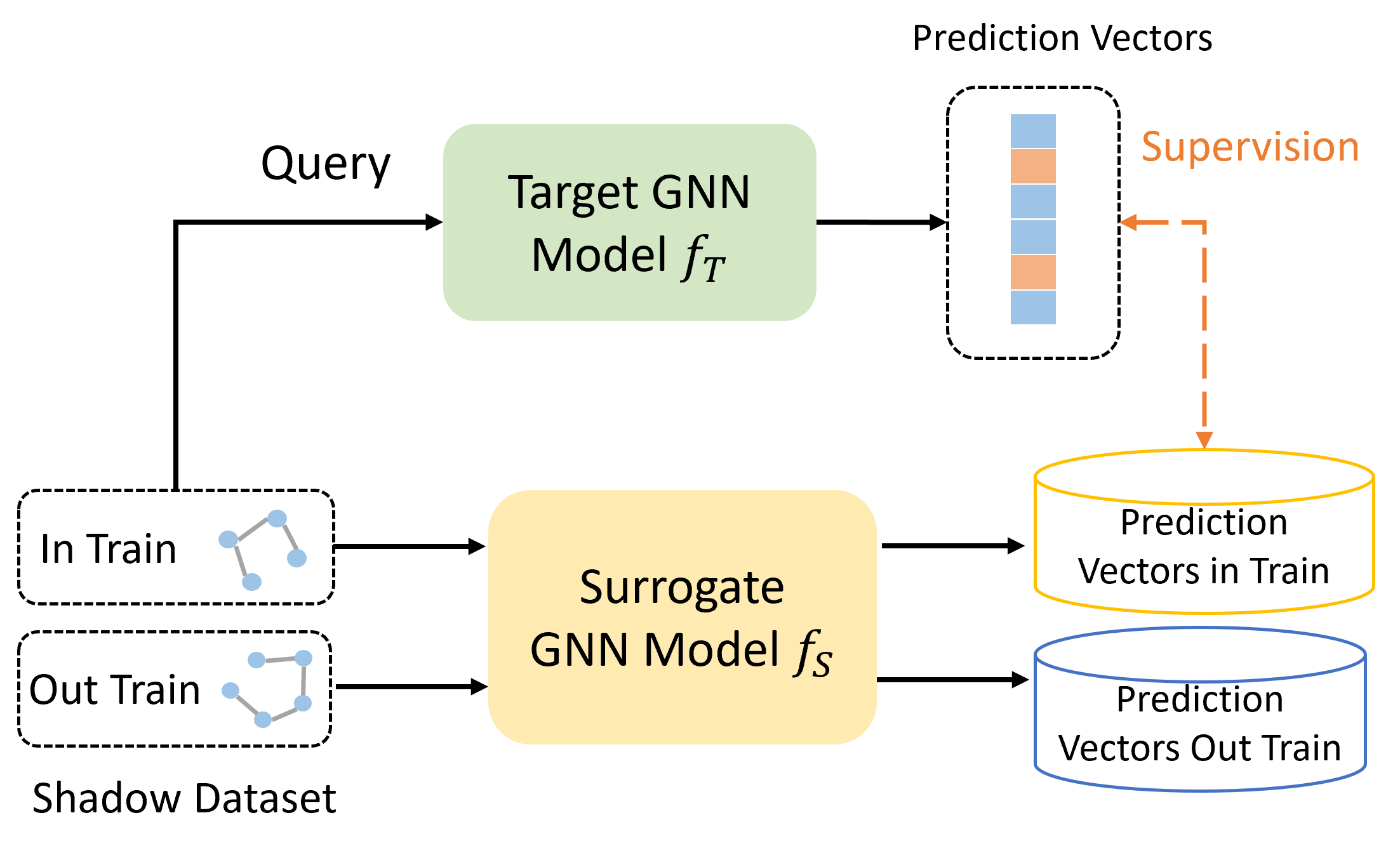}
    \vskip -1em
    \caption{Shadow training for membership inference.}
    \label{fig:privacy_shadow}
\end{subfigure}
\vskip -1em
\caption{The illustration of the privacy attack methods.}
\label{fig:privacy_attack}
\vskip -1em
\end{figure}

\vspace{0.2em} \noindent \textbf{The Unified Framework.}
Supervised privacy attack is a common design strategy of privacy attacks~\cite{duddu2020quantifying, olatunji2021membership,he2021node,wu2021adapting,zhang2021inference,he2021stealing,wu2020model,shen2021model}. The core idea of supervised privacy attack methods is to utilize the shadow dataset and the output of the target models to get the supervision for training a privacy attack mode, which can be described as a unified framework shown in Figure~\ref{fig:privacy_unify}. As shown in Figure~\ref{fig:privacy_unify}, the attacker uses a shadow dataset $\mathcal{D}_{S}$ as input to the target model $f_T$ to obtain the predictions or embeddings. Then, ground-truth of various types of privacy attacks on the shadow dataset can be attained. With the attack labels from the shadow dataset, attackers can train an attack model that performs inference based on the outputs of the target models by:
\begin{equation} \small
    \min_{\theta_A} \frac{1}{|\mathcal{D}_{S}|}\sum_{\mathcal{G}_i \in \mathcal{D}_{S}} l(f_A(f_T(\mathcal{G}_i)), y_i),
    \label{eq:privacy_attack}
\end{equation}
where $\mathcal{G}_i$ indicates the samples from the shadow dataset, which can be subgraphs of node $v_i$'s local graph for node classification or a sample graph for graph classification. $y_i$ represents the extracted attack labels of the sample $\mathcal{G}_i$. As illustrated in Fig.~\ref{fig:privacy_unify}, it can be varied from attributes to network properties for different privacy attacks. $l(\cdot)$ denotes the loss function such as cross entropy loss to train the attack model $f_A$. $\theta_A$ denotes the parameters of the attack model $f_A$. After the attack model is trained, privacy attack on the target example $\mathcal{G}_t$ can be conducted by $f_A(f_T(\mathcal{G}_t))$. Next, we will give more details about the methods of each types of privacy attacks.

\vspace{0.2em} \noindent \textbf{Membership Inference Attack.} The membership inference attack aims to identify if a target sample was used for training the target model $f_T$. The privacy leakage of membership is caused by the overfitting of the model on the training dataset, which leads to the prediction vectors (predicted label distributions) of training and test dataset follow different distributions~\cite{shokri2017membership,wu2021adapting}. Thus, an attacker can utilize the prediction vector to judge if a data instance was in the training set for $f_T$. To learn an membership inference attack model, the most common way is to apply shadow training to obtain supervision for membership inference and train an attack model. The process of shadow training is shown in Fig.~\ref{fig:privacy_shadow}. In shadow training, part of the shadow dataset $\mathcal{D}_S^{train}$ are used to train a surrogate model $f_{S}$ to mimic the behaviors of the target model as:
\begin{equation}\small
    \min_{\theta_S} \frac{1}{|\mathcal{D}_{S}^{train}|} \sum_{\mathcal{G}_i \in \mathcal{D}_{S}^{train}} l(f_{S}(\mathcal{G}_i), f_T(\mathcal{G}_i)),
    \label{eq:shadow_training}
\end{equation}
where $\mathcal{G}_i$ is a graph for graph classification or $k$-hop subgraph centered at node $v_i$ for node classification.  $f_{T}(\mathcal{G}_i)$ denotes the predicted label distribution of $\mathcal{G}_i$ and $l(\cdot)$ is loss function such as cross entropy loss to ensure $f_{S}(\mathcal{G}_i)$ is similar to $f_T(\mathcal{G}_i)$. Since $\mathcal{G}_i \in \mathcal{D}_S^{train}$ is used for training $f_S$, the predictive probability vectors of $\mathcal{G}_i$ from $f_{S}$, i.e., $[f_S(\mathcal{G}_i), y_i]$, can be labeled as positive, where $y_i$ is the ground-truth class label of $\mathcal{G}_i$, used to help the attack model to judge if $f_S(\mathcal{G}_i)$ is overfitting. Similarly, the probability vectors of other shadow samples $\mathcal{D}_{S}^{Out}=\mathcal{D}_S - \mathcal{D}_S^{train}$ can be labeled as negative samples for membership inference attack.
Then, the prediction vectors and the obtained labels for membership inference will be used to train an attackers model such as logistic regression, which can infer whether a sample is in the training dataset of the target model or not.

Membership inference attack has already been extensively investigated on i.i.d data~\cite{shokri2017membership,rahman2018membership,jayaraman2019evaluating}. Due to the great success of GNNs, recently, membership inference attack on graphs has raised increasing attention~\cite{olatunji2021membership,he2021node,duddu2020quantifying,wu2021adapting}, which generally follow the same scheme of shadow training. 
%Due to the utilization of graph topology, the shadow graph dataset and the target samples are largely different from that of i.i.d data, which can challenge the previous privacy attacks. 
In~\cite{olatunji2021membership}, membership inference based on subgraphs of the target node's local graph is investigated for node classification task. More specifically, the shadow dataset $\mathcal{D}_S$ in~\cite{olatunji2021membership} is a graph that comes from the same underlying distribution as the graph used for training. 
A GCN is deployed as the shadow model $f_S$. 
In~\cite{he2021node}, the authors further investigate the attack on target samples that are only provided with node features. To infer the membership of a singe test node, the attacker model is trained to distinguish $\{f_S(\mathbf{x}_i): v_i \in \mathcal{D}_{S}^{train}\}$ and $\{f_S(\mathbf{x}_i): v_i \in \mathcal{D}_{S}^{Out}\}$, where $f_S(\mathbf{x}_i)$ denotes the prediction of $f_S$ when only attributes of a single node $v_i$ is feed into the surrogate GNN.
In~\cite{wu2021adapting}, the authors also adopt the introduced framework. The major difference is that they achieve the membership inference attack on graph classification task.

\vspace{0.2em} \noindent \textbf{Reconstruction Attack.} As mentioned in Sec. \ref{sec:privacy_attack_types}, reconstruction attack aims to infer the sensitive attributes or/and links in target datasets. Due to the message passing of GNNs, the learned node/graph embeddings capture both node attributes and graph structure information.  
Thus, existing reconstruction attack method~\cite{duddu2020quantifying} reconstructs the information from node embeddings $\mathbf{H}=[\mathbf{h}_1,\dots, \mathbf{h}_N]$ learned by the target GNN. 
For attribute reconstruction, $f_A$ can simply be a multilayer perceptron (MLP) and reconstruct the attributes as $\mathbf{\hat X} = MLP(\mathbf{H})$.
For link inference, $f_A$ generally predicts the link based on the embeddings of node $v_i$ and $v_j$ by $w(i,j)=MLP(\mathbf{h}_i, \mathbf{h}_j)$. Following the unified framework, a shadow graph $\mathcal{G}_S$ is used to provide the adjacency matrix $\mathbf{A}_S$ and sensitive attributes $\mathbf{X}_S$ as supervision. The node embeddings $\mathbf{H}_S$ of the shadow graph are assumed to be available. Then, the attack model can be trained in a supervised manner. For attribute reconstruction attack, the training loss is given as $\min_{\theta_A} \sum_{v_i \in \mathcal{G}_S} l_{attr}(\mathbf{x}_i, \mathbf{\hat{x}}_i)$, where $l_{attr}$ could be MSE loss for continuous attributes and cross entropy loss for categorical attributes. As for the link inference, the objective function is: $\min_{\theta_A} \|\mathbf{\hat{A}}_S - \mathbf{A}_S\|_F^2$, where $\mathbf{\hat{A}}_S$ is the adjacency matrix reconstruted by $f_A$. Similar to link prediction, negative sampling may also be applied here. Reconstructing adjacency matrix on graph embeddings are also investigated in~\cite{zhang2021inference}. The main difference between~\cite{duddu2020quantifying} is that the attacker model directly infers the adjacency matrix with the graph embedding  by $\mathbf{\hat A} = MLP(\mathbf{h}^G)$, where $\mathbf{h}^G$ denotes a graph embedding.

% \suhang{notations and equations. Note that the unified framework provides a high-level. For each attack methods, we need to give specific questions to help explain. Otherwise, it is unclear}

For some applications, the embeddings might not be available while the prediction vector of an instance can be obtained by querying the target model. Therefore, He \textit{et al.}~\cite{he2021stealing} propose to infer the link of two nodes with their prediction vectors as $w(i,j) = MLP(\mathbf{y}_i, \mathbf{y}_j)$, where $\mathbf{y}_i$ is the prediction vector of node $v_i$ from target model $f_T$. Following shadow training, the authors firstly query the prediction vectors of shadow dataset from the target model $f_T$. Then, a surrogate model is trained on the shadow dataset and queried prediction vectors. Finally, the link inference attacker can be learned with the supervision generated from the surrogate model. 

The aforementioned methods all focus on black-box settings. However, with the development of pretraining GNN and federated learning that share the model parameters, white-box reconstruction attack methods  also start to attracting attentions. For example,  GraphMI~\cite{zhang2021graphmi} proposes to reconstruct the adjacency matrix in a white-box setting where the trained model parameters are known. Intuitively, the reconstructed adjacency matrix $\mathbf{A}$ will be similar to the original adjacency matrix if the loss between true node label $y_i$ and predictions using the reconstructed matrix, $f_{\theta}(\mathbf{X},\mathbf{X})_i$  is minimized. In addition, the graph structure is updated to ensure accurate predictions from GNN model under the feature smoothness constraint. Formally, GraphMI aims to solve the following optimization problem:
\begin{equation} \small
    \min_{\mathbf{A} \in \{0,1\}^{N \times N}} \frac{1}{|\mathcal{V}_L|} \sum_{v_i \in \mathcal{V}_L} l(f_\theta(\mathbf{A},\mathbf{X})_i, y_i) + \alpha \text{tr}(\mathbf{X}^T \mathbf{L} \mathbf{X}) + \beta \|\mathbf{A}\|_1
\end{equation}
where $\mathbf{L}=\mathbf{D}-\mathbf{A}$ is the Laplacian matrix of $\mathbf{A}$ and $\mathbf{D}$ is the diagonal matrix of $\mathbf{A}$. $\|\mathbf{A}\|_1$ is the $\ell_1$-norm to make $\mathbf{A}$ sparse. $\alpha$ and $\beta$ control the contributions of feature smoothness constraint and sparsity constraint on the adjacency matrix $\mathbf{A}$.

\vspace{0.2em} \noindent \textbf{Property Inference Attack.} The property inference attack is still in an early stage. An initial effort is taken in~\cite{zhang2021inference} to infer the properties of a target graph by its embedding. The proposed method also follows the unified framework. %The properties of shadow datasets will label the graph embeddings to train the attack model.  
Let $p_i$ denote the property of a graph $\mathcal{G}_i$ in the shadow dataset $\mathcal{D}_S$, the attack model is trained by $\min_{\theta_A} \frac{1}{|\mathcal{D}_S|} \sum_{\mathcal{G}_i \in \mathcal{D}_S} l(f_A(\mathbf{h}^G_i), p_i)$, where $\mathbf{h}^G_i=f_T(\mathcal{G}_i)$ is the embedding of $\mathcal{G}_i$. $l(\cdot)$ can be MSE loss or cross entropy loss for different types of properties.

\vspace{0.2em} \noindent \textbf{Model Extraction Attack}: The model extraction attack aims to learn a surrogate model that behaves similarly to the target model. The process of training a surrogate model is also included in the membership inference attack, which is shown in Fig.~\ref{fig:privacy_shadow}. Generally, the attacker will first query the target model to obtain predictions on the shadow dataset. It then leverages the shadow dataset and the corresponding predictions to train the surrogate model for model extraction attack, which has been formulated in Eq.(\ref{eq:shadow_training}). Following this framework, recent works~\cite{wu2020model,shen2021model} have investigated the GNN model extraction attacks with different levels of knowledge on shadow dataset and training graph of the target model. For example,  the general framework of model extraction is applied for training on the shadow dataset that provided with graph structures~\cite{wu2020model,shen2021model}. If no structure is given in the shadow dataset, the missing graph structures can be firstly learned. For example, in~\cite{shen2021model}, the graph structure is firstly initialized by KNN on the node attributes then  updated by a graph structure learning framework~\cite{chen2020iterative}.

\begin{table}[t]
    % \small
    \scriptsize	
    \centering
    \caption{The categorization of privacy-preserving graph neural networks.}
    \vskip -1.5em
    \begin{tabular}{lll}
        \toprule
         Category& Privacy Attacks to Defense & References   \\
         \midrule
         Differential Privacy (DP) & Membership Inference  & 
        \cite{olatunji2021releasing}, ~\cite{sajadmanesh2020locally},~\cite{xu2018dpne},~\cite{zhang2021graph} \\
         \midrule
         Federated Learning & - & \cite{he2021spreadgnn},~\cite{pei2021decentralized},~\cite{xie2021federated},~\cite{wang2020graphfl},~\cite{zheng2021asfgnn} \\
         Federated Learning + DP & Membership Inference &  \cite{wu2021fedgnn},~\cite{zhou2020vertically},~\cite{liu2021federated} \\
         \midrule
         Machine Unlearning & - & \cite{chen2022graph}, ~\cite{wu2023certified}, ~\cite{wu2023gif}, ~\cite{chien2022efficient}\\
         \midrule
         \multirow{2}{*}{Adversarial Privacy-Preserving} & Attribute Reconstruction & \cite{li2020adversarial},~\cite{liao2021information} \\
         & Structure Reconstruction & \cite{wang2021privacy}\\
         \midrule
         Model Ownership Verification & Model Extraction & \cite{xu2023watermarking},~\cite{zhao2021watermarking},~\cite{waheed2023grove} \\
         \midrule
          Other privacy protection methods & Membership Inference & \cite{dai2023unified}~\cite{olatunji2021membership}\\
         \bottomrule
    \end{tabular}
    \label{tab:pri_GNN}
\end{table}

\subsection{Privacy-Preserving Graph Neural Networks}
% \enyan{have a table to show which it is designed for}
As GNNs are vulnerable to privacy attacks and may leak the private information of users, privacy-preserving GNNs are developed to protect the privacy. Current privacy-preserving GNNs generally fall into the following categories, i.e., \textit{differential privacy}, \textit{federated learning}, \textit{machine unlearning}, \textit{adversarial privacy-preserving} and \textit{model ownership verification}. The categorization of the privacy-preserving GNNs are listed in Table~\ref{tab:pri_GNN}. Next, we will introduce  representative and state-of-the-art methods in each category. 
% \subsubsection{Against Membership Attack:} 
\subsubsection{Differential Privacy for Privacy-Preserving GNNs}
Differential Privacy (DP)~\cite{dwork2006calibrating} is a popular approach that can provide privacy guarantee of training data. The core idea of differential privacy is that that if two datasets differ only by one record and are used by the same algorithm, the outputs of the algorithm on the two datasets should be similar. With differential privacy, the impact of a single sample is strictly controlled. Thus, the membership inference attack can be defensed by DP with theoretical guarantee. Formally, differential privacy is defined as follows:
\begin{definition} [($\epsilon$, $\delta$)-Differential Privacy~\cite{dwork2006calibrating}] Given $\epsilon>0$ and $\delta \geq 0$, a randomized mechanism $\mathcal{M}$ satisfies $(\epsilon, \delta)$ differential privacy, if for any adjacent datasets $D \text{~and~} D' \in \mathcal{R}$ and for any subsets of outputs $\mathcal{S}$, the following equation is met:
\begin{equation}
    %\max_{a} 
    {P(\mathcal{M}(D) \in \mathcal{S})}\leq e^{\epsilon} P(\mathcal{M}(D') \in \mathcal{S}) + \delta,
\end{equation}
where $\epsilon$ is the privacy budget to trade-off the utility and privacy. A larger $\epsilon$ will lead to stronger privacy guarantee but weaker utility. When $\delta=0$, it is equivalent to $\epsilon$-Differential Privacy. ($\epsilon$, $\delta$)-DP allows for the possibility that plain $\epsilon$-DP is broken with a small probability $\delta$. 
\end{definition}
To achieve ($\epsilon$, $\delta$)-Differential Privacy, some additive noise mechanisms such as Gaussian mechanism~\cite{dwork2014algorithmic} and Laplace mechanism~\cite{dwork2006calibrating} are widely adopted. Based on the privacy budget and the mechanism to be protected, certain levels of Gaussian noise or Laplace noise will be injected to achieve a differentially private mechanism.  Recently, various differential-privacy preserving deep learning methods~\cite{sajadmanesh2020locally,abadi2016deep,
arachchige2019local,papernot2016semi} are proposed to protect the training data privacy. For instance, NoisySGD~\cite{abadi2016deep} adds noises to the gradients during model training so the trained model parameters will not leak training data with certain guarantee.
PATE~\cite{papernot2016semi} firstly trains an ensemble of teacher models on  subsets of sensitive training data that are split disjointly. Then, the student model will be trained with the aggregated output of the ensemble on public data. As a result, the student model is unlikely to be affected by a change of a single sensitive data, which meets the differential privacy. Theoretical guarantee of PATE in privacy is also analyzed in~\cite{papernot2016semi}.
In differential privacy, a trusted curator will be required to apply calibrated noise to produce DP. To handle the situation of untrusted curator, local differential privacy methods~\cite{arachchige2019local,sajadmanesh2020locally} that perturbs users' data locally before uploading to the central server are also investigated for privacy protection. 

% Various DP-based privacy-preserving deep learning methods~\cite{sajadmanesh2020locally,abadi2016deep,
% arachchige2019local,papernot2016semi,zhu2020private} have been.

To protect the privacy of graph-structured data, many differential privacy preserving network embedding methods are investigated~\cite{olatunji2021releasing,sajadmanesh2020locally,xu2018dpne,zhang2021graph}. For instance, DPNE~\cite{xu2018dpne} applies perturbations on the objective function of learning network embeddings. In~\cite{zhang2019graph}, a perturbed gradient descent method that guarantees the privacy of graph embeddings learned by matrix factorization is proposed. More recently, several works~\cite{sajadmanesh2020locally,olatunji2021releasing} that focus on differentially private GNNs are explored. For example, the locally private GNN~\cite{sajadmanesh2020locally} adopts local differential privacy~\cite{kasiviswanathan2011can} to protect the privacy of node features by perturbing the user's features locally. Furthermore, a robust graph convolution layer is investigated to reduce the negative effects of the injected noises.  PrivGnn~\cite{olatunji2021releasing} extends the Private Aggregation of Teacher Ensembles (PATE)~\cite{papernot2016semi} for graph-structured data to release GNNs with differential privacy guarantees. In particularly, random subsampling on the training set of teacher and noisy labeling mechanism on the public data are used in PrivGNN~\cite{olatunji2021releasing} to achieve practical privacy guarantees. 
As for~\cite{zhang2021graph}, both the user feature perturbation at input stage and loss perturbation at optimization stage are investigated to achieve a privacy-preserving GNN for recommendation. In addition, they also empirically show that their proposed method can help defend against the attribute inference attack as the noises added to node features can prevent the attacker from reconstructing the original sensitive attribute. 

\subsubsection{Other Methods for Membership Privacy} Beyond differential privacy approaches, recent developments have given rise to a variety of other methods aimed at preserving membership privacy. Next, we will delve into a detailed exploration of these newly established works.

\vspace{0.2em}
\noindent \textbf{LBP} and \textbf{NSD}~\cite{olatunji2021membership}: These two methods are preliminary explorations in defending against membership inference attacks. Specifically, LBP is an output perturbation method by where noise is infused to the posterior before it is released to end users. Intuitively, noise can obfuscate the posteriors, making it challenging to discern between member and non-member node posteriors. As for NSD, it randomly chooses neighbors of the queried node during inference. This can limit the amount of information used in the target model, thereby protecting membership privacy.

\vspace{0.2em}
\noindent \textbf{RM-GIB}~\cite{dai2023unified}. This work represents a pioneering effort to craft a unified graph neural network framework designed to both preserve membership privacy and defend against adversarial attacks. In this paper, \textit{Dai et al.}~\cite{dai2023unified} first connect the Information Bottleneck (IB) with the membership privacy preservation. 
In particularly, the proposed RM-GIB principle can be written as:
\begin{equation}
    \min_{\theta} -I(\mathbf{z}_x, \mathcal{N}_S;y) + \beta I(\mathbf{z}_x, \mathcal{N}_S; \mathbf{x}, \mathcal{N})
    \label{eq:GIB},
\end{equation}
where $\mathbf{z}_x$ represents the attribute bottleneck encoding the node attribute information,  $\mathcal{N}_S$ is a subset of $v$'s neighbors that bottleneck the neighborhood information, and $I(;)$ denotes the mutual information.
As analyzed in~\cite{dai2023unified}, the regularization in IB (latter term in Eq.(\ref{eq:GIB})) can constrain the mutual information between representations and labels on the training set. This constraint can narrow the gap between training and test sets to avoid membership privacy leakage. In addition, RM-GIB collects pseudo labels
on unlabeled nodes and integrate them with the given labels in the optimization, further benefiting the membership privacy.
The proposed RM-GIB also benefit the robustness, as an attribute bottleneck and a
neighbor bottleneck are deployed to remove the redundant information and/or adversarial perturbations in both node attributes and graph
topology.

\begin{figure}
    \centering
    \includegraphics[width=0.62\linewidth]{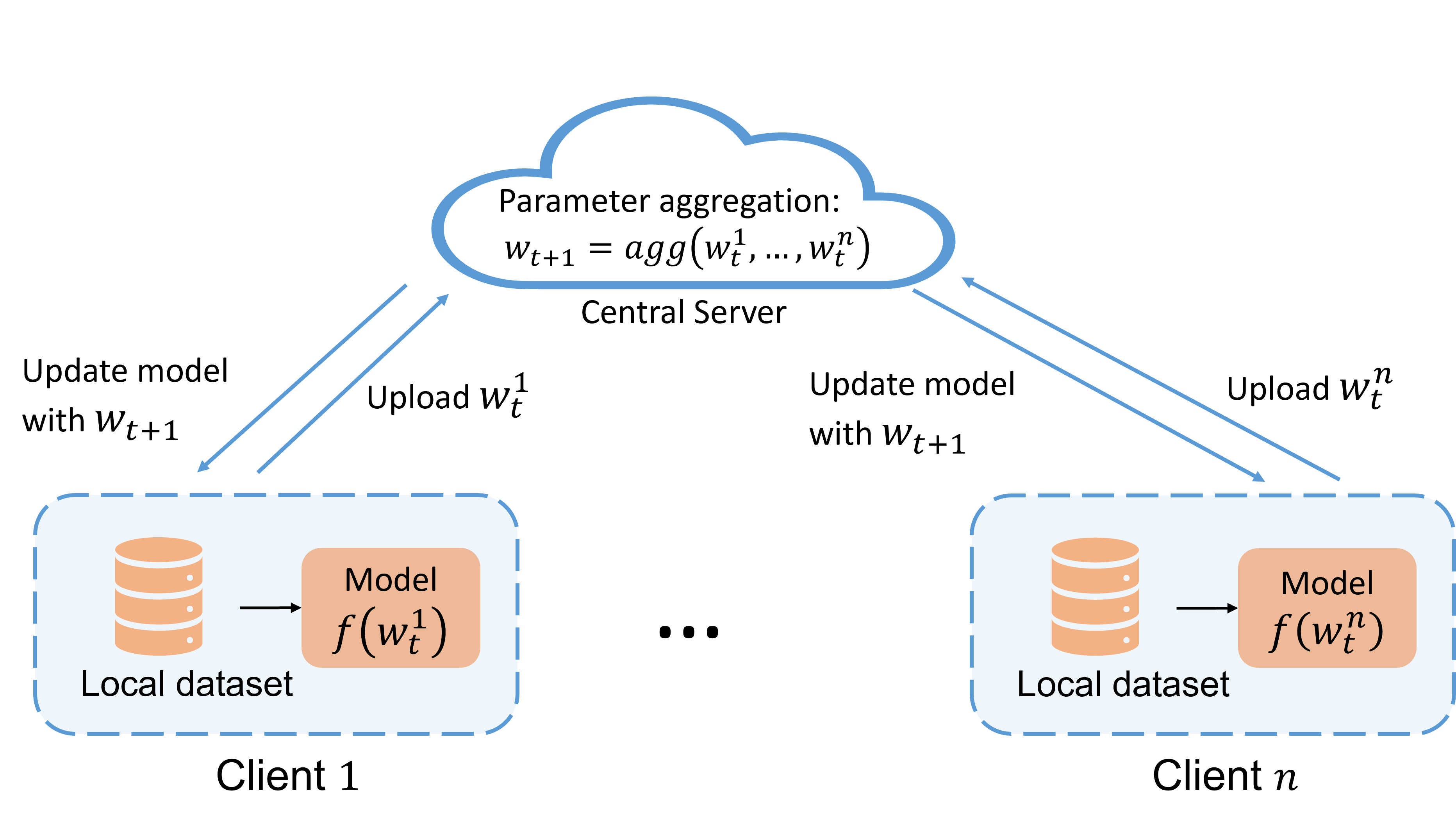}
    \vskip -1em
    \caption{The illustration of the federated learning.}
    \label{fig:federated}
    \vskip -1em
\end{figure}

\subsubsection{Federated Learning for Privacy-Preserving GNNs} 
Currently, the majority of deep learning methods require a centralized storage of user data for training. However, this can be unrealistic due to the privacy issue. For example, when several companies or hospitals want to combine their data to train a GNN model, the data of their users are not allowed to be shared according to the privacy terms. Furthermore, the users may be not willing to upload their data to the platform server due to the concern of information leakage. For such situations, the data will remain in the local devices of users or the data holder organizations. To address this problem, federated learning~\cite{konevcny2016federated,mcmahan2017communication} is proposed to collectively learn models with decentralized user data in a privacy-preserving manner. In particular, it aims to optimize the following objective function:
\begin{equation} \small
    \min_w \sum_{k=1}^n p_k \mathcal{L}_k(\mathcal{D}_{k},w),
    \label{eq:federated}
\end{equation}
where $n$ is the total number of devices/clients,  $\mathcal{D}_k$ is the local dataset stored in the $k$-th client, and $\mathcal{L}_k$ is the local objective function for the $k$-th device. The impact of each device is controlled by $p_k$ with $p_k \ge 0$ and $\sum_k p_k = 1$. The $p_k$ is often set as $\frac{1}{n}$ or $\frac{|\mathcal{D}_k|}{|\mathcal{D}|}$, where $|\mathcal{D}|=\sum_k |\mathcal{D}_k|$ is the total size of samples. 
A general framework of federated learning to solve Eq.(\ref{eq:federated}) is given in Fig.~\ref{fig:federated}, where 
the user data and a local model $f_k(w_t^k)$ are maintained locally in  $k$-th client in federated learning. In the training step $t$, each client will compute the local model updates based on their own data. Then, the central server will aggregate the model updates from clients and update the global model parameters to $w_{t+1}$. The updated global model will be distributed to the clients for future iterations. 
The federated learning framework is firstly proposed in FedAvg~\cite{mcmahan2017communication}, which is the most commonly used federated learning algorithm now. specifically, FedAvg aggregates the model parameters by averaging the updated model parameters from clients as $w_{t+1} = \sum_{k=1}^n \frac{|\mathcal{D}_k|}{|\mathcal{D}|} w_t^k,$, 
% \begin{equation}
%     w_{t+1} = \sum_{k=1}^n \frac{|\mathcal{D}_k|}{|\mathcal{D}|} w_t^k,
% \end{equation}
where $w_t^k$ denotes the updated parameters in $k$-th client at step $t$. More comprehensive survey about federated machine learning can be found in~\cite{yang2019federated}.

To protect the user privacy, federated learning has been extended to train GNNs~\cite{wang2020graphfl,pei2021decentralized,he2021spreadgnn,xie2021federated,wu2021fedgnn,liu2021federated,zhou2020vertically,zheng2021asfgnn}. To handle the challenges caused by non-i.i.d. graphs, Xie \textit{et al.}~\cite{xie2021federated} propose to dynamically cluster local systems based on GNN gradients, which can reduce the structure and feature heterogeneity among graphs. In~\cite{wang2020graphfl}, the work proposes a hybrid of  meta learning and federated framework. They view the training on a client as a task in meta-learning and learn a global model to mitigate the issue of non-i.i.d data. Then, federated learning methods are leveraged to further update the global model.
In~\cite{zhou2020vertically}, vertically federated learning that assumes that different clients hold different features and neighborhood information of the user data is proposed. 
To protect the private data, i.e., node attributes, edges and labels, the computations on private data are carried out by the data-holders. A semi-honest server focuses on computations on encoded node representations that are non-private.
Apart from federated learning GNNs in node/graph classification, federated frameworks on GNNs for recommendation are also developed~\cite{wu2021fedgnn,liu2021federated}. FedGNN~\cite{wu2021fedgnn} incorporates the high-order user-item interactions by building the local user-item graphs in a privacy-preserving way. Furthermore, noises are injected locally in federated learning to meet the local differential privacy for privacy projection. FeSoG~\cite{liu2021federated} further extends the federated GNNs for social recommendation which involves  social information for predictions.

Moreover, decentralized federated learning for GNNs 
%that does not require the central server for aggregation 
is also explored in~\cite{pei2021decentralized,he2021spreadgnn}, where a client can communicate with a set of neighbor clients for aggregation without the central server. Therefore, existing algorithms~\cite{pei2021decentralized,he2021spreadgnn} propose to aggregate the local model with neighbor local models. For example, decentralized periodic averaging SGD~\cite{he2021spreadgnn} applies SGD on each client locally and synchronizes parameters with their neighbors every certain number of iterations.

\subsubsection{Machine Unlearning} One of very important regulations on privacy is to ensure the right to be forgotten~\cite{smuha2019ethics}. This principle demands not only the deletion of data from storage but also the removal of related information from trained AI models. 
Retraining can be a possible solution for training data removal. However, this can be overwhelmingly expensive especially for models trained on large-scale datasets. Therefore, machine unlearning methods~\cite{} have been developed to efficiently erase the user information in deep models including graph neural networks. Based on the targets, machine unlearning can be split into \textit{exact unlearning} and \textit{Approximate unlearning}.

\begin{definition}[Exact unlearning~\cite{bourtoule2021machine}] Given a learning algorithm $A(\cdot)$, a dataset $\mathcal{D}$, and a forget set $\mathcal{D}_f \subset \mathcal{D}$, the exact unlearning process $U(\cdot)$ is required to meet:
\begin{equation}
    P(A(\mathcal{D}\backslash \mathcal{D}_f)) = P(U(\mathcal{D},\mathcal{D}_f, A(\mathcal{D}))),
\end{equation}
where $P(A(\mathcal{D}))$ denote the distributions of all models trained on $\mathcal{D}$ by the learning algorithm $A(\cdot)$.
\end{definition}
% It is challenging to meet the criteria for exact unlearning.
SISA (Sharded, Isolated, Sliced and Aggregated)~\cite{bourtoule2021machine} is one popular framework for exact unlearning. In this approach, the training set $\mathcal{D}$ will be first partitioned into $K$ disjointed shards. Subsequently, $K$ independent models are trained with the $K$ shards. Predictions are given by assembling the outputs from the these models. When a data point $x$ requests removal, only the model using the shard containing $x$ will be retrained on the small shard. For the graph-structed data, the partition will destroy the training graph’s structure which largely degrade the utility. Therefor, GraphEraser~\cite{chen2022graph} introduces a balanced graph partition which split graphs based on community/clusters. 

\begin{definition}[($\epsilon, \delta$)-Approximate Unlearning~\cite{guo2019certified}] Given $\epsilon>0$ and $\delta \geq 0$, a unlearn process $\mathcal{M}$ satisfies $(\epsilon, \delta)$ approximate unlearning, if for any data point $z \in \mathcal{D}$ and model sets $\mathcal{T}$, the following equation is met:
\begin{equation}
    %\max_{a} 
    {P(U(\mathcal{D},\mathcal{D}_f, A(\mathcal{D})) \in \mathcal{T})}\leq e^{\epsilon} P(A(\mathcal{D}\backslash \mathcal{D}_f) \in \mathcal{T}) + \delta,
\end{equation}
\end{definition}
\noindent where $\epsilon$ is the budget to trade-off the utility and privacy. A larger $\epsilon$ will lead to stronger privacy guarantee but weaker utility. ($\epsilon$, $\delta$)-approximate unlearning allows that plain $\epsilon$-approximate unlearning is broken with a small probability $\delta$. Existing methods~\cite{wu2023certified,wu2023gif,chien2022efficient} mainly adopt the influenced function-based parameter updating for approximate unlearning by: $\theta^{-} = \theta + H_{\theta}^{-1} \Delta$, 
where $H_{\theta}^{-1} \Delta$ is the influence function of the training point to be unlearned.
This is feasible because influence functions quantify how model parameters will adjust if the loss of a particular data point is excluded from training. However, merely adopting influence functions does not assure approximate unlearning because gradient residual persists after updating model parameters. Certain noises in the training  would be necessary  to ensure the gradient residual would not leak privacy. 

\subsubsection{Adversarial Privacy-Preserving GNNs}
To defend against the sensitive attributes/links leakage attack, adversarial learning is adopted for privacy-preserving GNNs~\cite{li2020adversarial,liao2021information,wang2021privacy}. Let $\mathbf{H}$ be node representations learned by the encoder/GNN as $\mathbf{H}=f_E(\mathcal{G};\theta)$. The core idea of the adversarial privacy-preserving is to adopt an adversary $f_A$ to infer sensitive attributes from node representations $\mathbf{H}$ while the encoder $f_E$ aims to learn representation that can fool $f_A$, i.e., making $f_A$ unable to infer sensitive attributes. It is theoretically shown in~\cite{wang2021privacy} that through this minmax game, the mutual information between learned representation and sensitive attribute, $MI(\mathbf{H},s)$, can be minimized, which protects the sensitive information from leakage. The process can be formally written as
\begin{equation} \small
    \min_{\theta_E} \max_{\theta_A} \mathcal{L}_{utility}(f_E(\mathcal{G};\theta_E)) - \beta \mathcal{L}_{Adversarial}(f_A(\mathbf{H};\theta_A)), 
\end{equation}
where $\theta_E$ and $\theta_A$ are parameters of encoder $f_E$ and adversary $f_A$, respectively. $\mathcal{L}_{utility}$ is the loss function to ensure the utility of the learned representations such as classification loss and reconstruction loss. $\mathcal{L}_{Adversarial}$ is the adversarial loss which generally is cross entropy loss of sensitive attribute prediction of the adversary based on node representations $\mathbf{H}$. $\beta$ is the hyperparameter to balance the contributions of these two loss terms. 

Adversarial privacy-preserving GNN is firstly proposed in~\cite{li2020adversarial} to defend against attribute inference attack, where link prediction loss and the  node attribute prediction loss are combined together for utility. Let $\mathbf{x}_v^c$ denotes the $c$-th attribute of node $v$ and $f_c(\mathbf{h}_v)$ denotes the prediction of $c$-th attribute based on the representation of node $v$. The utility loss can be written as
\begin{equation} \small
    \mathcal{L}_{utility} = \sum_{i=1}^N \sum_{j=1}^N l_{edge}(\mathbf{A}_{ij}, \mathbf{\hat A}_{ij}) + \alpha \sum_{v \in \mathcal{V}} \sum_{c \in \mathcal{C}} l_{attr}(\mathbf{x}_v^c, f_c(\mathbf{h}_v)),
\end{equation}
where $l_{edge}$ and $l_{attr}$ denote loss functions for link prediction and attribute reconstruction, $\mathcal{C}$ denotes the set of attribute to be reconstructed. As for the adversarial loss, it is $\mathcal{L}_{Adversarial}=\sum_{v\in \mathcal{V}_S} -[s_v \log(\hat{s}_v) + (1-s_v)\log (1-\hat{s}_v)]$, where $\hat{s}_v = f_A(\mathbf{h}_v)$ and $\mathcal{V}_S$ is the set of nodes with sensitive attributes.
GAL~\cite{liao2021information} deploys a WGAN-based adversarial privacy-preserving for information obfuscation of sensitive attributes. Both attribute privacy protection and link privacy protection with adversarial learning are discussed in~\cite{wang2021privacy}.

\begin{wrapfigure}{r}{0.38\linewidth}
    \vskip -1em
    \includegraphics[width=\linewidth]{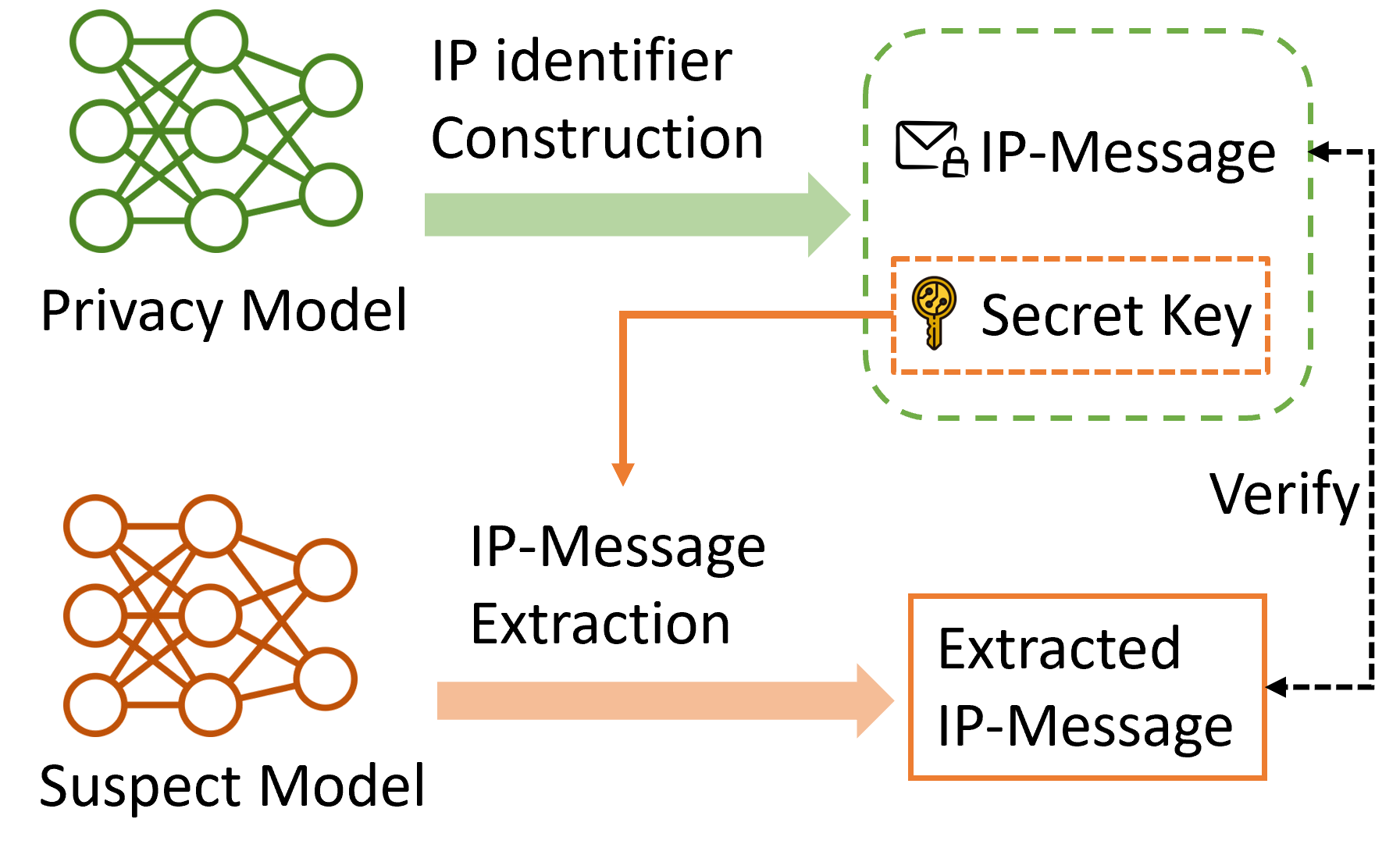}
    \vskip -1em
    \caption{An illustration of model ownership verification.} \label{fig:IP_pipline}
    \vskip -1.5em
\end{wrapfigure}
\subsubsection{Model Ownership Verification} Nowadays, it can be overwhelmingly expensive to train a high-performance model due to the demands in high-quality data collection and expensive computation cost. For example, the well-known ChatGPT is reported to cost around 12 million dollars to train once. Similarly, models for graph-structured data, such as graph contrastive learning on molecules~\cite{qiu2020gcc}, also tend to require massive computation on large-scale data. Therefore, trained DNNs is considered as deep Intellectual Property (IP) with high business values. It is necessary to protect the deep models from stealing and abusing by adversaries. Extensive model ownership verification methods have been proposed for deep neural networks for images and text~\cite{sun2023deep}. Recently, model ownership verification methods for graph neural networks are emerging~\cite{xu2023watermarking,zhao2021watermarking,waheed2023grove}. In this subsection, we will illustrate the general workflow of model ownership verification followed by recent advances in protecting GNNs. 

\vspace{0.2em}

\noindent \textbf{Workflow of Model Ownership Verification.} As it is shown in Fig.~\ref{fig:IP_pipline}, the workflow protecting the deep IP mainly consists of two stages, i.e., IP construction and IP verification. 
\begin{itemize}[leftmargin=*]
    \item In the IP construction phase, an IP identifier for the deep model will be built either in a invasive or non-invasive way. Generally, the IP identifier is in the form of \textit{key-message pair}. Model owners possess the secret keys, which can be predefined matrices or special input samples. The IP messages could be a bit string or model outputs that are triggered by the secret keys.
    \item In the verification phase, given the secret keys, we will testify whether the same IP identifier exists in the suspect model. Specifically, a stealing model will convey the same IP message. On the contrary, an independent model will not output the IP message.
\end{itemize} 
Based on whether the IP identifier is constructed in an invasive way, the model ownership verification for GNNs can be split into \textit{watermarking methods}~\cite{xu2023watermarking,zhao2021watermarking} and \textit{fingerprinting methods}~\cite{waheed2023grove}. 

\vspace{0.2em}
\noindent \textbf{GNN Watermarking}. It is an invasive solution that will embed the detectable and unforgettable IP identifier to the GNN model to obtain a watermarked GNN. Current methods~\cite{xu2023watermarking,zhao2021watermarking} focus on applying backdoor attacks to obtain watermarked GNN model. Specifically, backdoor attacks aim to learn a model that predicts the target class $y_t$ given arbitrary sample attached with the predefined trigger $\mathcal{G}_t$. Therefore, the trigger graph $\mathcal{G}_t$ can work as the secret key for watermarking. The target class $y_t$ will be the IP message. The training process in~\cite{xu2023watermarking,zhao2021watermarking} is the same as the graph backdoor attacks. Firstly, a trigger graph $\mathcal{G}_t$ will be generated. Then, samples with the defined trigger $\mathcal{G}_t$ will be labeled as a target class $y_t$ and join the training process of GNN. More details about the backdoor attacks are illustrated in Sec.~\ref{sec:backdoor}. As for watermarking GNN model without using backdoor attacks, it still remains further investigation. 

\vspace{0.2em}
\noindent \textbf{GNN Fingerprinting}. Fingerprinting is a noninvasive solution which aims to build IB identifier for a trained model. The main idea is based on the assumption that a trained neural network models will exhibit distinct characteristics when compared to an independently trained model. These differences can be model predictions, decision boundaries, adversarial samples, etc. Fingerprinting techniques have been deeply studied in the context of i.i.d data~\cite{sun2023deep}. However, for GNN fingerprinting, there is only one initial effort named GROVE~\cite{waheed2023grove}, which proposes a fingerprinting scheme based on GNN embeddings. Specifically, More precisely, a set of test samples, denoted as $\mathcal{D}_f$, serve as the secret key for fingerprinting. During the verification, the similarity between the embeddings from protected model and suspect model will be computed as the IP message. If the similarity of embeddings from the protected model and suspect model are high, it indicates a high probability of model stealing. 

\begin{table}[t!]
    % \small
    \scriptsize	
    \centering
    \caption{Datasets in Privacy}
    \vskip -1.5em
    \begin{tabular}{llllllll}
    \toprule
    Type & Dataset & \#Graphs & Avg.Nodes & Avg.Edges & \#Features & References\\
    \midrule
    &Cora & 1 & 2,708 & 5,429 & 1,433 & \makecell[l]{\cite{duddu2020quantifying} \cite{olatunji2021membership} \cite{he2021node} \cite{he2021stealing} \cite{zhang2021graphmi} \cite{wu2020model}}\\
    % \cmidrule{2-7}
    Citation & Citeseer & 1  & 3,312 & 4,715 & 3,703 & \makecell[l]{\cite{duddu2020quantifying} \cite{olatunji2021membership} \cite{he2021node} \cite{he2021stealing} \cite{zhang2021graphmi} \cite{wu2020model} \cite{shen2021model}}\\ 
    % \cmidrule{2-7}
    &PubMed & 1  & 19,717 & 44,338 & 500 & \makecell[l]{\cite{duddu2020quantifying} \cite{olatunji2021membership} \cite{he2021stealing} \cite{wu2020model} \cite{shen2021model}}\\
    % \cmidrule{2-7}
    &DBLP & 1  & 17,716 & 105,734 & 1,639 & \cite{shen2021model}\\
    \midrule
    
    Authorship & Coauthor & 1  & 34,493 & 247,962 & 8,415 & \cite{shen2021model}\\
    &ACM & 1  & 3,025 & 26,256 & 1,870 & \cite{shen2021model}\\
    \midrule
    
    &Facebook & 1  & 4,039 & 88,234 & - & \cite{duddu2020quantifying}\\ 
    Social Networks & LastFM & 1  & 7,624 & 27,806 & 7,842 & \cite{duddu2020quantifying} \cite{he2021node}\\
    &Reddit & 1  & 232,965 & 57,307,946 & 602 & \cite{olatunji2021membership}\\
    \midrule
    
    Image & Flickr & 1  & 89,250 & 449,878 & 500 & \cite{olatunji2021membership}\\
    \midrule
    
    &PROTEINS & 1,113  & 39.06 & 72.82 & 29 & \cite{wu2021adapting} \cite{he2021stealing}\\
    Bioinformatics & DD & 1,178  & 284.32 & 715.66 & 89 & \cite{wu2021adapting} \cite{zhang2021inference}\\
    &ENZYMES & 600  & 32.63 & 62.14 & 21 & \makecell[l]{\cite{wu2021adapting} \cite{zhang2021inference} \cite{he2021stealing} \cite{zhang2021graphmi}} \\
    \midrule
    
    &NCI1 & 4,110  & 29.87 & 32.30 & 37 & \cite{wu2021adapting} \cite{zhang2021inference}\\
    Molecule & AIDS & 2,000  & 15.69 & 16.20 & 42 & \cite{zhang2021inference} \cite{he2021stealing} \cite{zhang2021graphmi} \\
    &OVCAR-8H & 4,052  & 46.67 & 48.70 & 65 & \cite{wu2021adapting}  \cite{zhang2021inference}\\
    
    \bottomrule
    \end{tabular}
    \label{tab:privacy_datasets}
    \vskip -1.5em
\end{table}

% \subsubsection{Others}
\subsection{Datasets for Privacy-Preserving GNNs}
% \enyan{Junjie: add the description of datasets in the privacy papers as the fairness section and completed the related info as Table~\ref{tab:fair_datasets} do. }
In this subsection, we list the datasets that have been used in the literature about GNN's privacy.  The statistics of the datasets along with papers used the datasets are presented in Table~\ref{tab:privacy_datasets}. 
\begin{itemize}[leftmargin=*]
    \item \textbf{Cora, Citeseer, PubMed, DBLP} \cite{sen2008collective, pan2016tri}: Cora, Citeseer, PubMed, and DBLP are citation network datasets. Cora consists of seven classes of machine learning papers. CiteSeer has six classes. Papers are represented by nodes, while citations between two papers are represented by edges. Each node has features defined by the words that appears in the paper's abstract. Similarly, PubMed is a collection of abstracts from three types of medical papers. The data of DBLP comes from four research areas.
    \item \textbf{Facebook, LastFM} \cite{leskovec2012learning}: Facebook and LastFM are social network datasets. Different user accounts are represented by nodes that are connected by edges. Each user node in Facebook has features such as gender, education, hometown, and location. LastFM was collected through the following relationship in social network.
    \item \textbf{Flickr} \cite{mcauley2012image}: Flickr is an online photo management and sharing application. The edges reflect the common properties shared between two images, whereas the nodes represent an image submitted to Flickr. Users' node features are specified by their tags, which indicate their interests.
    \item \textbf{Reddit} \cite{hamilton2017inductive}: The Reddit dataset represents the post-to-post interactions of a user. An edge between two posts indicates that the same user commented on both posts. The labels correspond to the community that a post is associated with.
    \item \textbf{Coauthor} \cite{shchur2018pitfalls}: Coauthor is a dataset of co-authorship. Nodes represent authors, and edges connect two authors if they co-authored a paper. Features are collected by keywords in the author's papers. The label of a node is the area that the author focuses on.
    \item \textbf{ACM} \cite{wang2019heterogeneous}: In ACM, nodes are papers. Edges mean if there are same authors in two papers. Features of a node are keywords of the paper, while labels are the conferences that the papers published.
    \item \textbf{PROTEIN, DD, ENZYMES} \cite{morris2020tudataset}: DD, ENZYMES, and PROTEINS are macromolecules datasets. Nodes are secondary structural elements labeled with their type and a variety of physical and chemical data. If two nodes are neighbors along the amino acid sequence or one of the three nearest neighbors in space, an edge links them. ENZYMES assigns enzymes into six classes, which reflect the catalyzed chemical reaction. The labels in PROTEINS show whether a protein is an enzyme. In DD, nodes are amino acids, and edges are their spatial proximity.
    \item \textbf{AIDS, NCI1, OVCAR-8H, COX2} \cite{morris2020tudataset, zhang2021inference}: AIDS, NCI1, OVCAR-8H, and COX2 are molecule datasets. Atoms are represented by nodes, while chemical bonds are represented by edges. The node features consist of atom types. The label is decided by toxicity or biological activity in drug discovery projects.

\end{itemize}

\subsection{Applications of Privacy Preserving GNNs}

\vspace{0.2em} \noindent \textbf{Pretraining and Model Sharing.} Nowadays, there is an increasing trend of pretraining models~\cite{qiu2020gcc,hu2020gpt} on large-scale datasets to benefit the downstream tasks. In practice, the pretrained model will often be shared to other parties for their use. However, the pretrained model itself has embedded the information of the training data, which can cause private data leakage by privacy attacks such as membership inference and attribute inference. The privacy-preserving GNNs can be applied to address this concern.  For instance, differential privacy-preserving GNNs~\cite{olatunji2021releasing,sajadmanesh2020locally,zhang2021graph} can be adopted in the pretraining phase to defend against the membership inference attack.  

\vspace{0.2em} \noindent \textbf{Distributed Learning.} Due to challenges of processing the large amount of data such as privacy concerns, computational cost, and memory capability, the demand of distributed learning is increasing dramatically.  In this situation, federated learning on GNNs provide solutions to distributed learning by processing  data in local devices. In addition, combining the other privacy-preserving methods such as differential privacy~\cite{wu2021fedgnn}, the client's data can be protected from privacy attacks.

\vspace{0.2em} \noindent \textbf{Healthcare.} 
Graph-structured data such as protein molecules, brain network, and patient network are pervasive in healthcare domain. 
GNNs have been trained on these private healthcare data for various applications. For example, GNNs have been used to process electronic health record (EHR) of patients for diagnosis prediction~\cite{li2020graph}. GNNs are also deployed to better capture the graph signal of brain activity for medical analysis~\cite{ahmedt2021graph}. To ensure the privacy of sensitive data of patients, privacy-preserving GNNs are required for the applications in healthcare domain.
% Privacy attack such as reconstruction attack may lead to the information leakage of the discovered structure that are not intended to be shared.  
% \vspace{0.2em} \noindent \textbf{Social Network Analysis.}
% With the development of social media, analysis based on the social network have been widely conducted. As it is aforementioned, private information can be leaked during the analysis process. For instance, the embeddings of users from social platforms such as Facebook may be released for academic research. However, adversary can utilize the embeddings to infer sensitive attribute information such as political tendency, gender, and races~\cite{li2020adversarial}. And the social network is often utilized for various analysis that contain certain protected groups. For instace, GNN may be applied to the social networks of COVID-19 patients to analyze the propagation of virus. In this situation, membership inference attack can identify the patients. Therefore, privacy-preserving GNNs is important for social network analysis. 

\vspace{0.2em} \noindent \textbf{Recommendation System.} GNNs are widely applied in recommendation system to involve social context and better utilization of high-order neighbor information~\cite{wu2021fedgnn}. Similarly, information can be leaked from the GNN-based recommendation system. To protect the user privacy, various privacy-preserving recommendation systems have been proposed~\cite{wu2021fedgnn,liu2021federated,zhang2021graph}. 

% \vspace{0.2em} \noindent \textbf{Biomedical Analysis.} Graph-structured data are pervasive in biomedical domain such as protein molecules and chemical molecules. GNNs have been on the graphs in biomedical domain for various applications such as drug generation. 
% Privacy attack such as reconstruction attack may lead to the information leakage of the discovered structure that are not intended to be shared.  

\subsection{Future Research Directions of Privacy Preserving GNNs}
\vspace{0.2em} \noindent \textbf{Defense Against Various Privacy Attacks.} Though many privacy-preserving GNNs have been proposed, they mostly focus on defending against membership inference attack and attribute reconstruction attack. The privacy-preservation GNNs against structure attack, property inference attack, and model extraction attack are less studied. Therefore, it is promising to develop  privacy-preserving GNNs against various privacy attacks.

\vspace{0.2em} \noindent \textbf{Privacy Attack and Preservation in GNN Pretraining.} Model pretraining have been a common scheme to benefit the downstream tasks that are lack of labels.  
Recently, pretraining of GNNs with supervised tasks~\cite{hu2019strategies} and self-supervised tasks~\cite{qiu2020gcc,hu2020gpt} have achieved great success. The parameters of pretrained GNNs will be released for downstream tasks, which may lead to private information leakage. However, existing privacy attacks mostly focus on black-box settings and do not investigate the information leakage caused by model releasing. Hence, privacy attack and the corresponding defense methods for pretrained GNN modes need to be explored.

\vspace{0.2em} \noindent \textbf{Trade off Between Privacy and Utility.} Though methods that apply differential privacy, federated learning or adversarial learning have been proposed to protect the privacy of training data, the relations between the privacy protection performance and the prediction accuracy are rarely discussed. For example, in differentially private GNNs~\cite{olatunji2021releasing,sajadmanesh2020locally,zhang2021graph}, the actual performance in defending against various privacy attacks is generally not evaluated. And in adversarial privacy-preserving~\cite{li2020adversarial,liao2021information,wang2021privacy}, how to control the balance between the prediction performance and privacy protection is still not well discussed. 

\section{Robustness of Graph Neural Networks} \label{sec:robust}
As an extension of neural networks on graph-structured data, GNNs are also vulnerable to adversarial attacks. In addition, due to the message-passing mechanism and graph structure, GNNs can be negatively affected by adversarial perturbations on both graph structures and node attributes. For example, Nettack~\cite{zugner2018adversarial} can fool GNNs to give false predictions on target nodes by poison the training graph with unnoticeable perturbations to graph structure and node attributes. 
NIPA~\cite{sun2019node} manages to significantly reduce the global node classification performance of GNNs by injecting a small amount of labeled fake nodes to the training graph. The vulnerability of GNNs has arisen tremendous concerns on adopting GNNs in safety-critical domains such as credit estimation and healthcare. For example, fraudsters can create several transactions with deliberately chosen high-credit users to escape GNN-based fraud detectors, which can cause tremendous loss to individuals, and institutions. Hence, developing robust GNNs is another important aspect of trustworthiness and many efforts have been taken. %And we will introduce the recent advances of adversarial attacks and defense methods on graph-structured data in this section.
There are already several comprehensive surveys about adversarial attacks and defenses on graphs~\cite{jin2020adversarial,chen2020survey,sun2018adversarial,wu2022recent}. Therefore, in this section, we briefly give the overview of adversarial learning on graphs, but focus more on methods in emerging directions such as scalable attacks, graph backdoor attacks, and recent defense methods.

\begin{table}[t]
    \footnotesize
    \centering
    \caption{Categorization of representative graph adversarial attacks.}
    \vskip -1.5em
    \begin{tabular}{lll}
    \toprule
    Aspect & Category & References  \\
    \midrule
    \multirow{2}{*}{Knowledge}& White-box & \cite{wu2019adversarial}, ~\cite{xu2019topology},~\cite{chen2018fast},~\cite{dai2018adversarial},~\cite{geisler2021robustness}\\
    % \midrule
    & Black-box & \cite{dai2018adversarial},~\cite{ma2019attacking},~\cite{zugner2018adversarial},\cite{chen2018fast},~\cite{zugner2019adversarial},~\cite{xu2019topology},~\cite{wang2020scalable},~\cite{sun2019node},~\cite{bojchevski2019adversarial},~\cite{chang2020restricted}\\
    % \cite{ma2020towards},~\cite{tao2021single},~\cite{li2021adversarial},~\cite{xi2021graph},~\cite{zhang2021backdoor}
    \midrule
    \multirow{2}{*}{Capability}& Evasion Attack & \cite{wu2019adversarial}, ~\cite{xu2019topology},~\cite{chen2018fast},~\cite{dai2018adversarial},\cite{tao2021single},~\cite{ma2019attacking},~\cite{ma2020towards},~\cite{ma2019attacking},~\cite{geisler2021robustness}\\
    & Poisoning Attack & \cite{zugner2018adversarial},~\cite{zugner2019adversarial},~\cite{sun2019node},~\cite{bojchevski2019adversarial},~\cite{chang2020restricted},~\cite{wang2020scalable},~\cite{xi2021graph},~\cite{zhang2021backdoor},~\cite{li2021adversarial},~\cite{geisler2021robustness}\\
    \midrule
    \multirow{2}{*}{Attackers' Goal} & Targeted Attack  & \cite{dai2018adversarial},~\cite{wang2020scalable},~\cite{tao2021single},~\cite{xi2021graph},\cite{zhang2021backdoor},~\cite{zugner2018adversarial},~\cite{ma2019attacking},~\cite{chen2018fast},~\cite{li2021adversarial},~\cite{bojchevski2019adversarial} \\
    & Untargeted Attack  & \cite{sun2019node}.~\cite{zugner2019adversarial},~\cite{xu2019topology},~\cite{geisler2021robustness},~\cite{bojchevski2019adversarial},~\cite{wu2019adversarial},~\cite{ma2020towards},~\cite{chang2020restricted} \\
    \bottomrule
    \end{tabular}
    \label{tab:adv}
\end{table}

\subsection{Threat Models of Graph Adversarial Attacks} 
\label{sec:attack_threat}
Graph adversarial attacks aim to degrade the performance of GNNs or to make GNN models give desired output by injecting deliberate perturbations to the graph dataset. Generally, attackers are constrained in the knowledge about the data and the model they attack, as well as the capability of manipulations on the graph. In this subsection, we introduce the threat models in various aspects to show different settings of graph adversarial attacks.

\vspace{0.2em} \noindent \textbf{Attackers' Knowledge}.
Similar to privacy attacks, attackers need to possess certain knowledge about the dataset and target model to achieve the adversarial goal. Based on whether the model parameters are known for the attacker, they can be split into white-box and black-box attack:
\begin{itemize} [leftmargin=*]
    \item \textbf{White-box Attack}: In this setting, the attacker knows all information about the model parameters and the training graph such as adjacency matrix, attribute matrix and labels~\cite{xu2019topology,geisler2021robustness,chen2018fast}. Since it is impractical to obtain all  information in the real world, the white-box attack is less practical but often used to show the worst performance of a model under adversarial attacks. 
    \item \textbf{Black-box Attack}: In black-box attack~\cite{zugner2018adversarial,bojchevski2019adversarial,dai2018adversarial}, attackers do not have access to the model’s parameters but can access graph dataset. More specifically, the full/partial of the graph structure and node features could be accessible for attackers. Attackers may be allowed to have labels used for training or query the outputs of the target GNN, which could be used to mimic the predictions of the target model to achieve  black-box attack.
    % Compared with white-box attack, black-box attack is more challenging and practical
\end{itemize}

% \enyan{general equations each category, Some equations for the representative works. Refer the reader to the survey papers for more details.}
\vspace{0.2em} \noindent \textbf{Attackers' Capability}. 
In adversarial attacks, adversarial perturbations are added to data samples to mislead the target GNN model to give the output desired by the attacker. According to the stage the attack occurs, the attacks can be divided into poisoning attack and evasion attack: 
\begin{itemize}[leftmargin=*]
    \item \textbf{Evasion Attack}: The perturbations in evasion attacks~\cite{dai2018adversarial,xu2019topology} are added to the graph in the test stage, where the GNN model parameters have been well trained and cannot be affected by attackers. Depending on whether the attacker knows the model parameters, evasion attack can be further categorized into white-box or black-box evasion attack.%It is quite often that the model architecture and parameters are assumed accessible in evasion attack.
    \item \textbf{Poisoning Attack}: In poisoning attacks~\cite{zugner2018adversarial,zugner2019adversarial,sun2019node}, the training graph is poisoned before GNNs are trained. The GNN model trained on the poisoned dataset will exhibit certain designed behaviors such as misclassifying target nodes or having low overall performance. As poisoning attack happens before model training, it belongs to the black-box attack where the model parameters are unknown. Thus, attackers usually train a surrogate model and poison the graph to reduce the performance of the surrogate model. Due to the transferability of adversarial attack, the poisoned graph can also reduce the performance of the target GNN trained on it.
\end{itemize}
Currently, many graph mining tasks such as semi-supervised node classification and link prediction are transductive learning, where the test samples participate in training phase. Therefore, most of existing works focus on poisoning attacks, which are often more practical for graph mining.

Based on the way that the graph data is perturbed,  the adversarial attacks can be categorized into manipulation attacks, node injection attack, and backdoor attacks:
\begin{itemize}[leftmargin=*]
    \item \textbf{Manipulation Attack}: In manipulation attack, an attacker manipulates either graph structure or node features to achieve the attack goal. For example, Nettack~\cite{zugner2018adversarial} perturbs the graph by deliberately adding/deleting edges and revising the node attributes with a greedy search algorithm based on gradients. To make the perturbation more unnoticeable,  ReWatt~\cite{ma2019attacking} poisons the training graph by rewiring edges with reinforcement learning.
    \item \textbf{Node Injection Attack}: 
    Different from manipulation attack that modifies the original graph, node injection attack aims to achieve the adversarial goal by injecting malicious nodes into the graph~\cite{sun2019node,wang2020scalable,tao2021single}. Compared to manipulation attack, node injection attack is more practical. For example, in e-commercial network, attackers need to hack servers or user accounts to manipulate the network; while injecting new malicious accounts is much easier.
    \item \textbf{Backdoor Attack}: Backdoor attacks~\cite{xi2021graph,zhang2021backdoor} inject backboor triggers to the training set to poison the model. The backdoor trigger is a predefined or learned pattern, such as a single node or a subgraph. The attacker relabel training nodes/graphs attached with the backdoor trigger to the target label so that a GNN trained on the poisoned dataset will predict any test sample with backdoor trigger to the target label. Compared with other types of adversarial attacks, backdoor attack on GNNs is still in an early stage. 
\end{itemize}

\vspace{0.2em} \noindent \textbf{Attackers' Goal}.
Based on whether the goal of the attacker is to misclassify a set of target instances or reduce the overall performance of GNN model, threat models can also be categorized as:
\begin{itemize}[leftmargin=*]
    \item \textbf{Targeted Attack}: The attacker aims to fool a GNN model to misclassify a set of target nodes~\cite{zugner2018adversarial,zhang2021backdoor}. Meanwhile, the attacker might want the performance of the target model on non-targeted samples remain unchanged to avoid being detected.
    \item \textbf{Untargeted Attack}: The untargeted attack~\cite{zugner2019adversarial,sun2019node} aims to reduce the overall performance of the target GNN model. Since evasion attack cannot affect the parameters, it can only be achieved by poisoning the dataset in the training stage. 
\end{itemize}
The categorization of the existing representative graph adversarial attacks in different aspects are summarized in Table~\ref{tab:adv}. Next we will give more details of these methods.

\subsection{Graph Adversarial Attack Methods}
In this subsection, we first give a unified formulation of adversarial attacks followed by representative methods in evasion attacks and poisoning attacks. Finally, we survey recent advances in backdoor attacks and scalable attacks on GNNs.

\subsubsection{A Unified Formulation of Adversarial Attack}
The adversarial attacks can be conducted on both node-level and graph-level tasks. Since the majority of the literature focuses on node classification problem, we give a unified formulation on node-level graph adversarial attacks as an example, which can be easily extended to other tasks.
\begin{definition} Given a graph $\mathcal{G}=\{\mathcal{V}, \mathcal{E}\}$ with adjacency matrix $\mathbf{A}$ and attribute matrix $\mathbf{X}$, let $\mathcal{V}_T$ be the set of nodes to be attacked, the goal of adversarial attack is to find a perturbed graph $\mathcal{\hat G}$ that meets the unnoticeable requirement by minimizing the following objective function:
\begin{equation} \small
\begin{aligned}
        & \min_{\mathcal{\hat G} \in \Phi(\mathcal{G})} \mathcal{L}_{atk}(\mathcal{\hat G}) = \sum_{v \in \mathcal{V}_T} l_{atk}(f_{\theta^*}(\mathcal{\hat G})_v,y_v) \quad \text{s.t.}  \quad \theta^* = \arg \min_{\theta} 
     \mathcal{L}_{train}(f_\theta(\mathcal{G}'))
     \label{eq:attack}
\end{aligned}
\end{equation}
where $l_{atk}$ is the loss for attacking, which is typically set as $-H(f_{\theta^*}(\mathcal{\hat G})_v,y_v)$ with  $H(\cdot)$ being the cross entropy. %$l_{atk}$ is typically set as $-H(f_{\theta^*}(\mathcal{\hat G})_v,y_v)$.  
$\mathcal{L}_{train}$ is the loss for training target model. Generally, in node classification, we apply the classification loss $\mathcal{L}_{train}=\sum_{v \in \mathcal{V}_L} H(f_\theta (\mathcal{\hat G})_v,y_v)$. As for
$\mathcal{G}'$, it can be either $\mathcal{\hat G}$ or $\mathcal{G}$, which correspond to poisoning attack and evasion attack, respectively. As for the search space $\Phi(\mathcal{G})$, apart from the way of perturbing the graph, it is also constrained by the budget to ensure unnoticeable attacks. Specially, the constraint from the budget is typically implemented as $\|\mathbf{\hat A} - \mathbf{A}\| + \| \mathbf{\hat X} - \mathbf{X}\| \leq \Delta$, where $\Delta$ denotes the budget value. For non-targeted attack, $\mathcal{V}_L$ is set all the unlabeled nodes $\mathcal{V}_U$ and $y_v$ will be the prediction of the unlabeled nodes fro a GNN trained on the clean graph $\mathcal{G}$
\end{definition}

% \subsubsection{Gradient-based Methods}
% \vspace{0.2em} \noindent \textbf{Evasion Attacks}

\subsubsection{Evasion Attacks} \label{sec:evasion}
Evasion attacks focus on the inductive setting, which aims to change the predictions on new nodes/graphs. Based on the applied techniques, evasion attack methods can be split into \textit{Gradient-Based Methods} and \textit{Reinforcement Learning-Based Methods}. % Next, we introduce the representative methods in each category.

\vspace{0.2em} \noindent \textbf{Gradient-Based Methods.}
In the early stage of adversarial attacks~\cite{wu2019adversarial,xu2019topology,chen2018fast,dai2018adversarial}, they generally focus on white-box evasion attack to demonstrate the vulnerability of GNNs and assess the robustness of model under worst situations. Since the model parameters are available in white-box evasion attacks, it is natural to optimize the objective function in Eq.(\ref{eq:attack}) by gradient decent.
More specifically, the objective function of white-box evasion attack can be rewritten as:
\begin{equation} \small
    \min_{\mathcal{\hat G} \in \Phi(\mathcal{G})} \mathcal{L}_{atk}(\mathcal{\hat G}) = \sum_{v \in \mathcal{V}_T} l_{atk}(f_{\theta}(\mathcal{\hat G})_v,y_v),
    \label{eq:evasion}
\end{equation}
where $\theta$ represents mode parameters of the target GNN. However,  due to the discreteness of the graph structure, it is challenging to directly solve the optimization problem in Eq.(\ref{eq:evasion}). Therefore, FGA~\cite{chen2018fast} and GradMax~\cite{dai2018adversarial} use a gradient-based greedy algorithm to iteratively modify the connectivity of the node pair within attack budget. Xu \textit{et al.}~\cite{xu2019topology} adopt projected gradient decent to ensure the discreteness of perturbed adjacency matrix.  Instead of directly using derivatives from the attacks, Wu \textit{et al.}~\cite{wu2019adversarial} use integrated gradients to identify the optimal edges and node attributes to be modified for attack. 

Recently, there are several attempts in developing gradient-based evasion attacks~\cite{ma2020towards,tao2021single} in a more practical black-box setting, which can be applied to graph classification task and node classification on evolving graphs. By exploiting the connection between the backward propagation of GNNs and random walks, Ma \textit{et al.}~\cite{ma2020towards} investigates the connections between the change of classification loss under perturbations and the random walk transition matrix. 
Based on that, they generalize the white-box gradients into a model-independent important scores of PageRank, which avoids using model parameters.  Tao \textit{et al.}~\cite{tao2021single} investigate the black-box evasion attack with single node injection. Without knowing the model parameters, they train a surrogate model on the train graph and attack the surrogate model to inject a fake node to the graph. A parameterized attribute generator and edge generator is adopted for the node injection attack on unseen test nodes. 

\vspace{0.2em} \noindent \textbf{Reinforcement Learning-Based Methods.} In black-box evasion attacks, lacking of model parameters will challenge the gradient-based methods. However, in many scenarios such as drug property prediction, the attacker is allowed to query the target model. In this situation, reinforcement learning can be employed to conduct evasion attacks to learn the optimal actions for graph perturbation~\cite{dai2018adversarial,ma2019attacking}. 
RL-S2V~\cite{dai2018adversarial} is the first work to apply reinforcement learning for black-box targeted attack. They model the attack process as a Markov Decision Process (MDP) defined as:
\begin{itemize}[leftmargin=*]
    \item \textbf{State}: The state $s_t$ at time $t$ is represented by the tuple $(\mathcal{\hat G}_t,v)$, where $\mathcal{\hat G}_t$ is the intermediate modified graph at time $t$ and $v$ is the target node to be attacked.
    \item \textbf{Action}: The attacker of RL-S2V needs to add or delete an edge in each step, which is equivalent to select a node pair. It decomposes the node pair selection action $a_t \in \mathcal{V} \times \mathcal{V}$ into a hierarchical structure of sequentially selecting two nodes, i.e., $a_t=(a_t^{(1)},a_t^{(2)})$, where $a_t^{(1)}, a_t^{(2)} \in \mathcal{V}$. 
    \item \textbf{Reward}: The goal of the attack is to fool the target classifier $f$ on the target node $v$.  In RL-S2V, no reward is given in intermediate steps. The non-zero reward is only given at the end as:
    \begin{equation} \small
        r(s_m, a_m) = 
        \left \{ \begin{array}{ll}
         1 & \mbox{if $f(\mathcal{\hat G}_m)_v \neq y_v$};\\
        -1 & \mbox{if $f(\mathcal{\hat G}_m)_v = y_v$},\end{array} \right.
        \label{eq:reward_graph}
    \end{equation}
    
    \item \textbf{Termination}: The process will stop once the agent modifies $m$ edges. 
\end{itemize}
RL-S2V adopts Q-learning algorithm to solve the MDP problem for targeted attack. A parameterized Q-learning is implemented for better transferablility. Similar reinforcement learning framework is also applied in ReWatt~\cite{ma2019attacking}. To make the perturbations more unnoticeable, ReWatt perturbs the graphs by rewiring, i.e., break an edge $e_{ij}$ and rewire the edge to another node to form $e_{ik}$. Hence,  ReWatt employs a different action space that consists of all the valid rewiring operations.
% \enyan{Graph-level task, it is reasonable.}
% \enyan{node classification: transaction network.}
\subsubsection{Poisoning Attacks by Graph Manipulation} \label{sec:poisoning}
Apart from the evasion attacks, extensive poisoning attack methods~\cite{dai2018adversarial,sun2019node,ma2019attacking} have been investigated for the transductive learning setting, which are more practical for semi-supervised node classification. The majority of them focus on perturbing the graph data by manipulating the original graph~\cite{zugner2018adversarial,zugner2019adversarial}. From a technical standpoint, most poisoning attacks through manipulation can be categorized under gradient-based methods.

% \vspace{0.2em} 
As Eq.(\ref{eq:attack}) shows, the poisoning attack can be formulated as a bilevel optimization problem. To address this problem, various methods~\cite{zugner2018adversarial,chen2018fast,zugner2019adversarial,xu2019topology} that perform gradient-based attacks on a static/dynamic surrogate model have been investigated. For instance, Nettack~\cite{zugner2018adversarial} deploys a tractable  surrogate model, i.e., $f_S(\hat{\mathbf{A}}, \mathbf{X})=\text{softmax}(\mathbf{\hat A}^2\mathbf{X}\mathbf{W})$, to conduct poisoning targeted attack. The surrogate model is firstly trained to capture the major information of graph convolutions. Then, the poisoning attack is reformulated to learning perturbations by attacking the surrogate model as:
\begin{equation}
    \underset{ (\mathbf{\hat A}, \mathbf{\hat X}) \in \Phi(\mathbf{A},\mathbf{X})}{\arg \max} \mathcal{L}(\mathbf{\hat A}, \mathbf{\hat{X}};\mathbf{W},v) = \max_{c \neq y_v} [\mathbf{\hat{A}}^2 \mathbf{\hat X} \mathbf{W}]_{vc} - 
    [\mathbf{\hat{A}}^2 \mathbf{\hat X} \mathbf{W}]_{vy_v},
    \label{eq:nettack}
\end{equation}
where $\Phi(\mathbf{A},\mathbf{X})$ denotes the search space under the unnoticeable constraint, which considers both attack budgets and maintaining important graph properties. To solve Eq.(\ref{eq:nettack}), Nettack proposes an effective way of evaluating the change of surrogate loss after adding/removing a feature or an edge. The final poisoned graph is obtained by repeatably selecting the most malicious operation in a greedy search manner until reaching the budget. Similarly, FGA~\cite{chen2018fast} adopts a static surrogate model for poisoning attack. Different from Nettack, FGA directly adopts the GCN model and select the perturbations based on gradients.  

As the static surrogate model is trained on the raw graph, which cannot accurately reflect the performance of GNN on the poisoned graph, there are also many works~\cite{zugner2019adversarial,xu2019topology} that adopt a dynamic surrogate model to consider the effects of added perturbations to the target model. In~\cite{xu2019topology}, min-max topology attack generation is employed for untargeted attack. Specifically, the inner maximization updates the surrogate model on the partial modified graph, and the outer minimization conducts projected gradient decent topology attack on GNN model. Metattack~\cite{zugner2019adversarial} introduces meta-learning to solve the bi-level optimization on the untargeted attack. Essentially, the graph structure matrix is treated as a hyperparameter and the meta-gradient is computed as:
\begin{equation}
    \nabla_{\mathcal{G}}^{meta} = \nabla_\mathcal{G}\mathcal{L}_{atk}(f_{\theta^{*}}(\mathcal{G})) = \nabla_f \mathcal{L}_{atk}(f_{\theta^*}(\mathcal{G}))\cdot [\nabla_{\mathcal G} f_{\theta^*}(\mathcal{G})+ \nabla_{\theta^*}(f_{\theta^*}(\mathcal{G})\cdot\nabla_{\mathcal{G}}\theta^*], 
\end{equation}
where $\theta^*$, i.e., the parameters of surrogate GNN model,  is usually obtained by gradient descent in $T$ iterations. For each inner iteration, the gradients of $\theta_{t+1}$ is obtained by $\nabla_{\mathcal{G}}\theta_{t+1}=\nabla_{\mathcal{G}}\theta_t - \alpha \nabla_{\mathcal{G}}\nabla_{\theta_t}\mathcal{L}_{train}(f_{\theta_t}(\mathcal{G}))$, where $\alpha$ is the learning rate of the gradient descent in the inner iteration.

Though surrogate model provides a way for poisoning attack, if the target model architecture differs a lot from the surrogate model, the poisoned graph might not able to significantly reduce the performance of the target model when trained on it. Thus, instead of using surrogate models, some works~\cite{bojchevski2019adversarial,chang2020restricted} design generalized attack loss functions to poison the graph to improve the transferability. %embedding learning and downstream tasks.
For example, Bojchevski \textit{et al.}~\cite{bojchevski2019adversarial} exploit the eigenvalue perturbation theory to efficiently approximate the unsupervised DeepWalk loss to manipulate the graph structure. It further demonstrates that the perturbed graph structure is transferable to various GNN models such as GCN. Graph embedding learning is formulated as a general signal process with corresponding graph filter in GF-Attack~\cite{chang2020restricted}, which enables a general attacker that theoretically provides transferability of adversarial samples. GF-Attack is able to perturb both adjacency matrix and feature matrix, which leads to better attacking performance than~\cite{bojchevski2019adversarial}.

% \vspace{0.2em} \noindent \textbf{Reinforcement Learning-based methods.} Several reinforcement learning-based methods~\cite{sun2019node} are also investigated for positioning attack. 

\subsubsection{Node Injection Attacks} 
Node injection attacks aim to achieve the adversarial goal by adding malicious nodes. This attack process will not affect the existing link structures and node attributes. Hence, node injection is more practical to execute compared with manipulation attacks.
A uniform objective function of node injection attack can be formulated as:
\begin{equation} \small
\begin{aligned} 
    \min_{\mathcal{E}_A, \mathcal{V}_A} \mathcal{L}_{atk}(\mathcal{\hat G}) = \sum_{v \in \mathcal{V}_T} l_{atk}(f_{\theta^*}(\mathcal{\hat G})_v,y_v) \quad \text{s.t.}  \quad \theta^* = \arg \min_{\theta} 
     \mathcal{L}_{train}(f_\theta(\mathcal{\hat G}))
\end{aligned}
\end{equation}
Where $\mathcal{E}_A$ is the set of edges that link nodes inside $\mathcal{V}_A$ and connect $\mathcal{V}_A$ with clean graph $\mathcal{G}$. 

Node injection attack is firstly proposed by NIPA~\cite{sun2019node}, which assigns fake nodes to degrade the overall performance of the target GNN. To achieve stronger attack performance, the attacker in NIPA will also add the labels of the malicious nodes into the training set.
The objective function is solved by reinforcement learning, which is similar to RK-S2V that described in Sec.\ref{sec:evasion}. But the action space and design of reward are designed for node injection positioning attack. Specifically, in each step, the action is to connect a fake node to the graph and assign a class label to the fake of node to effectively fool GNN trained on the poisoned graph. The action is decomposed into three steps: (i) select a fake node; (ii) select a real node and connect to the fake node selected; and (iii) assign the label to the selected fake node. The reward is defined based on the performance of the surrogate model trained on the poisoned graph. Let $A_t$ denote the attack success rate on the surrogate model that trained on $\mathcal{\hat G}_t$. The reward is defined as $r_t(s_t, a_t) = 1$ if $A_{t+1} > A_{t}$, where $s_t$ and $a_t$ denotes the state and action at time $t$; Otherwise, $r_t(s_t, a_t) = 0$.

Beyond the practicality, scalability is another advantage of node injection attacks. When we inject a malicious node for adversarial attacks, we only need to consider $d \cdot N$ options, where $d$ and $N$ are the expected degree of the fake node and the graph size. Since $d$ is generally very small and even can be set as one~\cite{tao2021single}, the node injection attack is naturally more scalable than the manipulation attack. Several following works~\cite{wang2020scalable,tao2021single,zou2021tdgia} further investigate the node injection attacks for better scalability. AFGSM~\cite{wang2020scalable} approximately linearizes the target GCN and derives a closed-form solution for a node injection targeted attack, which has much lower time cost. Experiments on Reddit with over 100K nodes demonstrate its effectiveness and efficiency for targeted attack. G-NIA~\cite{tao2021single} explore to conduct targeted attack by injecting a singe node, which avoids the cost of generating multiple nodes for the attack. To efficiently choose nodes connecting with the injected ones, TDGIA~\cite{zou2021tdgia} introduces a topological defective edge selection strategy. Specifically, a metric based on basic graph structural information is employed to assess nodes most impacted by perturbations in their neighbors. To generate the features for the injected nodes, a the smooth feature optimization objective is designed in TDGIA.
% It models the optimization problem of node injection attack by a parameterized model, which contains a node generator and an edge generator. The node generator and edge generator are trained by the attack loss on training nodes to gain the generalization ability in attacking test nodes. 

Recently, the limitations of node injections attacks in unnoticeability are explored in~\cite{chen2022understanding}. In this work, the authors observed that these injected nodes and edges can disrupt the homophily distribution in the original graphs, compromising its unnoticeability. To solve this issue, \textit{Chen et al.} propose homophily unnoticeability to regularize the node generation and attachment. In their experiments, the homophily regularization is applied on various existing node injection attacks. Extensive results on many massive size graphs indicate the effectiveness of the proposed homophily regularization.

% In this work, we explored the advantages and limitations of graph injection attacks and found they tend to break the unnoticeability that even destroys the homophily distribution in the original graphs. This issue can bring unreliable conclusions when evaluating GNNs’ robustness with injection attacks. Hence, we propose homophily unnoticeability to regularize this behavior as well as a practical objective to realize the constraint, with the hope that it can mitigate the ill-defined unnoticeability issue in general graph adversarial attacks

\subsubsection{Graph Backdoor Attacks} \label{sec:backdoor}
In this subsection, we review the recent emerging graph backdoor attacks. In backdoor attacks, the attacker aims to make the host system to misbehave when the pre-defined trigger is present. The backdoor attack on graphs can be applied in various scenarios, which can largely threat the applications of GNNs. For example, an attacker may inject backdoor trigger to the drug evaluation model that possessed by the community. Then, even a useless drug containing the backdoor trigger from the attacker will be classed as an effective medicine. In addition, the backdoor can be applied to federated learning~\cite{xu2022more}, which threatens the safety of the final global model. 
Though successful backdoor attacks may lead to enormous loss, there are only few works on backdoor attack methods on graphs~\cite{xi2021graph,zhang2021backdoor}. Next, we will introduce a general framework of graph backdoor attacks followed by details of representative works.

\begin{figure}[t]
    \centering
    \includegraphics[width=0.93\linewidth]{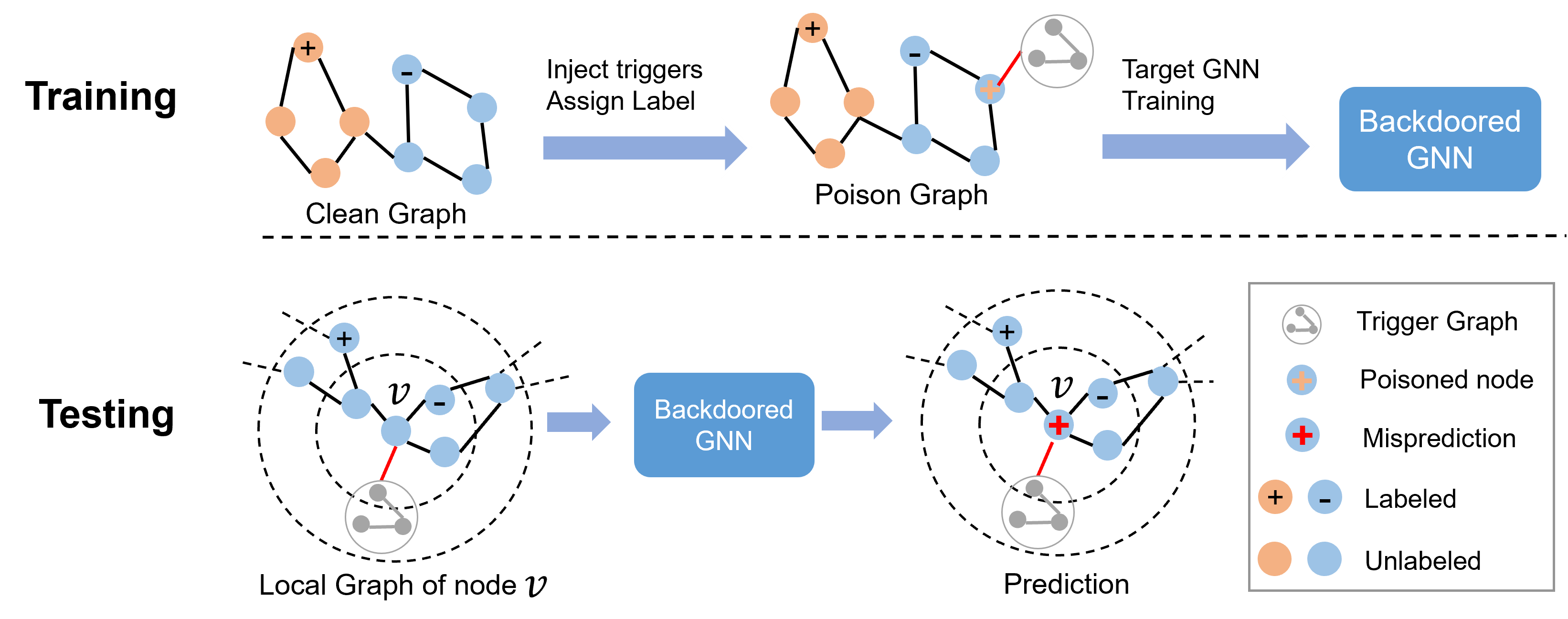}
    \vskip -1em
    \caption{General framework of graph backdoor attack~\cite{dai2023unnoticeable}.}
    \vskip -1em
    \label{fig:bkd_framework}
\end{figure}

\vspace{0.2em} \noindent \textbf{General Framework of Graph Backdoor Attacks}
The key idea of the backdoor attacks is to associate the trigger with the target class in the training data to mislead target models. 
As illustrated in Fig.~\ref{fig:bkd_framework}, during the poisoning phase, the attacker will attach a trigger $g$ to a set of poisoned nodes $\mathcal{V}_P \subseteq \mathcal{V}$ and associate $\mathcal{V}_P$ with the target class label $y_t$. This process generates a backdoored dataset. The GNNs trained on the backdoored dataset will be optimized to predict the poisoned nodes $\mathcal{V}_P$ (attached with the trigger $g$) as target class $y_t$. This association  will force the target GNN to relate the existence of the trigger $g$ in neighbors with the target class. In the test phase, the attacker can control the prediction on the target node $v$ to be $y_t$ by attaching the trigger $g$ to node $v$.

\vspace{0.2em} \noindent \textbf{SBA}~\cite{zhang2021backdoor} proposes a subgraph-based backdoor attack for graph classification.  In~\cite{zhang2021backdoor}, the subgraph trigger is generated by random graph generation algorithms such as Erdős-Rényi model and small world model. The set of poisoned nodes $\mathcal{V}_P$ is randomly sampled. In this paper, the authors also investigate the impacts of the trigger size, trigger density, and the size of poisoned graphs, which demonstrates the vulnerability of GNN models to backdoor attacks. 

\vspace{0.2em} \noindent \textbf{GTA}~\cite{xi2021graph} focuses on injecting backdoor to a pretrained model for node/graph-level tasks. The backdoor is expected to remain effective even after fine-tuning the pretrained model on downstream tasks. Hence, GTA formulates it as the following bi-level optimization problem
\begin{equation} \label{eq:obj_gta} \small
    \mathcal{G}_t^* = \arg \min_{\mathcal{G}_t} \mathcal{L}_{atk}(\theta^*(\mathcal{G}_t)) \quad \text{s.t.}~\theta^* = \arg \min_\theta \mathcal{L}_{train}(\theta, \mathcal{D}_C \cup \mathcal{\hat D}_P), 
\end{equation}
where $\mathcal{D}_C$ and $\mathcal{\hat D}_P$ represent the clean training samples and the samples poisoned by $\mathcal{G}_t$, respectively. In~\cite{xi2021graph}, the downstream tasks are assumed unknown. They adopt an unsupervised training and design an attack loss that forces the poisoned graphs/nodes have similar embeddings with a pre-defined sample. 
Eq.(\ref{eq:obj_gta}) is solved by iteratively conducting inner and outer optimization with a first-order approximation.
% \enyan{More details of difference on attacking node and graph classification}
In addition, a parameterized backdoor generator is adopted to obtain personalized backdoor for each test graph/node.

\vspace{0.2em} \noindent \textbf{UGBA}~\cite{dai2023unnoticeable}. While SBA and GTA demonstrate impressive attack performance, \textit{Dai et al.}~\cite{dai2023unnoticeable} observed that the triggers they utilize differ significantly from the attached poisoned nodes. This violates the homophily property of the graph. Hence, trigger attachments are vulnerable, as eliminating edges connecting dissimilar nodes can easily disrupt them. Moreover, existing works require expensive budget to backdoor GNNs trained on large datasets. To address these problems, \textit{Dai et al.}~\cite{dai2023unnoticeable} develop UGBA~\cite{dai2023unnoticeable} to execute unnoticeable grpah backdoor attacks with limited  budget. To make more efficient use of the budget, UGBA will attach triggers with deliberately chosen poisoned nodes denoted as $\mathcal{V}_P$. To ensure unnoticeability, an adaptive trigger generator is deployed to obtain trigger $g_i$ for node $v_i$ under the following constraint:
\begin{equation} \small
    \min_{(u,v) \in \mathcal{E}_B^i} sim(u,v) \geq T, 
    \label{eq:UGBA_const}
\end{equation}
where $\mathcal{E}_B^i$ denotes the edge set that contain edges inside trigger $g_i$ and edge attaching trigger $g_i$ to node $v_i$. $sim$ represents the cosine similarity on node features. $T$ is the threshold of the similarity score which can be tuned based on datasets. The objective functon of UGBA can be formluated as:
\begin{equation} \small
\begin{aligned} 
    \min_{\mathcal{V}_P, \theta_{g}} & \mathcal{L}_{atk} (\mathcal{V}_P, \theta_{g}) =  \sum_{v_i \in \mathcal{V}_U} l(f_{\theta^*}(\mathcal{\hat G}_i), y_t) \\
    s.t. ~~  & \theta^* = \mathop{\arg\min}_{\theta}  \sum_{v_i \in \mathcal{V}_L}l(f_{\theta}(\mathcal{G}_i), y_i) + \sum_{v_i \in \mathcal{V}_P}l(f_{\theta}(\mathcal{\hat G}_i), y_t), \\
    & \forall v_i \in \mathcal{V}_P \cup \mathcal{V}_{U}, ~~\text{$g_i$ meets Eq.(\ref{eq:UGBA_const}) and }|g_i| < \Delta_{g}, ~~ |\mathcal{V}_P| \leq \Delta_{P} 
    \label{eq:UGBA_opt}
\end{aligned}
\end{equation}
where $\mathcal{\hat G}_i=a(\mathcal{G}_i,g_i)$ denotes the computation graph of node $v_i$ after attaching the generated trigger $g_i$. $l(\cdot)$ represents the cross entropy loss and $\theta_g$ denotes the parameters of the adaptive trigger generator. UGBA splits the optimization process in Eq.(\ref{eq:UGBA_opt}) into poisoned node selection and adaptive trigger generator learning. During the poisoned node selection phase, representative nodes situated at the clustering center are chosen.  As for the training of the adaptive generator, a bi-level
optimization with a surrogate GCN model is applied. 

\subsubsection{Scalable Attack Methods} \label{sec:scalable_attack}
In real-world scenarios, the graph to be attacked is often in large scale. For example, the Facebook social network contains billions of users. It is challenging to conduct adversarial attacks on such large-scale network. \textit{First}, the majority of existing works  focus on manipulation attacks that try to find the optimal node pairs to be manipulated, which will require very high computation cost. For example, the time and space complexity of Mtattack is $O(N^2)$ with $N$ being the number of nodes in the graph, as it requires to compute the meta-gradient of each node pair. \textit{Second}, the large-scale graph may exhibit different properties. Therefore, the attack methods on small graphs could be ineffective on large graphs. However, there are only few works investigating the vulnerability of GNNs on large-scale graphs~\cite{geisler2021robustness}.
In this section, we present three promising directions of scalable attacks.

\vspace{0.2em} \noindent \textbf{Perturbation Sampling}: 
As mentioned, the gradient-based manipulation attack needs to compute the gradients of each pair of node to decide the perturbation on the topology, resulting in unaffordable time and space complexity. In~\cite{geisler2021robustness}, Projected Randomized Block Coordinate Descent (PR-BCD) is proposed to sample the perturbations and update their corresponding probability scores so as to reduce time complexity. Specifically, the  manipulation attack on graph topology is modeled as $\min_{\mathbf{P}} \mathcal{L}_{atk}(f_\theta(\mathbf{A}  \oplus \mathbf{P}), \mathbf{X})$, where $\sum \mathbf{P} \leq \Delta$  and  $\mathbf{P}_{ij}=1$ denotes an edge flip. The operation $\oplus$ stands for the operation of element-wise edge flipping. To optimize $\mathbf{P}$, it is relaxed to $\mathbf{P} \in [0,1]^{N \times N}$, where $\mathbf{P}_{ij}$ is the probability of flipping the edge for attacks. In each training iteration, PR-BCD randomly samples a fixed number of perturbations, which lead to a sparse adjacency matrix for both forward and backward computation. The final perturbations can be obtained by the flipping probability score matrix $\mathbf{P}$. This method is applicable for both targeted and untargeted attacks.
Experiments on massive graphs with over 100 million nodes empirically show the effectiveness of PR-BCD and the vulnerability of GNNs on large-scale graphs.

% As it mentioned,  the major bottleneck of scalable attacks is the time and space complexity, due to the need of update gra

\vspace{0.2em} \noindent \textbf{Candidates Reduction}: For attack on a target node $v$, the search space can actually be reduced as many candidate topology manipulations are ineffective to affect the prediction on $v$. For instance, linking two nodes far from the target node $v$ can hardly change the representation of node $v$. And linking a node $u$ that share the same label as $v$ to $v$ can even lead to a more robust graph structure~\cite{dai2022towards}. Therefore, SGA~\cite{li2021adversarial} develops a mechanism of perturbation candidate reduction to avoid excessive computation. 
First, for manipulation on a node pair $(v_1,v_2)$, one of the nodes, say $v_1$, should be in the computation graph of the target node $v$, i.e., $k$-hop subgraph of node $v$ in a $k$-layer GNN. Second, for the other node $v_2$, its class label should be the one that is most easily misclassified to the original class of target node $v$. Finally, SGA assumes that a node $v_2$ is more likely to be selected if node $v_2$ can largely affect the prediction of $v$ when the manipulation is directly on node pair $(v,v_2)$. Therefore, several best candidates of $v_2$ for each manipulation can be selected. With these strategies, the time complexity can be reduced to be linear with the graph size.

\subsection{Robust Graph Neural Networks}
As GNNs are vulnerable to adversarial attacks, various robust graph neural networks against adversarial attacks have been proposed, which can be generally categorized into three types: \textit{Adversarial Training}, \textit{Graph Denoising}, and \textit{ Certifiable Robustness}. Next, we will introduce the representative methods of each category and some defense methods lie in other category.

\subsubsection{Adversarial Training}
Adversarial training is a popular and effective approach to defend against adversarial evasion attacks, which has been widely applied in computer vision~\cite{goodfellow2014explaining} to defend against evasion attacks. Generally,  adversarial training simultaneously generate adversarial samples that can fool a classifier and force the classifier to give similar predictions for a clean sample and its perturbed version so as to improve the robustness of the classifier. Adversarial training~\cite{dai2019adversarial,deng2019batch,xu2019topology,feng2019graph,wang2022robust} is also investigated to 
defend against graph adversarial attacks, which can be generally formulated as the following min max game:
\begin{equation} \small
    \min_{\theta} \max_{\Delta_{\mathbf A} \in \mathcal{P}_{\mathbf{A}}, \Delta_{\mathbf x} \in \mathcal{P}_{\mathbf{X}}} \mathcal{L}_{train}(f_\theta(\mathbf{A} + \Delta_{\mathbf{A}},\mathbf{X} + \Delta_{\mathbf{X}})),
\end{equation}
where $\mathcal{L}_{train}$ is the classification loss on the labeled nodes. $\Delta_{\mathbf{A}}$ and $\Delta_{\mathbf{X}}$ stand for the perturbations on the topology structure and node attributes, respectively. $\mathcal{P}(\mathbf{A})$ and $\mathcal{P}(\mathbf{X})$ denote the allowable perturbations within the attack budget. Adversarial training on GNN is firstly explored in~\cite{xu2019topology}, where the perturbations on graph topology are generated by a PGD algorithm. Some variants of~\cite{xu2019topology} are investigated in~\cite{chen2019can,wang2019graphdefense}. Considering that node classification is a semi-supervised learning task by nature, virtual graph adversarial training~\cite{deng2019batch,feng2019graph} is applied to further encourage the smoothness of model predictions on both labeled and unlabeled nodes. In~\cite{deng2019batch,feng2019graph}, only node feature perturbations are considered in the virtual graph adversarial training as:
\begin{equation}
    \min_{\theta} \max_{\Delta_{\mathbf{X}} \in \mathcal{P}(\mathbf{X})} \mathcal{L}_{train} (f_\theta(\mathbf{A},\mathbf{X} + \Delta_{\mathbf{X}})) + \alpha \sum_{v_i \in \mathcal{V}} D_{KL}(p(y|\mathbf{x}_i; \theta)||p(y|\mathbf{x}_i+\Delta_{\mathbf{x}_i}; \theta)),
    \label{eq:vat}
\end{equation}
where $\alpha$ is the weight of virtual adversarial training regularizer. $p(y|\mathbf{x}_i$ denotes the prediction of node $v_i$. $D_{KL}(p(y|\mathbf{x}_i; \theta)||p(y|\mathbf{x}_i+\Delta_{\mathbf{x}_i}; \theta))$ enforces the predictions on unlabeled nodes to be similar with and without perturbation. %As shown in Eq.(\ref{eq:vat}), virtual adversarial training regularization will enforce the smoothness of predictions on all instances including labeled and unlabeled nodes. 

\subsubsection{Certifiable Robustness}
% In contrast to the adversarial training, certifiable robustness aims to provide guarantees that no
Though various approaches such as graph adversarial training have been proposed to improve robustness against adversarial samples, new attacks may be developed to invalidate the defense methods, leading to an endless arms race. To address this problem, recent works~\cite{zugner2019certifiable,wang2021certified} analyze the certifiable robustness of GNNs to understand how the worst-case attacks will affect the model. 
Certifiable robustness aims to provide certificates to nodes that are robust to potential perturbations in considered space. For each node $v \in \mathcal{V}$ with label $y$, the certificate $m(v;\theta)$ can be obtained by solving the following optimization problem
\begin{equation}
    m(v;\theta)= \min_{\mathcal{\hat G} \in \Phi (\mathcal{G})} \max_{i \neq y} f_\theta(\mathcal{\hat G})_{vy}-f_\theta(\mathcal{\hat G})_{vi},
\end{equation}
where $f_\theta(\mathcal{\hat G})_{vi}$ denotes the predicted logit of node $v$ in class $i$ and $\Phi(\mathcal{G})$ indicates all allowable perturbed versions of the graph. If $m(v)>0$, the GNN is certifiably robust w.r.t. node $v$ in considered space $\Phi(\mathcal{G})$. In other other, any adversarial samples in $\Phi(\mathcal{G})$ cannot change the prediction to node $v$ from the target model. 
The work~\cite{zugner2019certifiable} firstly investigates the certifiable robustness against the perturbations on node features. Some following works~\cite{wang2021certified,jin2020certified,zugner2020certifiable,bojchevski2019certifiable} further analyze the certifiable robustness under topology attacks. For instance, ~\cite{zugner2020certifiable} proposes a  branch-and-bound algorithm that obtains a tight  bound on the global optimum of the certificates for topology attacks. In~\cite{bojchevski2019certifiable}, the certificates of a page-rank and a family of GNNs  such as APPNP~\cite{klicpera2018predict} are efficiently computed by exploiting connections to PageRank and Markov decision processes. A technique of randomized smoothing is applied in~\cite{wang2021certified} to give certifiable guarantees to any GNNs. The randomized smoothing will inject noises to the test samples to mitigate the negative effects of the adversarial perturbations. And the obtained certificates are proven to be tight.

Apart from methods of computing certificates of a trained GNN, robust training that aims to increase the certifiable robustness is also investigated in~\cite{zugner2019certifiable,bojchevski2019certifiable}. The main idea is to directly maximize the worst-case margin $m(v;\theta)$ during training to encourage the model to learn more robust weights. In particular, a robust hinge loss can be added to the training loss to improve certifiable robustness, which can be formally written as 
\begin{equation} 
    \min_{\theta} \mathcal{L}_{train}(f_\theta(\mathcal{G})) + \sum_{v \in \mathcal{V}} \max(0, M-m(v;\theta)),
    \label{eq:cert_loss}
\end{equation}
where $\mathcal{L}_{train}$ denotes the classifcation loss, and $M>0$ is the hyperparameter for the hinge loss. The worst-case margin,i.e., $m(v;\theta)$  is encouraged to be larger than $M$ with Eq.(\ref{eq:cert_loss}).  

\subsubsection{Graph Denoising}
The adversarial training and certifiable robustness are effective to train robust GNNs to defend against evasion attacks, i.e., attacks happen in the test stage. However, they cannot deal with a poisoned training graphs which have been perturbed by adversarial attacks. In the early investigation about poisoning attacks~\cite{wu2019adversarial}, topology attack is found to be more effective and in favor by the positioning attacks; while feature-only perturbations generally fail to change the predictions of the target node due to the high dimension of node attribute. Therefore, a promising direction of defensing positioning attacks is to denoise the graph structure to reduce the negative effects of the injected perturbations. Based on the way of denoising the graph, existing methods can be split into \textit{Pre-processing}, \textit{Graph Structure Learning}, and \textit{Attention-Based methods}.

\vspace{0.2em} \noindent \textbf{Pre-processing.} Pre-processing based approaches first denoise the graph using heuristics about network properties or attack behaviors. Then, the GNN model can be trained on the denoised graph to give correct predictions that are not affected by the poisoning attacker. The work~\cite{wu2019adversarial} firstly proposes a simple and effective pre-processing defense method based on the following two observations on graph adversarial attacks : (i) perturbing graph structures are more effective than modifying the node attributes; and (ii) attackers tend to add adversarial edges by linking dissimilar nodes instead of deleting existing edges. Hence, GCN-Jaccard is proposed in~\cite{wu2019adversarial} to defend against adversarial attacks by eliminating the edges connecting nodes with low Jaccard similarity of node features. Experimental results show the effectiveness and efficiency of this  defense method. Apart from the observations about the properties of linked nodes, the adversarial attack  is found to result in high-rank spectrum of adjacency matrix, which corresponds to low singular values~\cite{entezari2020all}. Based on the observation of high-rank attack, low-rank approximation with truncated SVD is used to denoise the graph to resist poisoning attacks. Specifically, they retain a truncated SVD that contains only the top-k singular values of the adjacency matrix. Then, the denoised graph can be reconstructed from the truncated SVD. Their experiments show that only keeping the top 10 singular values of the adjacency matrix is able to defend against Nettack~\cite{zugner2018adversarial}.

\vspace{0.2em} \noindent \textbf{Graph Structure Learning.}
Graph structure learning methods~\cite{jin2020graph,luo2021learning,dai2022towards} aim to simultaneously learn a denoised graph and a GNN model that can give accurate predictions based on the denoised graph. Inspired by the fact that adversarial attacks will lead to high-rank adjacency matrix, Pro-GNN~\cite{jin2020graph} proposes to learn a clean adjacency matrix $\mathbf{S}$ with the constraint that (i) $\mathbf{S}$ is low-rank and close to the raw perturbed adjacency matrix $\mathbf{A}$ such that the adversarial edges are likely to be removed; (ii) $\mathbf{S}$ should facilitate node classification; and (iii) $\mathbf{S}$ should maintain feature smoothness, i.e., link nodes of similar features.  The overall objective function of Pro-GNN can be written as:
\begin{equation}
    \min_{\theta, \mathbf{S}} \mathcal{L}_{train}(f_\theta(\mathbf{X},\mathbf{S})) + \alpha\|\mathbf{A}-\mathbf{S}\|_F^2 + \beta\|\mathbf{S}\|_* + \lambda tr(\mathbf{X}^T \hat{\mathbf{L}} \mathbf{X}),
\end{equation}
%where $\mathbf{A}$ denotes the given perturbed graph adjacency matrix.
where $\mathcal{L}_{train}(f_\theta(\mathbf{X},\mathbf{S}))$ is the classification loss using $\mathbf{S}$.  $\|\mathbf{S}\|_*$ stands for the nuclear norm of the learned adjacency matrix $\mathbf{S}$ to encourage low-rank of $\mathbf{S}$. The last term $tr(\mathbf{X}^T \hat{\mathbf{L}} \mathbf{X})$ is to encourage the learned adjacency matrix to link nodes of similar features, where $\hat{\mathbf{L}}$ is the Laplacian matrix of $\mathbf{S}$. Similar to Pro-GNN, PTDNet~\cite{luo2021learning} also adopt a low-rank constraint to learn to drop noisy edges in an end-to-end manner. But different from Pro-GNN which directly optimizes the adjacency matrix of denoised graph, PTDNet deploys a parameterized denoising network to predict whether to remove the edge with the representations of two nodes from a GNN model. 

Though the defense methods using low-rank constraint are proven to be effective, the computation cost of nuclear norm is too expensive for large-scale graphs. 
Recently, a robust structural noise-resistant GNN (RS-GNN)~\cite{dai2022towards} is proposed to learn an link predictor that efficiently eliminate/down-weight the noisy edges with the weak supervision from the adjacency matrix. In real-world graphs, nodes with similar features and labels tend to be linked; while noisy edges would link nodes of dissimilar features. Therefore, RS-GNN deploys a MLP to predict the weight of the link between $v_i$ and $v_j$ by: $w(i,j)=f(\mathbf{h}_i^T \mathbf{h}_j)$, where $\mathbf{h}_i=MLP(\mathbf{x}_i)$ and $f$ is the activation function such as sigmoid. A novel feature similarity weighted edge-reconstruction loss to train the link predictor to encourage lower weights are assigned to noisy edges. The link prediction is further utilized to predict the missing links in the graph, which can involve more unlabeled nodes in the training to address the challenge of label sparsity. 

Reinforcement learning is also applied in graph structure learning for robust representation learning~\cite{wang2019learning}. Specifically, Graph Denoising Policy Network (GDPNet)~\cite{wang2019learning} focuses on denoising the one-hop subgraph of each node. Whether to involve the neighbors of a node is determined sequentially.
Therefore, the action space would be whether the selected node $u_t \in \mathcal{N}(v)$ at step $t$ should be linked with $v$. 
The state $s_t=[\mathbf{h}_v^t, \mathbf{h}_{u_t}]$ contains the representations of node $v$ by aggregating the previously selected neighbors $\mathcal{\hat N}_t(v)$ and the selected node $u_t$.
The prediction scores on the downstream tasks are used as the reward signal for the neighbor selection phase.

\vspace{0.2em} \noindent \textbf{Attention Mechanism.}
Attention-based defense methods~\cite{tang2020transferring,zhang2020gnnguard} aim to penalize the weights of adversarial edges or nodes in the aggregation of each GNN layer to learn robust representations. In~\cite{tang2020transferring}, PA-GNN utilizes  auxiliary clean graphs that share the same data distribution with the target poisoned graph to learn to penalize the adversarial edges.  Specifically, adversarial edges are injected to the clean graphs to provide supervision for the penalized aggregation mechanism. Let $\text{a}_{ij}^{(l)}$ be the attention score assigned to the edge linking $v_i$ and $v_j$ in $l$-th GNN layer. PA-GNN wants the attention weights of clean edges to be larger than the perturbed edges by a margin $\eta$ as 
\begin{equation}
    \mathcal{L}_{dist} = - \min \Big ( \eta, \  \mathop{\mathbb{E}}_{\substack{ (v_i.v_j) \in \mathcal{E}_C,  1 \leq l \leq L}} \text{a}_{ij}^{(l)} - \mathop{\mathbb{E}}_{\substack{(v_i.v_j) \in \mathcal{E}_P, 1 \leq l \leq L}} \text{a}_{ij}^{(l)} \Big ),
\end{equation}
where $\mathcal{E}_C$ and $\mathcal{E}_P$ are the set of clean edges and perturbed adversarial edges from auxiliary graphs. PA-GNN further adopts Meta-learning to transfer the ability of penalizing adversarial edges to GNN on the target graph. GNNGuard~\cite{zhang2020gnnguard} computes the attention scores based on the cosine similarity of node representations from last layer. With the similarity-based attention, the adversarial edges are likely to be assigned with small weights since they generally link dissimilar nodes.

\subsubsection{Other Types of Defense Methods Against Graph Adversarial Attacks}
In this subsection, we briefly introduce defense methods that do not belong to the aforementioned categories.

\vspace{0.2em} \noindent \textbf{Robust Aggregation.} Some efforts~\cite{zhu2019robust,geisler2021robustness,chen2021understanding} have been taken to design a robust aggregation mechanism the restrict the negative effects of perturbations in the graphs. For instance, RGCN~\cite{zhu2019robust} adopts Gaussian distributions as the hidden representations of nodes in each graph convolutational layer. As a result, the proposed RGCN could absorb the effects of adversarial changes in the variances of the Gaussian distributions. In~\cite{chen2021understanding}, a median aggregation mechanism is designed to improve the robustness of GNNs. In median aggregation, the median value of each dimension of the neighbor embeddings is used to capture the context information. Only when the portion of clean nodes in the neighbors is less than 0.5, the perturbed values will be selected in median aggregation, which implies its benefits to the model robustness. Following~\cite{chen2021understanding}, a soft median aggregation mechanism is applied for scalable defense~\cite{geisler2021robustness}, which computes the weighted mean where the weight of a an entry is based on the distance to the dimension-wise median.  Extensive experiments on large-scale graphs with up to 100M nodes demonstrate the validity and efficiency.

\vspace{0.2em} \noindent \textbf{Self-Supervised Learning Defense Methods.} To address the problem of lacking labels, various self-supervised learning tasks such as link prediction~\cite{kim2020find} and contrastive learning~\cite{you2020graph} have been proposed to help representation learning of GNNs. In addition to the prediction accuracy, it is found that some self-supervised tasks can improve the robustness of the GNNs. For instance, SimP-GCN~\cite{jin2021node} employs a self-supervised similarity preserving task which enforces similar representations for nodes with similar attributes. Therefore, the nodes whose local graph structures are perturbed can still preserve useful representations.
In contrastive learning, maximizing the representation consistency between the original graphs and the augmented views of edge perturbation~\cite{dai2021towards,you2020graph} can also result in a more robust model. Some adversarial graph contrastive learning and variants~\cite{guo2022learning,xu2020unsupervised,feng2022adversarial,wang2022robust,lin2022spectral} are developed to further improve the robustness by introducing an adversarial view of graphs. 

\subsection{Applications of Robust GNNs}
%In this subsection, we introduce the application scenarios of robust GNNs. 
Since robust GNNs can defend against adversarial attacks, applications in safety-critical domains will particularly benefit from robust GNNs. For instance, the investigations in bioinformatics graphs such as  protein-protein network~\cite{yang2020graph} and brain network~\cite{kawahara2017brainnetcnn} require robust GNNs to defend the attacks in bioinformatics~\cite{tang2020adversarial} to guarantee the safety. In addition, recent research also shows the vulnerability of GNNs in knowledge graph modeling~\cite{zhang2019data}. Nowadays, GNNs have been widely applied to learn useful representations from the knowledge graph to facilitate various downstream tasks such as recommendation system~\cite{wang2018dkn}. Therefore, robust GNNs is required for the application of knowledge graphs. Some works have attempted to apply GNNs in Financial analysis such as credit estimation and fraud detection~\cite{wang2021review}. Therefore, robust GNNs are urgent for the security of GNNs in real-world financial analysis.

\subsection{Future Research Directions of Robust GNNs}

\vspace{0.2em} \noindent \textbf{Scalable Robust GNNs.} As discussed in Sec.\ref{sec:scalable_attack}, some initial efforts have verified that adversarial attacks can be applied on extremely large graphs to  achieve the attackers' goal. 
However, scalable defense methods are rather limited~\cite{geisler2021robustness}.
Though some methods such as GCN-Jaccard is efficient, the defense performance is generally not good enough. For more advanced methods such as Pro-GNN, the computation cost is unaffordable for large-scale graphs. Thus, it is an emerging and promising direction to develop scalable robust GNNs.

\vspace{0.2em} \noindent \textbf{Robust GNNs on Heterogeneous Graphs.} Many real-world graphs such as product-user network are heterogeneous, which contain diverse types of objects and relations. Extensive Heterogeneous Graph Neural Networks (HGNNs) have been investigated for heterogeneous graphs. However, recent analysis~\cite{zhang2022robust} also shows that the adversarial attacks bring more negative effects to metapath-based HGNNs than general GNNs. Despite extensive works on robust GNNs, they are dedicated to homogeneous graphs, which can rarely handle heterogeneous graphs. Thus, developing robust HGNNs still remains an open problem.

\vspace{0.2em} \noindent \textbf{Robust GNNs Against Label Noises.}
Existing works mainly focus on defending the adversarial attacks on graph structure and node features; while the noises and attacks on labels such as label-flipping attack~\cite{zhang2020adversarial} can also significantly degrade the performance for GNNs. Several initial efforts~\cite{zhang2020adversarial,dai2021towards,li2021unified} are conducted to address the challenge of label noises.
For instance, the authors in~\cite{dai2021nrgnn} firstly develop a label noise-resistant GNN (NRGNN) by linking the unlabeled nodes with (pseudo) labeled nodes with similar features, which can improve the performance of GNNs against label noise or label flipping attack. 
Though promising, the robust GNNs against label noises is still in an early stage that needs further investigation.

\vspace{0.2em} \noindent \textbf{Robust Pre-training GNNs.} Recently, various pre-training GNN frameworks~\cite{hu2019strategies,you2020graph} have been investigated to leverage the large scale of data for downstream tasks. The adversarial attacks can also be applied to the pre-training GNN. For example, backdoor can be injected to a self-supervised learning GNN to mislead the model give target prediction to the target instance even after the fine-tuning~\cite{xi2021graph}.  Considering the pre-training GNN model will be utilized to various datasets and downstream tasks, a success adversarial attack on pretraining GNNs could cause huge losses. Thus, it is necessary to develop robust pre-training GNNs.
\section{Fairness of Graph Neural Networks} \label{sec:fairness}
Fairness is one of the most important aspects of trustworthy graph neural network. 
With the rapid development of graph neural network, GNNs have been adopted to various applications. However, recent evidence shows that similar to machine learning models on i.i.d data, GNNs also can give unfair predictions due to the societal bias in the data. The bias in the training data even can be magnified by the graph topology and message-passing mechanism of GNNs~\cite{dai2021say}.
For example, recommendation system based on random walk is found to prevent females from rising to the most commented and liked profiles~\cite{rahman2019fairwalk, stoica2018algorithmic}. A similar issue has been found in book recommendation with graph neutral network, where the GNN methods could be biased towards suggesting books with male authors~\cite{buyl2020debayes}.
% \suhang{give examples that the fairness of GNN on the application is important, and what kind of loss it will cause for unfair GNNs.} 
These examples imply that GNNs could be discriminated towards the minority group and hurt the diversity in culture. Moreover, the discrimination would largely limit the wide adoption of GNNs in other domains such as  ranking of job applicants~\cite{mehrabi2021survey} and loan fraud detection~\cite{xu2021towards}, and even can cause legal issues. Therefore, it is crucial to ensure that GNNs do not exhibit discrimination towards users. Hence, many works have emerged recently to develop fair GNNs to achieve various types of fairness on different tasks. In this section, we will give a comprehensive review of the cutting-edge works on fair GNNs. Specifically, we first introduce the major biases that challenge fairness of GNNs. We then describe various concepts of fairness that are widely adopted in literature, followed by categorization and introduction of methods for achieving fairness on graph-structured data. Finally, we present public datasets and applications and discuss future directions that need further investigation.

% Some background
\subsection{Bias Issues in Graph-Structured Data and Graph Neural Networks}
Biases widely exist in real-world datasets, which can lead to unfair predictions of machine learning models. Olteanu \textit{et al.} listed various biases that exist in social data~\cite{olteanu2019social}. Suresh \textit{et al.} further discussed various types of biases that cause discrimination issues of machine learning models~\cite{suresh2019framework}. According to~\cite{mehrabi2021survey}, the bias in machine learning can appear in different stages such as data, algorithm and user interaction. In this paper, we mainly focus on biases in graph-structured data and on GNNs. For a comprehensive review of biases that occur in other phases such as training and evaluation of machine learning models on i.i.d data, please refer to the survey~\cite{mehrabi2021survey}. 

First, similar to i.i.d data, node attributes/features are often available in graph-structured data. In addition, the data collection of graph-structured data follows similar procedures to i.i.d data such as data sampling and label annotation. Thus, the following biases that widely exist on i.i.d data also exist in graphs.
\begin{itemize}[leftmargin=*]
    \item \textbf{Historical Bias}. 
    Various biases such as gender bias and race bias exist in the real world due to historical reasons. These biases can be embedded and reflected in the data. For a system that reflect the world accurately, it can still inflict harm on a population who experience the historical bias~\cite{suresh2019framework}. An example of this type of bias is the node embedding learning for link prediction~\cite{rahman2019fairwalk}.
    In particular, users with the same sensitive attributes such as gender and race are more likely to be linked in a real-world graph. As a result, the learned node embeddings will tend to link users with the same gender/race. Then, applications such as friend recommendation that built on these types of node embeddings will reinforce the historical bias. 
     
    \item \textbf{Representation Bias.} \textit{Representation bias occurs when the collected samples under-represent some part of the population, and subsequently fails to generalize well for a subset of the use population}~\cite{suresh2019framework}. The representation bias can be caused in several ways: (i) the target population does not reflect the use population; (ii) the target population under-represent certain groups; and (iii) the sampled data does not reflect the target population.
    \item \textbf{Temporal Bias}. \textit{Temporal bias arises from differences in populations and behaviors over time~\cite{olteanu2019social}}. For graphs, temporal bias can be caused by both the change of node attributes and graph topology. One example is the social network, where the attributes and links of users will evolve over time.
    \item \textbf{Attribute Bias}. \textit{Given an attributed network and the corresponding group indicator (w.r.t. the sensitive attribute) for each node. For any attribute, if the distributions between different demographic groups are different, then attribute bias exists in graph~\cite{dong2021edits}.} Attribute bias focuses on the biases in the node attributes of the graphs.
\end{itemize}
In addition to the aforementioned biases, there are unique types of biases in graph-structured data due to the graph topology:
\begin{itemize}[leftmargin=*]
    \item \textbf{Structural Bias}. \textit{Given an undirected attributed network and the corresponding group indicator (w.r.t. sensitive attribute) for each node. If any information propagation promotes the distribution difference between different groups at any attribute dimension, then structural bias exists in the graph~\cite{dong2021edits}.}
    \item \textbf{Linking Bias}. \textit{Linking bias arises when network attributes obtained from user connections, activities, or interactions differ and misrepresent the true behavior of the users~\cite{olteanu2019social}}. For instance, it is found that younger people are more closely connected than older generations on social network~\cite{dong2016young}. Moreover,  low-degree nodes are more likely to be falsely predicted~\cite{tang2020investigating}, which leads to degree-related bias.
\end{itemize}

Generally, GNNs adopt a message-passing mechanism, which aggregates the information of neighbors to enrich the representation of the target nodes. As a result, the learned representations can capture both node attributes and its local topology, which can facilitate various tasks such as node classification. However, due to the biases in topology, \textit{the message passing mechanism of GNNs can magnify the biases} compared with MLP~\cite{dai2021say}. In graphs such as social networks, nodes of similar sensitive attributes are more likely to connect to each other~\cite{dong2016young,rahman2019fairwalk}. For example, young people
tend to build friendship with people of similar age on the social
network~\cite{dong2016young}. The message passing in GNNs will aggregate the neighbor features. Thus, GNNs learn similar representations for nodes of similar sensitive information while different representations for nodes of different
sensitive features, leading to severe bias in decision making, i.e.,
the predictions are highly correlated with the sensitive attributes of the nodes.  

\subsection{Fairness Definitions} \label{sec:fair_definition}
In this subsection, we introduce the most widely used fairness definitions, which can be generally split into two categories, i.e., group fairness and individual fairness.

\subsubsection{Group Fairness} The principle of group fairness is to ensure groups of people with different protected sensitive attributes receive comparable treatments statistically. Various criteria of group fairness have been proposed. Next, we will introduce the mostly used group fairness definitions.

\begin{definition}[Statistical Parity \cite{dwork2012fairness}] Statistical parity, also known as demographic parity, requires the prediction $\hat{y}$ to be independent with the sensitive attribute $s$, i.e., $\hat{y} \bot s$. The majority of the literature focus on binary classification and binary attribute, i.e., $y \in \{0,1\}$ and $s \in \{0,1\}$. In this situation, statistical parity can be formally written as:
\begin{equation}
    P(\hat{y}=1|s=0)=P(\hat{y}=1|s=1).
    \label{eq:SP}
\end{equation}
\end{definition}
\noindent According to statistical parity, the membership in the protected sensitive attributes should have no correlation with the decision from the classifier. %And it can be achieved by learning a classifier following the constraint Eq.(\ref{eq:SP}).  
Given the definition in Eq.(\ref{eq:SP}), the fairness in terms of statistical parity can be measured by:
\begin{equation}
   \Delta_{SP} =  |P(\hat{y}=1|s=0)-P(\hat{y}=1|s=1)|.
\end{equation}
A lower $\Delta_{SP}$ indicates a more fair classifier. The statistical parity can be easily extended to multi-class and multi-category sensitive attributes problem by ensuring $\hat{y} \bot s$. According to~\cite{locatello2019fairness}, let $y \in \{y_1,\dots,y_c\}$ and $\mathbf{s} \in \{s_1,\dots,s_{k}\}$ denotes the multi-class label and multi-category sensitive attribute, where $c$ is number of classes and $k$ is number of sensitive attribute categories, the evaluation metric can be extended to:
\begin{equation}
    \Delta_{SP} = \frac{1}{k} \sum_{i=1}^{k} \max_{y_j}{|P(\hat{y}=y_j)-P(\hat{y}=y_j|s=s_i)|}.
    \label{eq:SP_metric}
\end{equation}

Statistical parity is the first fairness definition and have been widely adopted. However, some following works~\cite{hardt2016equality} argue that statistical parity often cripples the
utility of the model. Hence, Equalized Odd is proposed to alleviate the issue, which is defined as:
\begin{definition}[Equalized Odds~\cite{hardt2016equality}] A predictor satisfies equalized odds with respect to protected attribute $s$ and class label $y$, if the prediction $\hat{y}$ and $s$ are independent conditioned on $y$, i.e., $\hat{y} \bot s \|y$. When $y \in \{0,1\}$ and $s \in \{0,1\}$, this can be formulated as:
\begin{equation}
    P(\hat{y}=1|y=v,s=0) = P(\hat{y}=1|y=v,s=1),
    ~~  \forall v\in \{0,1\}.
\end{equation}
\end{definition}
\noindent Equalized odds enforces that the accuracy is equally
high in all demographics, punishing models that perform well only on the majority.  In the binary classification, we often set $y = 1$ as the “advantaged” outcome, such as “not defaulting on a loan” or “admission to a college”. Hence, we can relax the equalized odds to achieve fairness within the “advantaged” outcome group, which is known as Equal Opportunity.
\begin{definition}[Equal Opportunity \cite{hardt2016equality}]
It requires that the probability of an instance in a positive class being assigned to a positive outcome should be equal for both subgroup members, i.e.,
 \begin{equation}
     P(\hat{y}=1|y=1,s=0) = P(\hat{y}=1|y=1,s=1). 
 \end{equation}
\end{definition}
\noindent Equal opportunity expects the classifier to give equal true positive rates across the subgroups, which allows a perfect classifier. Similar to statistical parity, equal opportunity can be measured by
\begin{equation}
    \Delta_{EO} = |P(\hat{y}=1|y=1,s=0) - P(\hat{y}=1|y=1,s=1)|.
\end{equation}
Equalized odds and equal opportunity can be naturally extended to multi-class and multi-category sensitive attributes setting by changing the range of sensitive attributes and labels.

\begin{definition}[Dyadic Fairness~\cite{li2020dyadic}]
This can be viewed as an extension of statistical parity for link prediction. It requires the link predictor to give predictions independent with the sensitive attributes of the target nodes. A link prediction algorithm satisfies dyadic fairness if the predictive score satisfies:
\begin{equation}
    P(g(u,v)|s(u)=s(v))=P(g(u,v)|s(u)\neq s(v)),
    \label{eq:dyadic}
\end{equation}
where $g(\cdot)$ is the link predictor, $s(u)$ and $s(v)$ denote the sensitive attributes of node $u$ and $v$, respectively. 
\end{definition}
Since the dyadic fairness is extended from the link prediction, the evaluation metric can be simply extended from $\Delta_{SP}$ in Eq.(\ref{eq:SP_metric}) by replacing the classification probability to link prediction probability, i.e., $\Delta_{DE} = | P(g(u,v)|s(u)=s(v)) - P(g(u,v)|s(u)\neq s(v))|$

% \suhang{any discussion on how to extend to multi-class and multi-category attribute? discussion on how to choose which measure to use for which problem? why we are interested in giving these definitions: design regularizer and evaluation metrics. Introduce {\bf evaluation metrics}}

\subsubsection{Individual Fairness}
While group fairness can maintain fair outcomes for a group of people, a model can still behave discriminatorily at the individual level. Individual fairness is based on the understanding that similar individuals should be treated similarly.
\begin{definition}[Fairness Through Awareness~\cite{dwork2012fairness}]
Any two individuals who are similar should receive similar algorithmic outcome. Let $u,v \in \mathcal{X}$ be two data points in dataset $\mathcal{X}$, and $f(\cdot)$ denotes a mapping function. The fairness through awareness can be formulated as:
\begin{equation}
    D(f(u),f(v)) \leq d(u,v),
\end{equation}
where $D(\cdot)$ and $d(\cdot)$ are two distance metrics required to be defined in the application context.
\end{definition}

% \begin{definition}[Individual Fairness from a Ranking Perspective]
% Given the oracle pairwise similarity matrix $\mathbf{S}_{\mathcal{G}}$ of the input
% graph $\mathcal{G}$, and the similarity matrix among instances in the outcome space (defined upon a similarity metric), we say the predictions
% are individually fair if for each instance, the two ranking lists that
% encode the relative order of other instances (ranked based on the similarity between instance and other instances in descending order)
% from $\mathbf{S}_{\mathcal{G}}$ and $\mathbf{S}_{\hat{Y}}$ are consistent with each other \enyan{An extension of fairness through awareness, may not required to be listed here?}\suhang{if it is not important, you can just explain without formally defining it}.
% \end{definition}
\begin{definition}[Counterfactual Fairness~\cite{kusner2017counterfactual}] The counterfactual fairness enforces that predictions for an individual in real-world should remain unchanged in a counterfactual world where the individual’s protected attributes had been different. Let $\hat{Y}_{S\leftarrow s}(U)$ and $\hat{Y}_{S \leftarrow s{'}}(U)$ denote the predictions of a sample with background variable $U$ whose sensitive attributes are set as $s$ and $s{'}$, respectively.
A predictor is counterfactually fair if under any context $X=x$ and $S=s$:
\begin{equation}
    P(\hat{Y}_{S\leftarrow s}(U)=y|X=x,S=s)=P(\hat{Y}_{S\leftarrow s{'}}(U)=y|X=x, S=s),
\end{equation}
for all $y$ and any value $s'$ of protected attribute $S$. 
\end{definition}
% The intuition behind the conterfactual fairness is that a fair predictor should give the same prediction to an individual in the actual world and the counterfactual world where the individual belongs to other sensitive attribute groups. 
% \subsection{Group Fairness}
The counterfactual fairness can be viewed as individual fairness whose similarity metric treats the individual and its counterfactual sample as a similar pair. How to evaluate individual fairness remains an under-explored research direction. In~\cite{kang2020inform}, a measure of individual fairness is proposed. The metrics of group fairness such as $\Delta_{SP}$ are also utilized for evaluating the counterfactually fair GNNs~\cite{agarwal2021towards}.

\subsection{Fairness-Aware Graph Neural Networks}
\begin{table}[t]
    \scriptsize
    \centering
    \caption{Categorization of fair models on graphs according to the revised stage.}
    \vskip -1.5em
    \begin{tabular}{ll}
    \toprule
        Category &  References  \\
        \midrule
        Pre-processing & \cite{kang2020inform}, \cite{dong2021edits}, \cite{spinelli2021biased} \\
        \midrule 
        In-processing & \cite{dai2021say},~\cite{bose2019compositional},~\cite{masrour2020bursting},~\cite{dong2021edits}, ~\cite{wang2021unbiased},~\cite{agarwal2021towards},~\cite{li2020dyadic},~\cite{kang2020inform},~\cite{dong2021individual},~\cite{buyl2020debayes},~\cite{kose2021fairness},~\cite{khajehnejad2021crosswalk},~\cite{rahman2019fairwalk}
        \\
        \midrule
        Post-processing & \cite{kang2020inform} \\
        \bottomrule
    \end{tabular}
    \label{tab:fair_category}
\end{table}

Extensive attempts have been proposed to eliminate the discrimination in machine learning models on i.i.d data~\cite{mehrabi2021survey}.  However, these methods cannot be directly applied to graph-structured data because of the unique biases brought by the graph topology and message passing mechanism. Recently, with the remarkable success of GNNs, the concern on fairness issue of GNNs is attracting increasing attention. In this section, we introduce the  debiasing methods for achieving fairness in GNNs.
Following the categorization of fair machine learning algorithms on i.i.d data~\cite{mehrabi2021survey,liu2021trustworthy}, existing fairness-aware algorithms can be split into \textit{pre-processing methods}, \textit{in-processing methods}, and \textit{post-processing methods}, based on which stage the debiasing is conducted. Pre-processing approaches are applied to eliminate the bias in data with fair pre-processing methods. In-processing approaches are designed to revise the training of machine learning models to ensure the predictions meet the target fairness definition. Post-processing methods directly change the predictive labels to ensure fairness.
Table~\ref{tab:fair_category} lists existing works on Fair GNNs into these three categories. Based on the techniques they apply, we categorize the debiasing methods on graph-structured data into \textit{Adversarial Debiasing}, \textit{Fairness Constraints}, and others. Next, we introduce the details of the methods following the categorization based on the techniques.

\subsubsection{Adversarial Debiasing} 
\begin{figure}[t]
\centering
\begin{subfigure}{0.49\linewidth}
    \centering
    \includegraphics[width=0.9\linewidth]{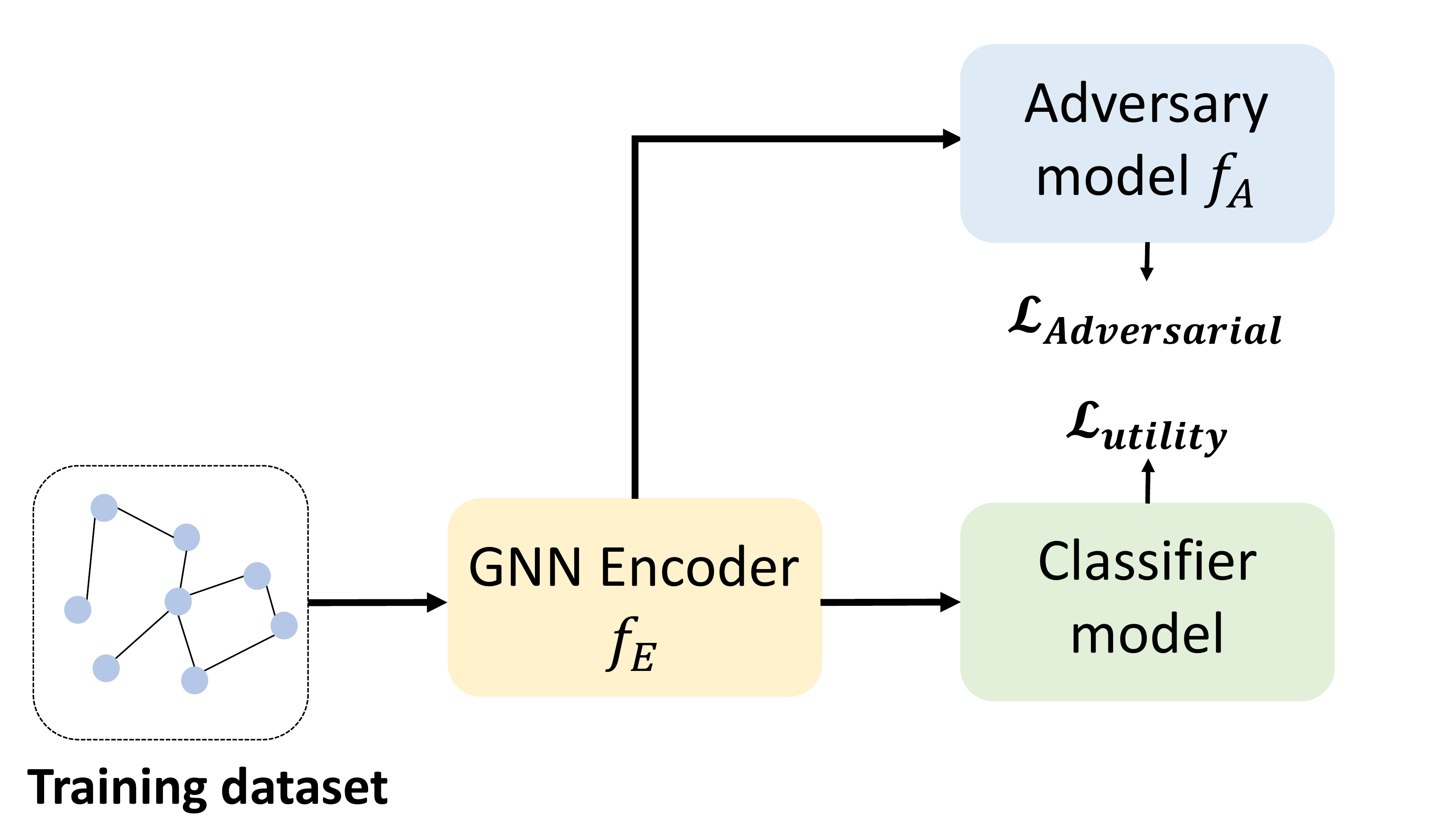}
    % \vskip -1em
    \caption{Framework of Adversarial Debiasing models.}
    \label{fig:adv_debias}
\end{subfigure}
\begin{subfigure}{0.49\linewidth}
    \centering
    \includegraphics[width=0.9\linewidth]{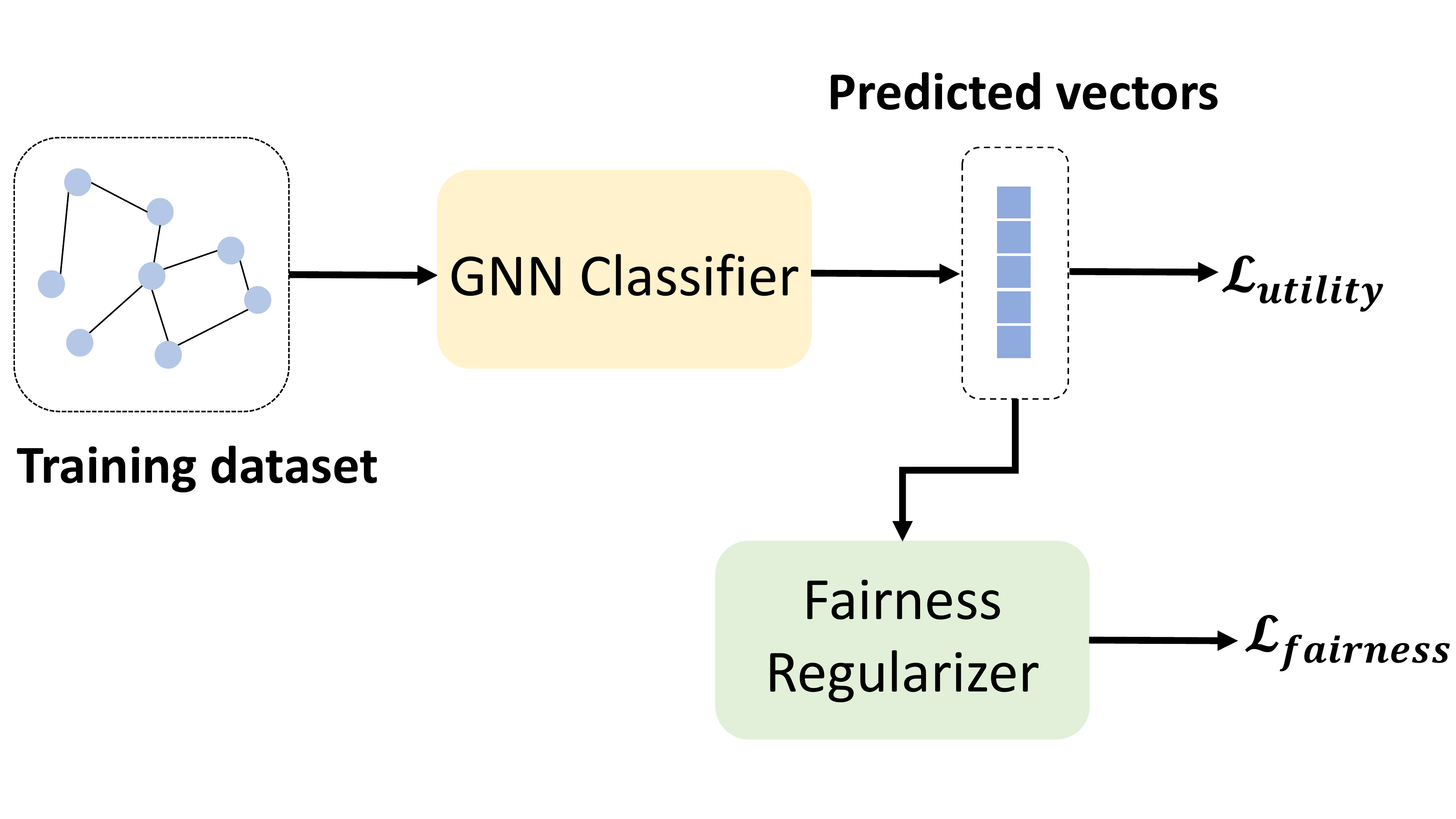}
    % \vskip -1em
    \caption{Framework of Fairness Constraint models.}
    \label{fig:fair_cons}
\end{subfigure}
\vskip -1em
\caption{An illustration of fairness-aware GNNs.}
\vskip -1em
\end{figure}

Using adversarial learning~\cite{goodfellow2014generative} to eliminate the bias is firstly investigated in the fair machine learning models on i.i.d data~\cite{beutel2017data,edwards2015censoring,liao2019learning,madras2018learning}. Several efforts~\cite{dai2021say,bose2019compositional,masrour2020bursting,dong2021edits} have been taken to extend the adversarial debiasing on graph-structured data. An illustration of adversarial debiasing is presented in Figure~\ref{fig:adv_debias}. Generally, adversarial debiasing adopts an adversary $f_A$ to predict sensitive attributes from the representations $\mathbf{H}$ of an encoder $f_E$; while the encoder aims to learn representations that can fool the adversary while can give accurate predictions for the task at hand, say node classification.
With the minmax game, the final learned representations will contain no sensitive information, resulting in fair predictions that independent with the sensitive attributes. Thus, statistical parity or dyadic fairness can be guaranteed with the adversarial biasing in node classification and link prediction, respectively.  The objective function of adversarial debiasing can be formulated as
\begin{equation}
    \min_{\theta_E} \max_{\theta_A} \mathcal{L}_{utility}(f_E(\mathcal{G};\theta_E)) - \beta \mathcal{L}_{Adversarial}(f_A(\mathbf{H};\theta_A)), 
\end{equation}
where $\theta_E$ and $\theta_A$ are parameters of encoder $f_E$ and adversary $f_A$, respectively. $\mathcal{L}_{utility}$ is the loss function to ensure the utility of the learned representations such as node classification loss and link reconstruction loss. $\mathcal{L}_{Adversarial}$ is the adversarial loss, which generally is cross entropy loss of sensitive attribute prediction of the adversary based on the learned node representations $\mathbf{H}$. $\beta$ is the hyperparameter to balance the contributions of these two loss terms.

\begin{table}[t]
    \scriptsize
    \centering
    \caption{Fair graph neural networks that adopt adversarial debiasing.}
    \vskip -1.5em
    \begin{tabular}{lllll}
    \toprule
    Methods  & Task & Fairness & $\mathcal{L}_{Adversarial}$ &  \\
    \midrule
    FairGNN~\cite{dai2021say} & Node Classification &  Statistical Parity & Cross entropy between $\hat{s}$ and $\Tilde{s}$  \\
    \midrule
    EDITS~\cite{dong2021edits}& Node Classification & Statistical Parity & Wasserstein distance \\
    \midrule
    \multirow{2}{*}{Compositional~\cite{bose2019compositional}} & Node classification & Statistical Parity &  \multirow{2}{*}{Cross entropy between $\hat{s}$ and $s$}\\
    & Link prediction &  Dyadic fairness &  \\
    \midrule
    FLIP~\cite{masrour2020bursting}& Link Prediction & Dyadic fairness &  Cross entropy between $\hat{s}$ and $s$ \\
    \bottomrule
    \end{tabular}

    \label{tab:debias}
    \vskip -1em
\end{table}

The seminar work~\cite{bose2019compositional} firstly applies adversarial debiasing on graph-structured data to learn fair node embeddings. Specifically, it adopts adjacency matrix reconstruction with negative sampling as the utility loss to learn node embeddings. Let $g(v_i,v_j)$ be the predicted probability that $v_i$ and $v_j$ are linked. The utility loss can be written as: $\mathcal{L}_{utility} = \sum_{v_i \in \mathcal{V}} \sum_{v_j \in \mathcal{N}(v_i)} -[\log(g(v_i,v_j)) + \sum_{n=1}^{Q} \mathbb{E}_{v_n\sim P_n(v_i)} \log(1-g(v_i,v_n))]$, where $\mathcal{N}(v)$ represents the neighbors of node $v$ and $P_n(v_i)$ is the distribution of sampling negative nodes for $v_i$. $Q$ is number of negative samples. 
An MLP is deployed as the adversary $f_A$ to predict the sensitive attributes from the node embeddings $\mathbf{H}$. The adversarial loss is given as the binary cross entropy loss between the predictions from the adversary and the real sensitive attributes, i.e., $\mathcal{L}_{Adversarial}=\sum_{v\in \mathcal{V}_S} -[s_v \log(\hat{s}_v) + (1-s_v)\log (1-\hat{s}_v)]$.
Aasrour \textit{et al.}~\cite{masrour2020bursting} investigate adversarial debiasing with similar implementation of losses to learn representations for fair link prediction to avoid the separation of users. In addition, they propose a metric to determine whether the predicted links will lead to further separation of the network to evaluate the model. 

Though the aforementioned methods achieve fairness with adversarial debiasing, they focus on learning node embeddings and do not use GNN model as the encoder. Recently, FairGNN~\cite{dai2021say} proposes a framework for fair node classification with graph neural networks. It uses the node classification loss as the utility loss, i.e., $\mathcal{L}_{utility} = \sum_{v \in \mathcal{V}_L} -[y_v\log(\hat{y}_v) + (1-y_v)\log(1-\hat{y}_v))$ where $\mathcal{V}_L$ is the set of labeled nodes. %Moreover, a more realistic setting of learning fair GNNs with limited number of sensitive attributes is investigated in~\cite{dai2021say}.  
For many real-world applications such as user attribute prediction in social medium, obtaining sensitive attributes of nodes for is difficult. To address the challenge of lacking sensitive attributes for adversarial debiasing, FairGNN adopts a GCN-based sensitive attribute estimator to estimate sensitive attributes $\Tilde{s}$ for nodes with missing sensitive attributes. It then uses the output of adversary $f_A$ and the estimated sensitive attribute $\Tilde{s}$
for adversarial loss. In addition, theoretical analysis in~\cite{dai2021say} demonstrates that the statistical parity can be guaranteed with adversarial debiasing given the estimated sensitive attributes. 

All the aforementioned methods focus on debiasing the node representations. Alternatively, adversarial debiasing can also be applied to debias the original graph data. For example, EDITS~\cite{dong2021edits} utilize a WGAN-based~\cite{arjovsky2017wasserstein} framework to eliminate the attribute bias and structural bias. The attribute matrix $\mathbf{X}$ is revised to ensure same attribute distribution between different demographic groups. Similarly, an adjacency matrix $\mathbf{A}$ that will not cause bias after information propagation is learned. More details of the representative adversarial debiasing methods are listed in Table~\ref{tab:debias}.
\begin{table}[t]
    \scriptsize
    \centering
    \caption{Fair graph neural networks using fairness constraints.}
    \vskip -1.5em
    \begin{tabular}{llll}
        \toprule
        Method & Task & Fairness  &  $\mathcal{L}_{fairness}$  \\
        \midrule
        FairGNN~\cite{dai2021say} & Node classification & Statistical Parity  & $|Cov(\hat{y},\hat{s})|$ \\
        \midrule
        UGE~\cite{wang2021unbiased} & Node classification & Statistical Parity &  $|g(\mathbf{x}_u,\mathbf{x}_v)-g(\Tilde{\mathbf{x}}_u,\Tilde{\mathbf{x}}_v)|$\\
        \midrule
        FairAdj~\cite{li2020dyadic} & Link prediction & Dyadic fairness & $|\mathbb{E}[g(u,v)|s_u=s_v]-\mathbb{E}[g(u,v)|s_u \neq s_v]| $ \\
        \midrule
        InFoRM~\cite{kang2020inform} & \makecell{PageRank\\Spectrual Clustering \\Embedding Learning } & Individual Fairness & $\sum_i \sum_j \mathbf{S}_{ij} (\mathbf{y}_i - \mathbf{y}_j)^2$ \\
        \midrule
        REDRESS~\cite{dong2021individual} & Node \& Link  & Individual Fairness & Consistency between $\mathbf{S}_{\mathcal{G}}$ and $\mathbf{S}_{\hat{Y}}$ \\
        \midrule
        NIFTY~\cite{agarwal2021towards} & Node classification & Counterfactual fairness & $|f(\Tilde{u})-f(u)|$ 
        \\
        \bottomrule
    \end{tabular}
    \label{tab:fc}
    \vskip -1em
\end{table}

\subsubsection{Fairness Constraints}
In addition to adversarial debiasing, directly adding fairness constraints to the objective function of machine learning modes is another popular direction. These constraints are usually derived from the fairness definitions introduced in Section~\ref{sec:fair_definition}. As the general framework of fairness constraints in Figure~\ref{fig:fair_cons}, they work as the regularization term and balance the  performance in prediction and fairness. The overall objective function can be written as
\begin{equation}
    \min_{\theta} \mathcal{L}_{utility} + \beta \mathcal{L}_{fairness}, 
\end{equation}
where $\theta$ is the set of model parameters to be learned, $\mathcal{L}_{utility}$ is the loss function for the utility of the model, $\mathcal{L}_{fairness}$ denotes the applied fairness constraint, and $\beta$ controls the trade-off between utility and fairness. 
To enforce different notion of fairness, various constraints have been investigated for fair GNNs~\cite{dai2021say,wang2021unbiased,agarwal2021towards,li2020dyadic,kang2020inform,dong2021individual}. The details of these methods are given in Table~\ref{tab:fc}.

\vspace{0.2em} \noindent \textbf{Statistical Parity \& Dyadic Fairness.}
In FairGNN~\cite{dai2021say}, apart from the adversarial debiasing, it also adopts the covariance constraint for statistical parity to further enforce fairness. Specifically, the covariance constraint minimizes the absolute covariance between the estimated sensitive attribute $\Tilde{s}$ and predictions $\hat{y}$, i.e., $|Cov(\Tilde{s},\hat{y})|=|\mathbb{E}[(\Tilde{s}-\mathbb{E}(\Tilde{s}))(\hat{y}-\mathbb{E}(\hat{y}))]|$. Enforcing the predictions to have no correlation with the estimated sensitive attributes will be helpful to learn classifier that give predictions independent with the protected attributes. 
UGE~\cite{wang2021unbiased} assumes that a bias-free graph can be generated from the pre-defined non-sensitive attributes. Then, a regularization term  pushes the embeddings to satisfy properties of the bias-free graph to eliminate bias. 
In particular, they enforce $g(\mathbf{x}_u,\mathbf{x}_v)$, i.e. the probability of predicting links between nodes $u$ and $v$ with complete attributes, is the same as $g(\Tilde{\mathbf{x}}_u,\Tilde{\mathbf{x}}_v)$, i.e., the probability of prediction links between $u$ and $v$ with bias-free attributes.
In FairAdj~\cite{li2020dyadic}, a regularization term, $|\mathbb{E}[g(u,v)|s_u=s_v]-\mathbb{E}[g(u,v)|s_u \neq s_v]|$, that directly derived from dynamic fairness based on Eq.(\ref{eq:dyadic}) is used to debias the adjacency matrix. 
% \suhang{the description is actually unclear. for example, the covariance constraint is add on what. you can give a little bit more details, say write down the cosntraint and explain the intuition, explain what kind of fairness it achieves and why. Note that this actually connect back to the fairness definitions we give in previous subsections}. 

\vspace{0.2em} \noindent \textbf{Individual Fairness.}
Moreover, two constraints for individual fairness are explored in InFoRM~\cite{kang2020inform} and REDRESS~\cite{dong2021individual}. InFoRM~\cite{kang2020inform} proposed a regularization term $\sum_i \sum_j \mathbf{S}_{ij} (\mathbf{y}_i - \mathbf{y}_j)^2$, where $\mathbf{y}_i \in \mathbb{R}^{c}$ and $\mathbf{y}_j \in \mathbb{R}^{c}$ denote the prediction vectors of nodes $v_i$ and $v_j$, and $\mathbf{S}_{ij} \in [0, 1]$ is the similarity score between $v_i$ and $v_j$. In this way, for two similar nodes, their predictions will be encouraged to be similar. As for REDRESS~\cite{dong2021individual}, it aims to optimize the consistency between the prediction similarity matrix $\mathbf{S}_{\hat{Y}}$ and the oracle similarity matrix $\mathbf{S}_{\mathcal{G}}$ from a ranking perspective.  For each node, the relative order of each node pair by $\mathbf{S}_{\hat{Y}}$ and that by $\mathbf{S}_{\mathcal{G}}$ are enforced to be the same.

\vspace{0.2em} \noindent \textbf{Counterfactual Fairness.} To achieve counterfactual fairness, NIFTY~\cite{agarwal2021towards} proposes to maximize the agreement between the original graph and its counterfactual augmented views. More specifically, the counterfactual sample $\Tilde{u}$ of an initial sample $u$ is generated by (i)  modifying the value of sensitive attribute; and (ii) randomly masking the other attributes and perturb the graph structure. Then, the constraint described in Table~\ref{tab:fc} can be applied to reduce the the gap between the predictions on the original graph and its counterfactual samples, which will enable the counterfactual fairness. Our survey~\cite{guo2023counterfactual} about counterfactual learning give more details about Counterfactual Fairness. 

\begin{table}[t]
    \footnotesize
    \centering
    \caption{Fair models on graphs belonging to other categories}
    \vskip -1 em
    \begin{tabular}{lll}
    \toprule
    Methods  & Task & Fairness   \\
    \midrule
    FairWalk~\cite{rahman2019fairwalk} & Link Prediction & Statistical parity \\
    CrossWalk~\cite{khajehnejad2021crosswalk} & Node \& Link & Statistical parity\\ 
    FairDrop~\cite{spinelli2021biased} & Node Classification & Statistical parity  \\
    Debayes~\cite{buyl2020debayes} & Link Prediction & Statistical parity \\
    % Fair PageRank~\cite{tsioutsiouliklis2021fairness}\\
    GMMD~\cite{zhu2023fairness} & Node Classification & Statistical parity\\
    \bottomrule
    \end{tabular}
    \vskip -1em
    \label{tab:fair_other}
\end{table}

\subsubsection{Fairness-aware GNNs in Other Categories}
Apart from aforementioned fair GNNs, there are several methods that do not belong to the adversarial debiasing or fairness constraints, which are presented in Table~\ref{tab:fair_other}. More specifically, users in the same sensitive attribute group are  more likely be sampled into a trace with the generally random walk, resulting to the correlation between the sensitive attributes and predictions. Therefore, unbiased sampling strategies in the random walk are investigated in~\cite{rahman2019fairwalk,khajehnejad2021crosswalk} to learn unbiased embeddings for downstream tasks such as node classification and link prediction. FairDrop~\cite{spinelli2021biased} proposes to drop more connections between nodes sharing the same sensitive attribute to reduce the bias of homophily in sensitive attributes. Debayes~\cite{buyl2020debayes} investigates a Bayesian method that is capable of learning debiased embeddings by using a biased prior. GMMD~\cite{zhu2023fairness} designs a fairness-awareness message-passsing mechanism that will encourage a node to aggregate representations of other nodes from diverse sensitive groups, resulting in fair representations.

\begin{table}[h]
    \footnotesize
    \centering
    \caption{Recent advances in fair graph neural networks.}
    \vskip -1em
    \begin{tabular}{lll}
    \toprule
    Catergory & Task & Reference \\
    \midrule
    \multirow{2}{*}{Fair Augmentation View Methods}    
    & Node Classification & \cite{agarwal2021towards},~\cite{kipf2016variational}, ~\cite{guo2023towards}, ~\cite{wang2022improving}    \\
    & Contrastive Learning & \cite{kose2023demystifying},~\cite{kose2021fairness},~\cite{ling2022learning}\\
    \midrule
    Explanation-Enhanced Fairness & Node Classification & \cite{dong2022structural},~\cite{dong2023interpreting}\\
    \bottomrule
    \end{tabular}
    \label{tab:fair_recent}
\end{table}

\subsubsection{Recent Advances} In addition to the previously discussed approaches adopting adversarial debiasing and fairness constraints, there are methods emerging in fair augmentation views and enhancement of fairness through model explanations, which are listed in Tab.~\ref{tab:fair_recent}. Next, we introduce these cutting-edge categories of fair GNNs in details. 

\vspace{0.2em} \noindent \textbf{Fair Augmentation View for Prediction and Contrastive Learning}. The main idea of the fair augmentation is to generate fairness-aware augmentation views and enforce the agreement between the fair views and original graphs. In this way, the learned representations will achieve fairness and whilst maintaining useful information from original graphs. 
NIFTY~\cite{agarwal2021towards} is one of the earliest methods that fail in this category. Specially, counterfactual views of graphs are generated by randomly perturbations on node attributes, sensitive attributes, and graph structures. Then, similarity between the original graph and its counterfactual fair augmented representations is maximized to ensure fairness for node classification. Similar fairness-aware augmented views are further extended to graph contrastive learning for fair node representation learning~\cite{kose2023demystifying,kose2021fairness}. 

Recently, further improvements have been made in the fair augmentation view generation~\cite{guo2023towards,ma2022learning,wang2022improving,ling2022learning}. For example, GEAR~\cite{ma2022learning} considers the sensitive information in the latent space and uses a GraphVAE~\cite{kipf2016variational} to model the potential biases from neighboring nodes and the overall graph structure. Through counterfactual data augmentation, GEAR generates perturbed data based on sensitive attributes, aiming to minimize discrepancies between original and counterfactual representations. CAF-GNN~\cite{guo2023towards} using a matching-based method to find potential views with different sensitive attributes and labels. Thus, the additional views can be more realistic. Then, CAF-GNN derives proper constraints to ensure the fairness and completeness of the learned representation.

\vspace{0.2em} \noindent \textbf{Explanation-Enhanced Fairness}. It has been empirically proven that utilizing graph structure with the messaging-passing mechanism in GNNs can yield bias in predictions~\cite{dai2021say}. Gaining insights into the origins of such discriminatory predictions is very useful in developing strategies to mitigate biases in GNNs. There are several initial efforts~\cite{dong2022structural,dong2023interpreting} in explaining source of model biases, aiming to enhance the fairness of GNNs through the interpretations to unfairness. Next, we present the existing two works of explanation-enhanced fairness in more details:
\vspace{0.2em}
\begin{itemize}[leftmargin=*, nolistsep]
    \item \textbf{REFEREE}~\cite{dong2022structural}: While biased GNN predictions can arise from various factors, the biased network topology plays a pivotal role in originating and amplifying the discrimination of GNNs. Hence, REFEREE focuses on explaining the unfairness by exploring the edges that maximally account the bias. Let $P(\mathcal{\hat Y}_0)$ and $P(\mathcal{\hat Y}_1)$ denote the distributions of predictions on group 0 and group 1, respectively. The problem of finding biased edge set in node $v_i$'s computation graph can be formulated as $\max_{\mathcal{E}_i} W_1(P(\mathcal{\hat Y}_0),P(\mathcal{\hat Y}_1))$, where $W_1$ measures the Wasserstein-1 distance. Similarly, a fair edge set can be found by minimizing the same loss function. Some other regularization terms are applied for higher quality of obtained biased and fair edge sets. Fair predictions can be given by removing the biased edges while keeping the fair edges.
    \item \textbf{BIND}~\cite{dong2023interpreting}: This approach aims to ascertain the extent to which a GNN model's bias is influenced by the presence of a specific training node within the graph. Specifically, BIND use the influential function to compute the change in the GNN parameters, denoted as $\Delta \mathbf{W}$, when node $v_i$ removed during the training. With the $\Delta \mathbf{W}$, the contribution of node $v_i$ to the predictive bias can be quantified. can be computed. Then, the fairness can be improved by eliminating the nodes that contribute most to the model biases. 
\end{itemize}

\subsection{Datasets for Fair GNNs}
Generally, graph datasets utilized to evaluate the performance of GNNs in terms of fairness need to (i) exhibit bias issue; and (ii) have both node label and node sensitive attributes available if the task is node classification.
Below, we list some of the widely used datasets that are suitable for evaluating the performance of fair GNNs for node classification and/or link prediction problems. The statistics of the datasets along with  papers using these datasets are presented in Table~\ref{tab:fair_datasets}.

\begin{itemize}[leftmargin=*, nolistsep]
    \item \textbf{Pokec-n \& and Pokec-z}~\cite{dai2021say}: Pokec-n and Pokec-z datasets collect users' data from Pokec social networks of two provinces in Slovakia in 2012, which is similar to Facebook. Each node in the graph contains attributes such as gender, age, hobbies, interest, education, working field and etc. The datasets target at predicting the occupations of users and the sensitive attribute is region.
    % \item \textbf{Pokec-z:} The Pokec-z dataset is similar to Pokec-n dataset and it's also from the same social network. It contains the same node attributes and sensitive attributes. The classification task is also to predict the working field of the users.
    \item \textbf{NBA}~\cite{dai2021say}: The NBA dataset utilizes 400 NBA players and their social relations on Twitter to construct the graph. The performance statistics of players in the 2016-2017 season and other information e.g., nationality, age, and salary are provided. The task is to predict whether the players' salary is over median. The sensitive attribute for this dataset is nationality, which is binarized to two categories, i.e., U.S. players and oversea players.
    \item \textbf{German Credit}~\cite{agarwal2021towards}: The German Credit dataset collects data from a German bank~\cite{asuncion2007uci}.  Nodes in the graph represent clients and edges are built between clients if their credit accounts are similar. With clients gender as the sensitive attribute, the node classification task aims to predict whether the credit risk of the clients is high. 
    \item \textbf{Recidivism}~\cite{agarwal2021towards}: In the Recidivism dataset, nodes are defendants released on bail during 1990-2009~\cite{jordan2015effect}. Two nodes are connected if two defendants' past criminal records and demographics are similar. The task is to predict a defendant as bail (i.e., unlikely to commit a violent crime if released) or no bail (i.e., likely to commit a violent crime) with race being the sensitive attribute.
    \item \textbf{Credit Defaulter}~\cite{agarwal2021towards}: In this dataset, nodes represent credit card users and they are connected based on the similarity of their purchase and payment records. The sensitive attribute of this dataset is age and the task is to classify whether a user will default on credit card payment.
    \item \textbf{MovieLens}~\cite{bose2019compositional}: The MovieLens dataset is a recommender system  benchmark~\citep{harper2015movielens}, whose target task is to predict the rating that users assign movies. Sensitive attributes about the user features, such as age, gender, and occupation, are covered in the dataset.
    \item \textbf{Reddit}~\cite{bose2019compositional}: The Reddit dataset is based on the social media website Reddit where users can comment on content in different topical communities, called “subreddits”. 
    % The edge prediction task is to predict interactions between users and subreddit communities. 
    For sensitive attributes, this dataset treats certain subreddit nodes as sensitive nodes, and the sensitive attributes for users are whether they have an edge connecting to these sensitive nodes.
    \item \textbf{Polblog}~\cite{palowitch2019monet}: Polblog is a blog website network ~\cite{adamic2005political}. Nodes represent blogs and links denote hyperlink between blogs. The sensitive attributes for this dataset are blog affiliation communities.
    \item \textbf{Twitter} ~\cite{khajehnejad2021crosswalk}: This is a subgraph of the Twitter dataset~\cite{babaei2016efficiency, cha2010measuring}. The sensitive attribute is the political leaning of each user, including neutrals, liberals and conservatives. 
     \item \textbf{Facebook} \cite{kang2020inform}: The dataset is collected from the Facebook website . Nodes are users and edges represent friendship between users ~\cite{mcauley2012learning}. The sensitive attribute for this dataset is gender.
    \item \textbf{Google+} \cite{masrour2020bursting}: It is collected from Google+ ~\cite{mcauley2012learning}. Nodes in the dataset are users and they are connected concerning their social relationships. The sensitive attribute for this dataset is gender.
    \item \textbf{Dutch} \cite{masrour2020bursting}: This is from the school network ~\cite{snijders2010introduction} with gender as the sensitive attribute. It corresponds to friendship relations among 26 freshmen at a secondary school in the Netherlands.
\end{itemize}

\begin{table}[t]
    \scriptsize
    \centering
    \caption{Datasets for fair graph neural networks.}
    \vskip -1.5em
    % \resizebox{0.98\linewidth}{!}{
    \begin{tabular}{llllllll}
    \toprule
    Task & Dataset & Labels & Sens. & \#Nodes & \#Edges & \#Features & References\\
    \midrule
    &Pokec-n & Job & Region & 66,569 & 729,129 & 59 & \cite{dai2021say} \cite{wang2021unbiased} \cite{kose2021fairness}\\ 
    &Pokec-z &  Job & Region & 67,797 &  882,765 & 59 & \cite{dai2021say} \cite{wang2021unbiased} \cite{kose2021fairness}\\
    Node & NBA & Salary & Nationality & 403 & 16,570 & 39 & \cite{dai2021say}\\
    Classification & German Credit& Credit Risk & Gender & 1,000 &  22,242 & 27 & \cite{agarwal2021towards} \cite{dong2021edits} \\
    &Recidivism& Bail & Race &  18,876 &  321,308 & 18 & \cite{agarwal2021towards} \cite{dong2021edits} \\
    &Credit Def.& Default & Age & 30,000 &  1,436,858 & 13 & \cite{agarwal2021towards} \cite{dong2021edits} \\
    \midrule
     & MovieLens & - & Multi-attribute & 9,940 &  1,000,209 & - & \cite{bose2019compositional} \cite{buyl2020debayes} \cite{palowitch2019monet}\\
    &Reddit & - & Multi-attribute & 385,735 &  7,255,096 & - & \cite{bose2019compositional} \\
    &Polblog &- & Community & 1,107 & 19,034 & - & \cite{palowitch2019monet}\\
    Link &Twitter & - & Politics & 3,560 & 6,677 & - & \cite{khajehnejad2021crosswalk}\\
    Prediction &Facebook & - & gender & 22,470 & 171,002 & - & \cite{masrour2020bursting} \cite{spinelli2021biased} \cite{kang2020inform} \cite{dong2021individual} \cite{li2020dyadic}\\
    &Google+ & - & gender & 4,938 & 547,923 & - & \cite{masrour2020bursting}\\
    &Dutch  & - & gender & 26 & 221 & - & \cite{masrour2020bursting}\\
    \bottomrule
    \end{tabular}
    % }
    \label{tab:fair_datasets}
    \vskip -1em
\end{table}

\subsection{Applications of Fair GNNs}
% In this subsection, we discuss representative applications of fair graph neural networks.

\vspace{0.2em} \noindent \textbf{Social Network Analysis.} With the emerging of social media platforms such as Facebook, Twitter, and Instagram, the social network analysis is widely conducted to provided better service to users. For example, the platforms may use GNNs to recommend new friends to a user~\cite{fan2019graph}. Node classification are also widely conducted on social networks to further complete user profile for better service~\cite{dai2021say}. However, recent works~\cite{dai2021say,rahman2019fairwalk,stoica2018algorithmic} indicate that GNNs can be biased to the minority in friends recommendation and node classification on social networks. For instance, the algorithm have been found to prevent minorities from becoming influencers. 
The message-passing on graphs can magnify the bias~\cite{dai2021say}. Therefore, several fair GNNs~\cite{dai2021say,wang2021unbiased,kose2021fairness,masrour2020bursting,khajehnejad2021crosswalk} for social network analysis have been proposed.

\vspace{0.2em} \noindent \textbf{Recommender System.} The user interactions on products such as books can link the users and products to compose bipartite graph. In addition, the social context of users may also be utilized for recommendation. Because the great power of GNN in process graphs, many platforms have applied GNNs for the recommendation system~\cite{hamilton2017inductive,ying2018graph}. But fairness issue is also reported in recommendation system. For instance, it is found that a GNN-based algorithm on book recommendation may be biased towards suggesting books with male authors~\cite{buyl2020debayes}. Hence, it is necessary to develop fair graph neural networks for recommendation system.

\vspace{0.2em} \noindent \textbf{Financial Analysis.} Recently, there is a growing interest in applying GNNs to financial applications such as loan default risk prediction~\cite{liang2021credit,cheng2019risk} and fraud detection~\cite{rao2020xfraud}. In loan default risk, the guarantee network~\cite{cheng2019risk} or user relational graph~\cite{liang2021credit} can be applied to learn more powerful representations for predictions. In fraud detection, GNNs on transaction~\cite{rao2020xfraud} are also investigated.  Similar to the applications in other domains, GNNs also exhibit bias towards protected attributes such as genders and ages in financial analysis~\cite{agarwal2021towards}. Using fair GNNs~\cite{agarwal2021towards,dong2021edits} in finance can ensure the fairness to users and avoid the social and legal issues caused by the bias in the GNN model. 

\subsection{Future Research Directions of Fair GNNs}
Though many fair models on graph-structured data have been investigated, there are still many important and challenging directions to be explored. Next, we list some promising research directions.

\vspace{0.2em} \noindent \textbf{Attack and Defense in Fairness.} 
Recent works have shown that an poisoning attacker can fool the fair machine learning model to exacerbate the algorithmic bias~\cite{solans2020poisoning,mehrabi2020exacerbating,van2021poisoning}. For instance, one can generate poisoned data samples by maximizing the covariance between the sensitive attributes and the decision outcome and affect the fairness of the model. Thus, a seriously biased model caused by the attacker might be treated as a fair model by the end-user due to the deployment of fair algorithms, which can result in social, ethical and legal issues. 
Since GNNs are an extension of deep learning on graphs, fair GNNs are also at a risk of being attacked. Without understanding the vulnerability and robustness of fair GNNs, we cannot fully trust a fair GNN.  Despite the initial efforts on attacking fair models~\cite{solans2020poisoning,mehrabi2020exacerbating,van2021poisoning}, all of them focus on i.i.d data; while the studies on vulnerability of fair GNNs are rather limited. Note that to achieve a trustworthy GNN, the robustness and fairness should be simultaneously meet. However, as it is discussed in Section~\ref{sec:robust}, current robust GNNs generally focus on the robustness in terms of performance and rarely investigate the robust models against attacks in both accuracy and fairness. Therefore, it is crucial to investigate the vulnerability of fair GNNs and develop robust fair GNNs.  
% \suhang{connect to robust section. Note that we want to make fair, robust, privacy, interpretability connected, instead of separated. For example, explainability will be useful to exam if a model is fair or if some data sample are poisoned}

\vspace{0.2em} \noindent \textbf{Fairness on Heterogeneous Graphs.} Many real-world graphs such as social networks, knowledge graph, and biological networks are heterogeneous graphs, i.e., networks containing diverse types of nodes and/or relationships. Various GNNs  have been proposed to address the challenge of representation learning on heterogeneous graphs, such as learning with meta-path~\cite{wang2019heterogeneous,hu2020heterogeneous} and designing new message-passing mechanisms for heterogeneous graphs~\cite{schlichtkrull2018modeling}. Recently, it is reported that the representations learned by heterogeneous GNNs can contain discrimination~\cite{zeng2021fair}, which could result in societal prejudice in the applications. For example, the social biases have been identified in knowledge graphs~\cite{shrestha2019fairness}. And the representations of knowledge graph learned by heterogeneous GNNs are widely adopted to facilitate the searching and recommender system. Hence, the encoded biases in the representations could lead to detrimental societal consequences.  
However, existing fair algorithms are generally designed for GNNs on homogeneous graphs, which are not able to mitigate the bias brought by the meth-path neighbors or the message-passing mechanism specifically designed for heterogeneous information networks.
Therefore, it is necessary to develop fair GNns to address the unique challenges brought by the heterogeneity graphs.

\vspace{0.2em} \noindent \textbf{Fairness without Sensitive Attributes.} Despite the ability of the aforementioned methods in alleviating the bias issues, they generally require abundant sensitive attributes to achieve fairness; while for many real-world applications, it is difficult to collect sensitive attributes of subjects due to various reasons such as privacy issues, and legal and regulatory restrictions. As a result of, most of existing fair GNNs are challenged due to the lacking of sensitive attributes in training data. Though investigating fair models without sensitive attributes is important and challenging, only some initial efforts on i.i.d data have been conducted~\cite{hashimoto2018fairness,zhao2021you,lahoti2020fairness}. How to learn fair GNN without sensitive attributes is a promising research direction.

% \noindent \textbf{Fair Pretraining GNNs.} To address the problem of lacking labels, various pretraining methods~\cite{qiu2020gcc,hu2019strategies,hu2020gpt} have been proposed to benefit downstream tasks. Supervised tasks~\cite{hu2019strategies} and many self-supervised tasks~\cite{qiu2020gcc,hu2020gpt} such as link prediction and contrastive learning have been explored to learn better node representations.  Despite the achievements of pretraining GNNs, the bias in the pretraining dataset may lead to discrimination, affecting the fairness of downstream models. In contrast, literature that focus on fairness in pretraining GNNs are rather limited. Therefore, developing a fair algorithm for pretraining GNNs that eliminate the discrimination from the pretrained dataset and even reduce the bias in downstream task predictions is a promising direction. \suhang{maybe remove this part for now}
\section{Explainability of Graph Neural Networks} \label{sec:explain}

Parallel to the effectiveness and prevalence of deep graph learning systems, the property of lacking interpretability is shared by most deep neural networks (DNNs). DNNs typically stack multiple complex nonlinear layers~\cite{zhang2018visual}, resulting in predictions difficult to understand. To expose the black box of these highly complex deep models in a systematic and interpretable manner,  explainable DNNs~\cite{nauta2022anecdotal}, 
have been explored recently. However, most of these works focus on images or texts, which cannot be directly applied to GNNs due to the discreteness of graph topology and the message-passing of GNNs. And it is very important to understand GNNs' predictions for two reasons. \textit{First}, it enhances practitioners' trust in GNN models by enriching their understanding of the network characteristics. \textit{Second}, it increases the models' transparency to enable trusted applications in decision-critical fields sensitive to fairness, privacy and safety challenges. High-quality explanations can expose the knowledge captured, helping users to evaluate the existence of possible biases, and make the model more trustworthy. For example, counterfactual explanations are utilized in ~\cite{sharma2020certifai} to analyze the fairness and robustness of black-box models, in order to build a responsible artificial intelligence system. A model-agnostic explanation interface is also designed in~\cite{baniecki2020dalex} to continuously monitor model performance and validate its fairness. Therefore, explainable GNNs are attracting increased attention recently and many efforts have been taken.

In this section, we will provide a comprehensive survey on the current progress in the explainability of GNNs. First, we introduce the background of explaining GNN models and provide a motivating example. Following that, a comprehensive review of existing explanation methods would be presented. Popular datasets and evaluation metrics in this domain are also introduced. Finally, we go through some future research directions. Compared to the existing review in ~\cite{yuan2020explainability}, the main improvement of this survey is that we covered more recent progress such as self-explainable GNNs~\cite{zhang2021protgnn,dai2021towards} and discuss more reliable evaluation settings~\cite{faber2021comparing}.

%\suhang{Tianxiang, can you discuss some motivations or applications of explainability, such as detecting unfairness or adversarial examples, so we can connect with other parts on fairness and robustness. There are several papers utilize explainability for fairness and robustness. Please search and read.}

\subsection{Backgrounds}

\subsubsection{Aspects of Explanation}
As discussed in ~\cite{mittelstadt2019explaining,danilevsky2020survey}, the term explainability itself needs to be explained. Generally, explainability in GNNs should (i) guide end-users or model designers to understand how the model arrives at its results; (ii) enable users to have an expectation on the decisions; (iii) and provide information on how and when the trained model might break. To cope with the needs of obtained explanations, explainability must be considered within the context of particular disciplines. As a result, developed explanation methods often show a high level of variety and provide explainability at different levels. Generally, explanations can be categorized from the following perspectives: 
\begin{itemize}[leftmargin=*]
    \item \textbf{Global} or \textbf{Local} explanations. A local explanation (or instance-level explanation) provides justification for prediction on each specific instance. Currently, most GNN explanation works~\cite{ying2019gnnexplainer,luo2020parameterized} fall within this group. On the other hand, global explanations (or model-level explanations) ~\cite{yuan2020xgnn} reveal how the model's inference process works, independently of any particular input.
    \item \textbf{Self-explainable} or \textbf{Post-hoc} explanations. Self-explainable GNNs design specific GNN models that are interpretable intrinsically, which can simultaneously give the prediction and corresponding explanation. The explainability arises as part of the prediction process for self-explaining methods.
    On the contrary,  post-hoc explanations focus on providing explanations on a trained model. An additional explainer model is generally adopted for the post-hoc explanations~\cite{ying2019gnnexplainer}. However, due to the adoption of the explainer model, the post-hoc explanations may misinterpret the actual inner working of the target model. %are model-agnostic and require to conduct additional operations after predictions are made.
    \item \textbf{Explanation forms} to be presented to the end users. Explanations should help users to understand GNNs' behaviors. Various explanation forms have been investigated such as \textit{bag-of-edges}~\cite{baldassarre2019explainability}, \textit{attributes importance}~\cite{ying2019gnnexplainer}, \textit{sub-graphs}~\cite{yuan2021explainability}, etc. Different explanation forms can give different visualizations and offer different information to end users.
    \item \textbf{Techniques} for deriving the explanations. To enable the explainability, various explanation techniques have been developed including perturbing the input~\cite{ying2019gnnexplainer}, analyzing internal inference process~\cite{baldassarre2019explainability}, designing intrinsically interpretable GNN models~\cite{dai2021towards}, etc. These methods make different assumptions about the model and their advantages may vary across datasets.
\end{itemize}
A comprehensive taxonomy and detailed introduction of methods are presented in Section ~\ref{sec:taxonomy_expl}.

\subsection{Desired Qualities of Explanations}
%\suhang{please include the criteria of explanations, such as fidelity, faithful, user-friendly and so on}
With the explanation model and its produced explanations obtained, different aspects regarding explanation quality can be evaluated. Whereas a gold standard exists for comparing predictive models, there is no agreed-upon evaluation strategy for explainable AI methods. As argued in ~\cite{nauta2022anecdotal}, evaluating the plausibility and convincingness of an explanation to humans is different from evaluating its correctness, and those criteria should not be conflated. In this part, we systematically analyze certain properties that good explanations should satisfy. Considerations of these qualities motivate the design of different explanation methods and evaluation metrics, which will be discussed in Section~\ref{sec:taxonomy_expl} and Section~\ref{sec:expl_evaluation} respectively. %which are closely relate to the design and evaluation of explanations:
\begin{itemize}[leftmargin=*]
    \item \textbf{Correctness}: The obtained explanations should be correct, and truthfully reflect the reasoning of the target predictive model (either locally or globally). This quality addresses the faithfulness of explanations and requires that \textit{descriptive accuracy} of the explanation is high~\cite{murdoch2019definitions}.
    \item \textbf{Completeness}: Completeness addresses the extent to which identified explanations explain the target model. Ideal explanations should contain ``the whole truth''~\cite{kulesza2013too}. High completeness is desired to provide enough details, and it should be balanced with correctness~\cite{nauta2022anecdotal}.
    \item \textbf{Consistency}: The obtained explanations should be consistent concerning to inputs. In other words, explaining should be deterministic and the same explanation should be provided for identical inputs~\cite{robnik2018perturbation}. It has also been argued that explanations should be invariant toward small perturbations~\cite{alvarez2018robustness}.
    \item \textbf{Contrastivity}: Contrastivity facilitates distinctions of obtained explanations of different predictions. Explanation of a certain event should be discriminative in comparison to those of other events~\cite{miller2019explanation}. For models taking different decision strategies, explanations of their behaviors should also be distinct~\cite{adebayo2018sanity}.
    \item \textbf{User-friendly}: An ideal explanation is expected to be user-friendly. Explanations should be presented in a form that is clear, easy to interpret, and ``agree with human rationales''~\cite{atanasova2020diagnostic}. For example, it is argued that explanations should be compact, sparse and avoid redundancy~\cite{nauta2022anecdotal}. Some formats are also found to be more easily interpretable than others~\cite{huysmans2011empirical}.
    \item \textbf{Causality}: A causal explanation provides insights into the cause-and-effect relationships that determines model decisions~\cite{bajaj2021robust}. Such explanations not only describe the correlations or patterns identified by the model but also delve into the reasons or mechanisms driving those patterns~\cite{verma2020counterfactual}. Ideally, a causal explanation should help users differentiate between spurious correlations and actual causal influences, thereby enabling more reasonable explanations~\cite{lin2021generative}.
\end{itemize}

\subsubsection{Explanation Example}

\begin{wrapfigure}{r}{0.38\linewidth}
    \vskip -1em
    \includegraphics[width=0.38\textwidth]{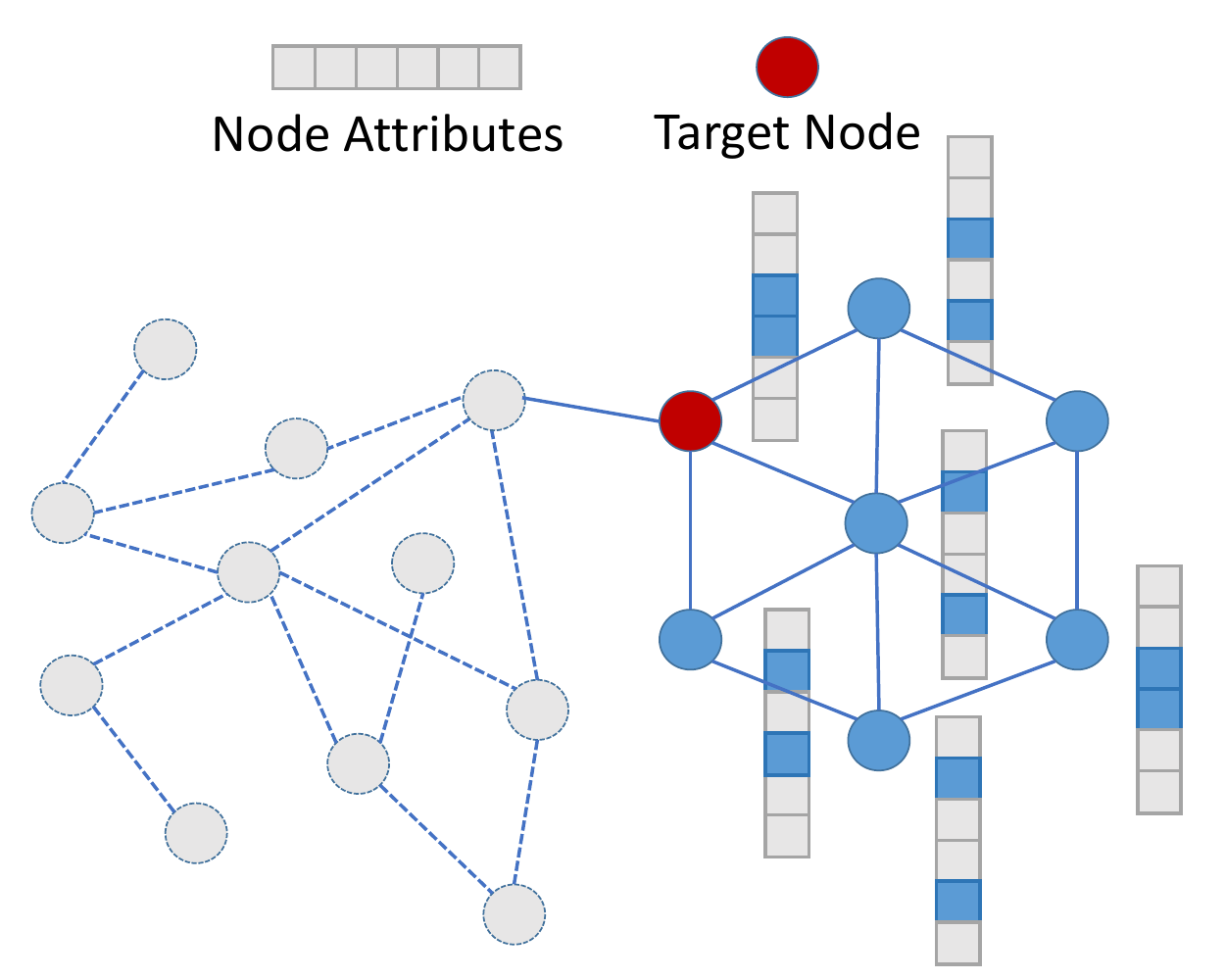}
    \vskip -1em
    \caption{An example of instance-level explanation, where important nodes, attributes and edges for predictions are highlighted} \label{fig:expl_target}
    \vskip -1em
\end{wrapfigure}

Due to diverse settings and the complexity of existing algorithms, discussing and comparing GNN explanation methods can often become abstract. To make this explanation task more concrete, we give an example of instance-level explanation, which has been widely taken in various works~\cite{ying2019gnnexplainer,luo2020parameterized,yuan2021explainability}.
As shown in Figure~\ref{fig:expl_target}, the explanation objective is to find discriminative substructures, including edges and node attributes, that are important for the prediction of to-be-explained GNN. Note that there are works investigating other explanation forms, like class-wise prototypical structures~\cite{zhang2021protgnn} and interpretable surrogate models~\cite{huang2020graphlime}.

\subsubsection{Challenges of Explainability of GNNs}
Besides common difficulties in explaining deep models, there are certain properties making the explanation in graph domain particularly challenging than in other domains like images or texts. \textit{First}, GNN captures both node attributes and structural topology information, relying upon the message-passing~\cite{gilmer2017neural} along edges. It is difficult to estimate contribution of edges as they are discrete, and could appear in the computational graphs multiple times at different layers. In turn, identifying important structures is even more complicated as interactions among nodes and edges are involved. \textit{Second}, graph is less intuitive than images or texts. It is difficult for users to analogize the commonalities and dissimilarities among graphs, making the evaluation and interpretation of obtained explanations (still in the form of graphs) challenging. Domain knowledge often is required to understand the obtained graph ``explanations''.

\subsection{Taxonomy of Explainability of Graph Neural Networks}\label{sec:taxonomy_expl}

% The explanation methods for GNN can be categorised into 2 level, the instance-level and model-level. For instance-level, we will introduce three kinds of methods, the attribution methods, the decomposition methods and the surrogate methods.

To make deep models applicable in real scenarios, researchers have made extensive attempts to get an explanation from deep models, especially in the image and text domains. However, due to the complexity of graph data and the less human-understandable message-passing mechanism in GNNs, it is difficult to directly extend explanation methods for image or text data to graph data. Recently, to address these challenges, researchers begin to focus on the explainability of GNN models and propose many specific models. %In this section, we introduce the interprete and explaina graph neural network models. Following the categorization of \cite{yuan2020explainability}, the explanation methods can be split into \textbf{instance-level methods} and \textbf{model-level methods}. However, besides these post-hoc techniques which explain the predictions, there are some works about self-explainable graph neural networks like \cite{dai2021towards,zhang2021protgnn} so we have the third categorization of \textbf{self-explanation approaches}. 
In this section, we provide a high-level summary of existing GNN explanation methods and categorize them into three groups: (1) instance-level post-hoc  explanations, (2) model-level post-hoc  explanations, and (3) self-explainable methods that are intrinsically interpretable. Most existing GNN explanation methods are designed for the instance-level post-hoc setting, and we further arrange them based on their adopted techniques for achieving explainability. A summary of method taxonomy is shown in Table~\ref{tab:explain_category}.

\begin{table}[t]
    %\vskip -1em 
     \scriptsize
     \centering
     \caption{Categorization of explanation models on graphs}
     \vskip -1.5em
     \begin{tabular}{lccccl} 
     \toprule
     \multicolumn{5}{c}{Category} & References \\
     \midrule
     \multirow{5}{*}{Instance-level Post-hoc}& Gradient & Perturbation & Decomposition & Surrogate & \\ 
     & $\surd$ &  &  &  &~\cite{baldassarre2019explainability,pope2019explainability} \\ 
      & & $\surd$  &  &  & ~\cite{yuan2021explainability,ying2019gnnexplainer,luo2020parameterized} \\
      & &  &  $\surd$ &   & ~\cite{baldassarre2019explainability, schwarzenberg2019layerwise,pope2019explainability}\\
     & &  &  & $\surd$  &  ~\cite{huang2020graphlime,zhang2021relex,vu2020pgm}\\ 
     \midrule
     \multicolumn{5}{l}{Model-level Post-hoc} & ~\cite{yuan2020xgnn,zhang2021protgnn}\\ 
     \midrule
     \multicolumn{5}{l}{Self-explainable} & ~\cite{dai2021towards,zhang2021protgnn}\\ 
     \bottomrule
     \end{tabular}
     \label{tab:explain_category}
     \vskip -1em
 \end{table}

\subsubsection{Instance-level Post-hoc Explanation}

%\suhang{the descriptions are too general. For representative methods, please provide equations to help understand the details. For example, for post-hoc instance level explanations, you can write down the general loss function. Then discuss how existing work optimize the equation.}

Instance-level post-hoc explanation identifies elements (like node attributes and edges) that are crucial for model's prediction for each specific input instance. Typically, given an input graph $\mathcal{G}$, which could be a graph sample for graph-level tasks or the local graph of a node for node-level tasks, it aims to find a sub-graph $\mathcal{G}_s \subset \mathcal{G}$ that accounts for prediction output of the target GNN model. Based on different strategies in identifying input substructures as explanations, we can further summarize existing methods into three groups: (1) \textit{Attribution Methods}, which directly analyze the influence of input elements on the prediction using gradients or through perturbations; (2) \textit{Decomposition Methods}, which examine the inference of deep models by decomposing prediction result into importance mass on the input; (3) \textit{Surrogate Methods}, which train an interpretable model that mimics the behavior of to-be-explained deep model within the neighborhood of current input. Next, we introduce the details of these three categories.

\vspace{0.2em} \noindent \textbf{Attribution Methods.}
Attribution, also referred to as relevance~\cite{xiong2018looking}, aims to reveal components of high importance in the input. These methods provide explanations through measuring the contribution of input elements in the target decision, and finding the subgraph with top contribution weights as explanation $\mathcal{G}_s$. Based on their strategies in estimating each element's contribution, we can further categorize them into two types: \textit{Gradient-based} and \textit{Perturbation-based}. %\enyan{@Tianxiang,  two-level bullets with long content seems a little bit difficult to read. Should we consider just split these bullets as paragraphs with some connection sentences? And a table would be also make the categorization more clear.}

\textit{Gradient-based Attribution}.  Gradient-based attribution methods estimate importance weights by back-propagated gradients. Based on Taylor expansion, gradients from model output w.r.t input elements reflect its sensitivity towards them, which can be utilized as an importance estimator~\cite{baehrens2010explain}. For example, importance of node $v$ with attribute $\mathbf{x}_v$ for prediction $y^c$ can be computed as $\|Relu(\frac{\partial y^c}{\mathbf{x}_v}) \|_1$~\cite{pope2019explainability}. One weakness of these methods is that local gradients could be unreliable because of gradient saturation problem~\cite{sundararajan2017axiomatic}. Therefore, integrated gradients (IG)~\cite{sundararajan2017axiomatic} that consider the gradients along a path were proposed to address this problem. On graphs, IG score of node attributes can be computed in the form of: %\enyan{ is $x_i$  one dim of attribute vector? or attribute vector of node $v_i$?.}:
    \begin{equation}\label{eq:integrated_gradient}
        IG(\mathbf{x}_v) = (\mathbf{x}_v -\mathbf{x}_v')\times \int_{\alpha=0}^1 \frac{\partial f(\mathbf{x}_v'+ \alpha \cdot (\mathbf{x}_v-\mathbf{x}_v'))}{\partial \mathbf{x}_v} d \alpha,
    \end{equation} 
where $\mathbf{x}_v'$ represents the baseline attributes which can be set to the global average. Essentially, Eq.(\ref{eq:integrated_gradient}) integrates the gradients at all points along the path from $\mathbf{x}_v'$ to $\mathbf{x}_v$, instead of relying on gradient at $\mathbf{x}_v$ which may suffer from saturated gradients. Existing works have investigated various gradient-based methods to explain GNNs, such as SA~\citep{baldassarre2019explainability}, Guided BP~\citep{baldassarre2019explainability}, CAM~\citep{pope2019explainability} and Grad-CAM~\citep{pope2019explainability}. These methods share similar ideas to identify important input elements. The main difference lies in the procedure of gradient back-propagation and how different hidden feature maps are combined~\citep{yuan2020explainability}. For example, Guided BP~\cite{baldassarre2019explainability} clips negative gradients during conducting back-propagation to estimate contribution weights. CAM~\cite{pope2019explainability} requires that a linear layer is used for classification, and calculates heat maps over nodes using node embeddings from the last GNN layer along with weights of that linear classification layer. Grad-CAM~\cite{pope2019explainability} generalizes it to the model-agnostic setting by using average class-wise gradients in place of linear classification parameters. It is straightforward to estimate importance weights of node attributes using these methods, and edges connecting important nodes would also be taken as important~\cite{9679041}.
    
\textit{Perturbation-based Attribution}. Perturbation-based attribution methods try to learn a perturbation mask and examine prediction variations w.r.t perturbations. Specifically, the mask is optimized to maximize the perturbation (mask out as many edges and nodes as possible) while preserving original predictions. Those left-out unperturbed nodes and edges are taken as the most important elements contributing to the prediction, which corresponds to the explanations. Let $\mathbf{M}_{A} \in \{0,1\}^{n \times n}$ and $\mathbf{M}_{X} \in \{0,1\}^{n \times d}$ denote binary masks on edges and node attributes, respectively. It can be formulated as the following optimization problem:
\begin{equation}\label{eq:perturb_framework}
    \begin{aligned}
    \min_{\mathbf{M}_{A}, \mathbf{M}_{X}} &\mathcal{L}_{dif}\big(f(\mathbf{A}, \mathbf{X}), f(\hat{\mathbf{A}}, \hat{\mathbf{X}} ) \big) + \beta \cdot \mathcal{R}(\mathbf{M}_{X},\mathbf{M}_{A}), 
     \quad \hat{\mathbf{A}} \sim \mathcal{P}(\mathbf{A}, \mathbf{M}_{A}), \quad \hat{\mathbf{X}} \sim \mathcal{P}(\mathbf{X}, \mathbf{M}_{X}).
    \end{aligned}
\end{equation}
where $\mathcal{P}$ denotes the perturbations on original input with provided importance masks, and we use $\hat{\mathbf{A}}$ to represent perturbed $\mathbf{A}$. So is the case of $\hat{\mathbf{X}}$. With this optimization objective, explanations are found by finding input elements that preserve model predictions. For example, in GNNExplainer~\cite{ying2019gnnexplainer}, $\mathcal{P}(\mathbf{A}, \mathbf{M}_{A}) = \mathbf{A} \odot \mathbf{M}_{A}$ and $\mathcal{P}(\mathbf{X}, \mathbf{M}_{X}) = \mathbf{Z} + (\mathbf{X}-\mathbf{Z}) \odot \mathbf{M}_{X}$, where $\mathbf{Z}$ is sampled from marginal distribution of node attributes $\mathbf{F}$ and $\odot$ denotes elementwise multiplication. $\mathcal{L}_{dif}$ is usually implemented as cross entropy loss~\cite{ying2019gnnexplainer}, which encourages consistency on prediction outputs. $\mathcal{R}$ regularizes identified explanations and is usually implemented as sparsity constraint. This objective promotes the correctness and non-redundancy of found explanations. 
    
%These algorithms are developed under the assumption that smaller variations indicate less importance of corresponding attributes or edges, and the desired explanation is the minimal input that can keep GNN's prediction unchanged. 
Eq.(\ref{eq:perturb_framework}) is a discrete optimization problem, which is difficult to solve directly. %Due to the combinatory nature of graphs, it is infeasible to test all possible substructures as inputs. 
Various learning paradigms have been proposed to efficiently find the masks~\cite{yuan2021explainability,ying2019gnnexplainer,luo2020parameterized}. Based on the adopted strategy in learning perturbation masks, these approaches can be further divided into three groups.

The first group identifies effective perturbation masks by conducting \textit{Searching}~\cite{yuan2021explainability,funke2021hard}. These methods optimize Eq.(\ref{eq:perturb_framework}) and learn perturbation masks with the explicit \textit{test-and-run} paradigm. And the optimization directions are found by search algorithms. For example, SubgraphX~\citep{yuan2021explainability} employs Monte Carlo Tree Search (MCTS) algorithm to search for the most important subgraph as the explanation of predictions. The Shapley value is used as the measurement of component's importance during the search phase. ZORRO~\citep{funke2021hard} revises the search process by explicitly encoding fidelity into the objective. Causal Screening~\citep{wang2020causal} also falls into this group, which incrementally selects input elements by maximizing individual causal effects at each search step. CF-GNNExplainer~\cite{lucic2022cf} focuses on counterfactual explanations, which is derived from identifying minimal perturbations that can change prediction results.

The second group uses \textit{Attention} mechanism to learn perturbation masks~\cite{ying2019gnnexplainer,velivckovic2017graph}. By relaxing binary masks  $\mathbf{M}_{A} \in \{0,1\}^{n \times n}$ and $\mathbf{M}_{X} \in \{0,1\}^{n \times d}$ into soft ones, i.e., $\mathbf{M}_{A} \in [0,1]^{n \times n}$ and $\mathbf{M}_{X} \in [0,1]^{n \times d}$, the soft perturbation masks can be directly optimized in an end-to-end manner. For example, GNNExplainer~\cite{ying2019gnnexplainer} employs a soft mask on attributes by element-wise multiplication and a mask on edges with Gumbel softmax. Then, these two masks are directly optimized with the objective of minimizing size of unperturbed parts while preserving prediction results. 

The last group~\cite{luo2020parameterized,wang2021towards} use an \textit{Auxiliary Model} to predict effective perturbation masks with the information from graph and target model. The auxiliary model is optimized with Eq.(\ref{eq:perturb_framework}) on training samples. And it is assumed to be general and can safely explain new-coming graphs after training. For example, PGExplainer~\cite{luo2020parameterized} adopts an explanation network $g_\phi$ to predict preserving probability of each given edge based on node embeddings derived from the target model $f_t(\mathcal{G})$. It can be formally written as $\mathbf{M}_{A} \sim g_{\phi}(\mathcal{G}, f_t(\mathcal{G}))$. Unlike the previous two strategies, the perturbation mask does not need to be re-learned from scratch for each to-be-explained graph, and this group of methods is much faster to use in the test time. Another representative method is GraphMask~\citep{schlichtkrull2020interpreting} which trains a model to produce layer-wise edge perturbation masks. It takes node embeddings from the corresponding layer as input, and provides different edge importance masks at different propagation steps. Gem~\cite{lin2021generative} introduces the notion of causality into the explanation generation process, which trains a causal explanation model equipped with a loss function based on Granger causality. It has better generalization ability as it has no requirement w.r.t internal GNN structures or knowledge of learning tasks. RCExplainer~\cite{bajaj2021robust} aims to modeling the common decision logic of GNNs across similar input graphs. This approach ensures noise resistance by leveraging shared decision boundaries, and guarantees counterfactual integrity by ensuring prediction changes upon the removal of identified edges.

\vspace{0.2em} \noindent \textbf{Decomposition Methods.}
These methods seek to decompose the prediction of target GNN model into contribution of input features. A contribution score would be assigned to each input element, and an explanation is obtained via identifying inputs with the highest scores. Concretely, the influence mass is back-propagated layer-by-layer onto each input element. And the influence mass from input to output will be decomposed based on neural excitation at each layer. During this process, nonlinear components of the GNN model are generally neglected to ease the problem. Popular decomposition strategies on image data include Layer-wise Relevance Propagation (LRP)~\cite{bach2015pixel} and Excitation BP~\cite{springenberg2014striving}. Several efforts have been made to extend them to graphs data~\citep{baldassarre2019explainability, schwarzenberg2019layerwise,pope2019explainability}. For example, $\alpha\beta$-rule and $\epsilon$-stabilized decomposition rule in original LRP algorithm are extended to work on message passing mechanism of GNN~\cite{baldassarre2019explainability}. GNN-LRP~\cite{schnake2020higher} evaluates contributions of bag-of-edges and deduces back-propagation rules on graph walks with high-order Taylor decomposition. For example in GCN, as shown in ~\cite{schnake2020higher}, the back-propagation rule decomposing contribution mass of node $K$ at layer $l+1$ to node $J$ at layer $l$ can be written as Eq.(\ref{eq:LRP}). This algorithm starts by assigning full contribution mass to the target output, and redistributes it with a backward pass on GNN layer by layer. For simplicity, we use $R_{K}^{l+1,k}$ to represent contribution mass of $k$-th dimension of node K's embedding at layer $l+1$, and use $R_{JK}^{l,j}$ to denote its decomposed contribution to $j$-th dimension of node J's embedding at layer $l$. Assuming node embedding dimension to be $d$, we have:
\begin{equation}
    \begin{aligned}
    R_{JK}^{l,j} &= \sum_{k=1}^{d}\frac{\lambda_{JK}h_{J}^{l,j}w_{jk}}{\sum_{J\in \mathcal{V}}\sum_{j=1}^{d}\lambda_{JK}h_{J}^{l,j}w_{jk}} R_{K}^{l+1,k},
    \end{aligned}
\label{eq:LRP}
\end{equation}
where with $h_J^{l,j}$ is $j$-th dimension of node $J$ embedding at layer $l$, $w_{jk}$ represents the weight linking neuron $j$ in layer $l$ to neuron $k$ in layer $l+1$ which is a scalar, and $\lambda_{JK}$ is the edge weight connecting node $J$ to node $K$. Following this rule, contributions can be back-propagated to the inputs, and those nodes with highest $R$ value are preserved as the explanation $\mathcal{G}_s$.

%the $\epsilon$-stabilized back-propagation rule at layer $l$ with $h_i^{(l)}$ indicating $i$-th node and $w_{ij}$ as weights from $h_i^{(l)}$ to $h_j^{(l+1)}$, provided by ~\cite{baldassarre2019explainability} is as follows:
%\begin{equation}
%    \begin{aligned}
%    z_{ij} &= h_i^{(l)}w_{ij}, \quad
%    R_i^{(l)} &= \sum_j\frac{z_{ij}^+}{\sum_{i'}z_{i'j}^+ + b_{j}^{+} + \epsilon} R_{j}^{(l+1)},
%    \end{aligned}
%\end{equation}
%where $R_{i}^{(l)}$ represents contribution weight of node $i$ at layer $l$, $b$ is the bias term, $\epsilon$ is a small value to avoid division by zero, and $z_{ij}$ is the computed contribution mass from $h_i^{(l)}$ to $h_j^{(l+1)}$. Contributions are back-propagated to the inputs, and those nodes with highest $R$ value are preserved as the explanation $\mathcal{G}_s$.

\vspace{0.2em} \noindent \textbf{Surrogate Methods.}
Neural networks are treated as black-box models due to their deep architectures and nonlinear operations. It is observed that they hold a highly-complex loss landscape~\cite{li2018visualizing} and are challenging to be explained directly. To circumvent nonlinear classification boundary of the trained DNN model, many attempts are made~\cite{ribeiro2016should} to approximate DNN's local prediction around each instance $\mathbf{x}$ with simple interpretable models such as logistic regression. Specifically, for the target instance $\mathbf{x}$, a group of prediction records can be obtained by applying small random noises to it and collecting the target model's prediction on these perturbed inputs. Then, the interpretable surrogate model can be trained on these records to mimic target model's behavior locally, which serves as explanations. The searching process of interpretable local model $\xi(x)$ can be written as:
\begin{equation}
    \xi(x) = \arg\min_{f' \in F'} \mathcal{L}(f, f', \pi_x) + \Omega(f').
\end{equation}
where $F'$ represents candidate interpretable model families and $\pi_x$ denotes local neighborhood around instance $x$. $\mathcal{L}(f, f', \pi_x)$ measures faithfulness of the surrogate model $f'$ in approximating the target model $f$ in the locality $\pi_x$. $\Omega()$ measures model complexity to encourage simple surrogate models~\cite{ribeiro2016should}. Once the surrogate model $\xi(\mathbf{x})$ is trained, explanation on the prediction of $f$ on $\mathbf{x}$ can be obtained by examining the interpretable function $\xi(\mathbf{x})$.

There are several works extending this idea to explain GNNs~\cite{huang2020graphlime,zhang2021relex,vu2020pgm}. They differ from each other mainly in two aspects: the strategy of obtaining local prediction records ($\pi_x$), and the interpretable model family ($F'$) selected as candidate surrogates. For example,  GraphLime~\cite{huang2020graphlime} takes neighboring nodes as perturbed inputs. And it employs a nonlinear surrogate model which can assign large weights to features that are important in inference. However, GraphLime ignores graph structures and can only find important node attributes. PGM-Explainer~\citep{vu2020pgm} randomly perturbs node attributes to collect local records, and trains a probabilistic graph model (PGM) to fit them. PGM is interpretable and can show the dependency among nodes inside input graph. RelEx~\citep{zhang2021relex} randomly samples subgraphs as inputs, and uses GCN as the surrogate model. It can assign importance weights to edges, but requires an additional running of explanation methods on GCN as it is also non-interpretable. 

\subsubsection{Model-level Post-hoc Explanation}

Compared to instance-level methods, model-level methods focus more on providing general insights by giving high-level explanations that are independent to inputs for deep graph models. The model-level explanations can be representative and discriminative instances for each class~\cite{yuan2020xgnn,zhang2021protgnn, huang2023global} or logic rules to depict knowledge captured by deep model~\cite{yang2019learn}. However, due to the highly diverse topology and complex semantics in the graph domain, it is a very challenging task and few attempts are made to provide model-level post-hoc explanations for GNNs. XGNN~\cite{yuan2020xgnn} aims to expose what input graph patterns can trigger certain predictions of the target GNN model, and adopt a graph generation module to achieve that. They employ input optimization methods and train a graph generator to generate graphs that can maximize the target predictions using reinforcement learning. After training, generated graphs are expected to be representative of each class and can provide global knowledge on the captured knowledge of the target GNN model. Concretely, the desired prototypical explanation for class $c$  is obtained by solving the following objective:
\begin{equation}
    \mathcal{G}^{*} = \arg\max_{\mathcal{G}} P(f_t(\mathcal{G}) = c),
\end{equation}
where $f_t$ denotes the target GNN model, and a graph generator is trained for finding $\mathcal{G}^{*}$ for class $c$. Another work, GCFExplainer~\cite{huang2023global}, explore the global explainability of GNNs using global counterfactual reasoning, aiming to identify a concise set of representative counterfactual graphs that elucidate all input graphs. To achieve this, they employ vertex-reinforced random walks on a graph's edit map and use a greedy technique to get the summarization for each class.

\subsubsection{Self-explainable Approaches}
Different from the post-hoc explanation, self-explainable approaches~\cite{zhang2021protgnn,dai2021towards,wu2020graph} aim to give predictions and provide explanations for each prediction simultaneously. Specific GNN architectures are adopted to support built-in interpretablity, similar to attention mechanism in GAT~\cite{velivckovic2017graph}. However, although these methods are explainable by design, they are often restricted in the modeling space and struggle to generalize across tasks at the same time. There are several representative self-explainable GNNs proposed recently, i.e., SE-GNN~\cite{dai2021towards}, GIB~\cite{wu2020graph}, and ProtGNN~\cite{zhang2021protgnn}. Many causal-based methods can also be categorized into self-explainable methods, i.e., DIR~\cite{wu2022discovering}, DisC~\cite{fan2022debiasing} and CIGA~\cite{chen2022learning}.

SE-GNN~\cite{dai2021towards} focuses on instance-level self-explanation. It obtains the self-explanation for node classification via identifying interpretable $K$-nearest labeled nodes for each node and utilizes the $K$-nearest labeled nodes to simultaneously give label prediction and explain why such prediction is given. More specifically, SE-GNN~\cite{dai2021towards} adopts an interpretable similarity modeling to compute the attribute similarity and local structure similarity between the target nodes and labeled nodes. A contrastive pretext task is further deployed in SE-GNN to provide self-supervision for interpretable similarity metric learning. GIB~\cite{wu2020graph} balances expressiveness and robustness of the learned graph representation by learning the minimal sufficient representation for a given task. Following the general information bottleneck, it maximizes the mutual information between the representation and the target, and simultaneously constrains the mutual information between the representation and the input data. ProtGNN~\cite{zhang2021protgnn} is better at global-level explanations by finding several prototypes for each class. Newly-coming instances are classified via comparing with those prototypes in the embedding space. A conditional subgraph sampling module is designed to conduct subgraph-level matching and several regularization terms are used to promote diversity of prototypical embeddings. DIR~\cite{wu2022discovering} utilizes the intrinsic interpretability of graph neural networks and aim to identify a subset of input graph features, termed "rationale", to guide the model predictions. They formulate the problem into a invariant learning problem and design distribution intervener to get the rationales. DisC~\cite{fan2022debiasing} is a disentangled GNN framework that seperates input graphs into causal and bias substructures, using a parameterized edge mask generator and training two GNN modules with respective loss functions. CIGA~\cite{chen2022learning} provides an alternative causal-based framework, which can capture graph invariance for reliable OOD generalization across diverse distribution shifts. CIGA utilizes an information-theoretic objective to identify subgraphs enriched with invariant intra-class information, ensuring resilience to distribution shifts.

\subsection{Datasets for Explainability of GNNs}
For comparing various explanation methods, it is desired to have datasets where the rationale between input graphs and output labels are intuitive and easy-to-obtain, so that it would be easier to evaluate identified explanatory substructures. In this subsection, we summarize popular datasets used by existing works on explainable GNNs, which can roughly be categorized into synthetic datasets and real-world datasets. Several representative benchmarks of both groups will be introduced in detail.

% maybe we can consider the category of node classification and graph classification, or like synthetic or real datasets, or with groundtruth or without groundtruth

% \textcolor{red}{datasets}

% \subsection{Synthetic Data}

% \begin{itemize}
%     \item BA-shapes
%     \item BA-Community
%     \item Tree-Cycle
%     \item Tree-Grids
%     \item BA-2Motifs
% \end{itemize}

% \subsection{Real World Data}

% \subsubsection{Sentiment graph Data}

% \begin{itemize}
%     \item Graph-SST2
%     \item Graph-SST5
%     \item Graph-Twitter
% \end{itemize}

% \subsubsection{Molecular Data}

% \begin{itemize}
%     \item MUTAG
%     \item BBBP
%     \item Tox21 
% \end{itemize}

% Synthetic test
% Node classification follow GNNExplainer, BA-shape
% Benzene ring test
% QM9 for graph classification
% Removal test.
% Remove one to five in the order of importance of the edges and get AUC
% Graph classification
% Scene graph, country vs. urban, outdoor vs. indoor
% Graph classification.
% Molecular graph binary classification, BBBP, BACE, NR-ER
% Synthetic dataset
%  BA graph
% Others.
% Activation task, pruning task, sentiment analysis task, SchNet task, VGG-16
% Graph classification
% Molecular graph, BBBP, BACE, NR-ER
% Graph classification
% Scene graph, Visual Genome

\subsubsection{Synthetic Data}
With carefully-designed graph generation mechanism, we can constrain unique causal relations between input elements and provided labels in synthetic datasets. GNN models must capture such patterns for a successful training and obtained explanations are evaluated with those ground-truth causal substructures. Several common synthetic datasets are listed below:
\begin{itemize}[leftmargin=*]
    \item \textbf{BA-Shapes}~\cite{ying2019gnnexplainer}: It is a single graph consisting of a base Barabasi-Albert (BA) graph (contains $300$ nodes) and $80$ ``house”-structured motifs (contains $5$ nodes). ``House'' motifs are randomly attached to the base BA graph. Nodes in the base graph are labeled as $0$ and those in the motifs are labeled as $1,2, \text{and } 3$ based on their positions. Explanations are evaluated on attached nodes of motifs, with edges inside the corresponding motif as ground-truth.
    \item \textbf{Tree-Cycles}~\cite{ying2019gnnexplainer}: It is a single network with a $8$-layer balanced binary tree as the base graph. $80$ cycle motifs (contains $6$ nodes) are randomly attached to the base graph. Nodes in the base graph are labeled as $0$ and those in the motifs are labeled as $1$. Ground-truth explanations for nodes within cycle motifs are provided for evaluation.
    \item \textbf{BA-2motifs}~\cite{ying2019gnnexplainer}: This is a graph classification dataset containing $800$ graphs. Half of the graphs are constructed by attaching a ``house'' motif to BA base graphs, while the other half graphs attach a five-node cycle motif. A binary label is assigned to each graph according to its attached motif. The motif serves as the ground-truth explanation.
    \item \textbf{Infection}~\cite{faber2021comparing}: This is a single network initialized with a ER random graph. $5\%$ of nodes are labeled as infected (class $0$), and the remaining nodes are labeled as their shortest distances to those infected ones. In evaluation, nodes with multiple shortest paths are discarded. All remaining nodes have one distinct path as the oracle explanation towards their labels. 
    \item \textbf{Syn-Cora}~\cite{dai2021towards}:  This is synthesized from the Cora~\cite{kipf2016semi} to provide ground-truth of explanations, i.e., $K$-nearest labeled nodes and edge machining results. To construct the graph, motifs are obtained by sampling local graphs of nodes from Cora. Various levels of noises are applied to the motifs in attributes and structures to generate similar local graphs.
\end{itemize}

\subsubsection{Real-world Data}
Due to high complexity in patterns and possible existence of noises, it is challenging to obtain human-understandable rationale from node features and graph topology to labels for real-world graphs. Typically, strong domain knowledge is needed. Thus, real-world graph datasets with groundtruth explanations are limited.  Below are two benchmark datasets:
\begin{itemize}[leftmargin=*]
    \item \textbf{Molecule Data}~\cite{wu2018moleculenet}: This is a graph classification dataset. Each graph corresponds to a molecule with nodes representing atoms and edges for chemical bonds. Molecules are labeled with consideration of their chemical properties, and discriminative chemical groups are identified using prior domain knowledge. Chemical groups \textit{$NH_2$} and \textit{$NO_2$} are used as ground-truth explanations.
    \item \textbf{Sentiment Graphs}~\cite{yuan2020explainability}:  It contains three graph classification datasets created from text datasets for sentiment analysis, i.e., Graph-SST3, Graph-SST5 and Graph-Twitter. Each graph is a text document where nodes represent words and edges represent relationships between word pairs constructed from parsing trees. Node attributes are set as word embeddings from BERT~\cite{devlin2018bert}. There is no ground-truth explanation provided. Heuristic metrics are usually adopted for evaluation.
\end{itemize}

A summary of representative benchmark datasets is provided in Table~\ref{tab:expl_dataset} along with their key statistics, tasks and papers used the datasets.

\begin{table}[t]
    \scriptsize
    \centering
    \caption{Datasets for Explainability of GNNs}
    \vskip -1.5em
    % \resizebox{0.98\linewidth}{!}{
    \begin{tabular}{llllllll}
    \toprule
    Tasks &  Dataset &  Graphs & Avg.Nodes & Avg.Edges & Features & References\\
    \midrule
    \multirow{5}{*}{Node Classification} & BA-Shapes &  1 & 700 & 4,110 & 10 & \makecell[l]{\cite{ying2019gnnexplainer}~\cite{luo2020parameterized}~\cite{yuan2021explainability}~\cite{vu2020pgm}~\cite{dai2021towards}~\cite{zhang2021protgnn}}\\
    % \midrule
    &BA-Community  & 1  & 1,400 & 8,920 & 1 & \makecell[l]{\cite{ying2019gnnexplainer}~\cite{luo2020parameterized}~\cite{funke2020hard}}\\  
    % \midrule
    &Tree-Cycles  & 1  & 871   & 1,950 & 10 & \makecell[l]{\cite{ying2019gnnexplainer}~\cite{luo2020parameterized}~\cite{vu2020pgm}}\\ 
    % \midrule
    &Tree-Grid & 1  & 1,231 & 3,410 & 10 & \makecell[l]{\cite{ying2019gnnexplainer}~\cite{luo2020parameterized}~\cite{vu2020pgm}}\\ 
    &Syn-Cora & 1 &1,895	 & 2,769 & 1,433 & \makecell[l]{\cite{dai2021towards}} \\
    \midrule
    \multirow{6}{*}{Graph Classification}
    &BA-2motifs & 1,000  & 25 & 51.4  & 10 & \makecell[l]{\cite{luo2020parameterized}~\cite{yuan2021explainability}}\\
    % \midrule
    &Infection & 10  & 1000 & 3996 & 2 & \makecell[l]{\cite{faber2021comparing} ~\cite{luo2020parameterized}}\\
    % \midrule
    &Graph-SST2 & 70,042  & 10.199 & 9.20 & 768 & \makecell[l]{\cite{yuan2020explainability}~\cite{yuan2021explainability}~\cite{zhang2021protgnn}}\\
    % \midrule
    &Graph-SST5 &  11,855  & 19.849 & 18.849 & 768 & \makecell[l]{\cite{yuan2020explainability}}\\
    % \midrule
    &Graph-Twitter & 6,940  & 21.103 & 21.10 & 768 & \makecell[l]{\cite{yuan2020explainability}~\cite{zhang2021protgnn}}\\
    % \midrule
    &MUTAG & 188  & 19.79 & 17.93 & 14 & \makecell[l]{\cite{debnath1991structure}~\cite{ying2019gnnexplainer}~\cite{luo2020parameterized}~\cite{yuan2021explainability}~\cite{zhang2021relex}~\cite{yuan2020xgnn}~\cite{zhang2021protgnn}}\\

    %\midrule
    %BBBP & \makecell[l]{Graph Classification} & 2039  & 24.06 & 25.95 & ??? & ??? &  \cite{yuan2020explainability}\\
    %\midrule
    %Tox21 & \makecell[l]{Graph Classification} & ???  & ??? & ??? & ??? & ??? &  \cite{wu2018moleculenet}\\
    % Reddit-binary gnnexplainer
    % scene graph classification
    % reddit-multi-5k causal screening
    % tree-grid-ba relex
    % pgmexplainer ba-grid ba-bottle
    % xgnn is_acyclic
    % protgnn bbbp 
    
    \bottomrule
    \end{tabular}
    \label{tab:expl_dataset}
    \vskip -1em
\end{table}

\subsection{Evaluation Metrics}\label{sec:expl_evaluation}
Besides visualizing identified explanations and conducting expert examinations, several metrics have been proposed for quantitatively evaluating explanation approaches from different perspectives. Next, we would present the major categories of metrics used and introduce their distinctions.
\begin{itemize}[leftmargin=*]
    \item \textit{Explanation Accuracy}. For graphs with ground-truth rationale known, one direct evaluation method is to compare identified explanatory parts with the real causes of the label~\cite{ying2019gnnexplainer,luo2020parameterized}. F1 score and ROC-AUC score can both be computed on identified edges. The higher these scores are, the more accurate obtained explanation is. 
    \item \textit{Explanation Fidelity}. In the absence of ground-truth explanations, heuristic metrics can be designed to measure the fidelity of identified substructures~\cite{yuan2020explainability}. The basic idea is that explanatory substructures should play a more important role in predictions. Concretely, Fidelity+~\cite{yuan2020explainability} is computed by removing all input elements first, then gradually adding features with the highest explanation scores. Heuristically, a faster increase in GNN's prediction indicates stronger fidelity of obtained explanations. In turn, Fidelity-~\cite{yuan2020explainability} is computed by sequentially removing edges following assigned importance weight. In turn, a faster performance drop represents stronger fidelity of removed explanations.
    \item \textit{Sparsity}. Good explanations are expected to be minimal structure explanations, as only the most important input features should be identified. This criteria directly measures sparsity of obtained explanation weights~\cite{funke2021zorro}, and a better explanation should be sparser. 
    \item \textit{Explanation Stability}. As good explanations should capture the intrinsic rationale between input graphs and their labels, this criterion requires identified explanations to be stable with respect to small perturbations~\cite{agarwal2021towards}. Stability score can be computed via comparing explanation changes after attaching new nodes or edges to the original graph. However, GNN's prediction is sensitive towards input. Thus, it is challenging to select a proper amount of perturbations.
    
\end{itemize}

\subsection{Future Research Directions on Explainability of GNNs}
%GNN explanation is a fast-growing research field, with many methods being proposed each year. A systematic overview of existing works is presented in previous sections. 
In this part, we provide our opinions on some promising future research directions. We hope it can inspire and encourage the community to work on bridging the gaps in GNN interpretation.     

\vspace{0.2em} \noindent \textbf{Class-level Explanations}. Despite many interpretation approaches, class-wise explanations remain an under-explored area. Instance-level explanations provide a local view of GNN's prediction. However, it is important to recognize that they may provide anecdotal evidence and scale poorly to large graph sets. On the other hand, class explanations could provide users with both a global view of model's behavior and a discriminative view grounded in each class, making it easier to expose and evaluate learned knowledge.

\vspace{0.2em} \noindent \textbf{Benchmark Datasets for Interpretability}. One great obstacle in designing and evaluating interpretation methods is the measurement of provided interpretability. To a large extent, the difficulty is inherent. For example, it is impossible to provide gold labels for what is a correct explanation. After all, we would not need to design an explanation approach in the first place had we known those real explanations. Currently, we rely upon heuristic metrics and approximations, but a set of principled proxy measurements and benchmark datasets are yet to be established.

\vspace{0.2em} \noindent \textbf{User-Oriented Explanations}. The purpose of designing explanation approaches is to use them on real-world tasks to expose learned knowledge of trained models. Based on the needs of users, different requirements may arise in the form of explanation and levels of interpretability. Hence, we encourage researchers to consider real user cases, select suitable design choices, and evaluate explanation algorithms in real-world scenarios for the integrity in interpretability field. One promising direction is to provide flexible fine-grained multi-level explanations so that end-users can select the level of explanations they can understand or satisfy their criteria.

\vspace{0.2em} \noindent \textbf{Causal and Counterfactual Explanations}. While traditional explanations offer insights into model behavior, causal and counterfactual explanations dig deeper into understanding the cause-and-effect relationships that drive predictions. Causal explanations focus on identifying the direct influences or drivers behind a particular prediction, while counterfactual explanations provide insights by answering "what if" scenarios, showcasing how a prediction might change if input features were altered. Incorporating such explanations in GNNs can be beneficial, especially in critical applications where understanding the underlying causes or potential outcomes of a decision is crucial. It is paramount to develop methods that can reliably extract and present these types of explanations, ensuring that users not only understand what the model predicts but also the underlying reasons and potential alternatives.
\section{Conclusion}
In this survey, we conduct a comprehensive review on the trustworthy GNNs from the aspects of privacy, robustness, fairness, and explainability. This fills the gap in lacking systematically summary about privacy-preserving GNNs and fairness-aware GNNs. For the robustness and explainability, we introduce the recent trends in more details apart from the representative methods that are reviewed before. 
More specifically, for each aspect, we introduce the core definitions and concepts to help the readers to understand the defined problems. The introduced methods are categorized from various perspectives. 
The unified framework of each category is generally given followed by the detailed implementations of the representative methods. In addition, we also list the used datasets in privacy, fairness, and explainability, where the proposed methods have special requirements on the datasets to be trained or evaluated.  Numerical real-world applications of trustworthy GNNs are also provided to encourage the researcher to develop practical trustworthy GNNs. Finally, we discuss the future research directions of each aspect at the end of each section, which includes promising directions in a single aspect and interactions between aspects for trustworthy GNNs.

\bibliographystyle{acm}
\bibliography{ref}
\end{document}